\documentclass{article} 
\usepackage{iclr2027_conference,times}

\usepackage{amsmath,amsfonts,bm}

\def\eqref#1{equation~\ref{#1}}

\def\1{\bm{1}}

\DeclareMathAlphabet{\mathsfit}{\encodingdefault}{\sfdefault}{m}{sl}
\SetMathAlphabet{\mathsfit}{bold}{\encodingdefault}{\sfdefault}{bx}{n}

\newcommand{\E}{\mathbb{E}}

\newcommand{\R}{\mathbb{R}}

\newcommand{\KL}{D_{\mathrm{KL}}}
\newcommand{\Var}{\mathrm{Var}}

\newcommand{\Cov}{\mathrm{Cov}}

\usepackage{hyperref}
\usepackage{url}

\usepackage{amsmath}
\usepackage{amssymb}
\usepackage{amsthm}
\usepackage{mathtools}
\usepackage{bm}
\usepackage{tikz}
\usetikzlibrary{positioning,arrows.meta,calc}

\usepackage{graphicx}
\usepackage{booktabs}
\usepackage{algorithm}
\usepackage{algpseudocode}
\usepackage{placeins}
\usepackage{subcaption}

\usepackage{tabularx}
\usepackage{array}

\providecommand{\R}{\mathbb{R}}
\providecommand{\E}{\mathbb{E}}
\providecommand{\KL}{\operatorname{KL}}

\providecommand{\Cov}{\operatorname{Cov}}
\providecommand{\Var}{\operatorname{Var}}
\providecommand{\Range}{\operatorname{Range}}

\providecommand{\cvec}{\bm c}
\providecommand{\etavec}{\bm\eta}
\providecommand{\lambdavec}{\bm\lambda}
\providecommand{\Phivec}{\bm\Phi}

\providecommand{\Fib}{\mathcal F}
\providecommand{\Risk}{\mathcal R}
\providecommand{\Act}{\mathcal A}

\providecommand{\norm}[1]{\left\lVert #1 \right\rVert}

\newtheorem{theorem}{Theorem}
\newtheorem{proposition}{Proposition}
\newtheorem{corollary}{Corollary}
\newtheorem{lemma}{Lemma}
\newtheorem{remark}{Remark}

\newtheorem{example}{Example}

\title{FIDE: Fiber-Informed Differentiable Experimental Design for Reference-Relative Law Dynamics}

\author{
Pietro Zanotta \& J{\'a}n Drgo{\v{n}}a \\
Department of Civil and Systems Engineering \\
Johns Hopkins University \\
Baltimore, MD, USA \\
\texttt{pzanott1@jh.edu}
}

\iclrfinalcopy 
\begin{document}

\maketitle
\lhead{Preprint}

\begin{abstract}
    Aggregate measurements generally leave an evolving population law underdetermined, so different measurement systems can be equally adequate for a scientific task while constraining different aspects of the law. We study how to resolve this residual design freedom in baseline-preserving settings, where an existing dynamical model remains the working baseline outside quantities explicitly valued by the task. \textbf{F}iber-\textbf{I}nformed \textbf{D}ifferentiable \textbf{E}xperimental Design (\textbf{FIDE}) completes each candidate's aggregate constraints by information projection relative to a common frozen reference, then ranks scientifically adequate designs by the dynamical revision required to realize the resulting completed paths. The resulting \emph{Full action} is the minimum mean-squared velocity correction needed to realize a completed path; equivalently, its square root is the sharp reference-relative expectation-rate discrepancy over a gradient-normalized Sobolev class. This defines a conditional minimum-revision principle for baseline-preserving design rather than a generic truth-recovery objective. Experiments show that this preference changes sensor selection, can help or harm withheld downstream predictions depending on baseline reliability, can be used prospectively from aggregate predictions, and remains computationally usable for a $32$-dimensional configuration law.
\end{abstract}

\section{Introduction}
\label{sec:introduction}

Consider an experiment that deploys a small number of aggregate sensors to answer a prespecified scientific question about an evolving population. Several sensor layouts may achieve essentially the same task performance while constraining different aspects of the intermediate law. Because aggregate measurements generally do not identify the full law, these task-equivalent designs can imply different conclusions outside the quantities used to evaluate the task. The primary objective tells us which designs are scientifically adequate, but not how to choose among these remaining law-level differences. Resolving that choice requires an explicit secondary design preference.

Classical and modern optimal experimental design typically chooses measurements to improve information gain, posterior uncertainty, reconstruction quality, or a task-specific utility \citep{chaloner1995bayesian,foster2019variational,attia2018goal}. Our question begins after such a primary objective has already identified several scientifically adequate designs. We focus on the baseline-preserving regime: an existing dynamical model is the declared working baseline, and the present experiment is scoped to improve or assess prescribed quantities. Among designs that satisfy that task, we prefer not to introduce additional departures from the baseline unless those departures are valued by another objective or supported by independent evidence.

This regime is natural when new sensors augment an existing modeling or decision pipeline rather than replace it. Related model-credibility and observing-system frameworks likewise evaluate models and measurements within a specified context of use \citep{morrison2019assessing,viceconti2021silico,hoffman2016future,mourre2012benefit}. Once the scientific task is already satisfied, leaving the residual choice unspecified means that it may be determined by incidental details of the parameterization, optimization procedure, or sensor family rather than by a declared design preference. Our formulation makes that preference explicit: preserve the working baseline in directions left unresolved by the task unless another objective provides a reason to depart from it. Departures that are themselves scientifically valuable should therefore enter the primary risk.

Within this baseline-preserving setting, we propose \textbf{F}iber-\textbf{I}nformed \textbf{D}ifferentiable \textbf{E}xperimental Design (\textbf{FIDE}) to resolve the residual design choice. Aggregate measurements do not determine a unique population law, so FIDE first information-projects each candidate's constraints relative to the same frozen reference \citep{kullback1951information,jaynes1957information,csiszar1975idivergence}. This gives every candidate a completed law path under the same reference-relative rule before any dynamical score is computed.

The completed path need not follow the frozen reference dynamics, so FIDE asks how much those dynamics must change to realize it. The \emph{Full action} is the minimum mean-squared velocity correction required to do so. Equivalently, its square root is the largest reference-relative error in expectation rate, per unit gradient energy, over a normalized class of smooth observables. Thus, after scientific adequacy has been enforced, Full measures dynamical revision of the declared baseline across a broad observable class.

Our contributions are:
\begin{itemize}
    \item \textbf{Residual experimental design.} We formulate baseline-preserving measurement selection as a hierarchical $\varepsilon$-constraint problem in which scientific risk determines adequacy and minimum dynamical revision resolves only the remaining design freedom.
    \item \textbf{Reference-relative dynamical revision.} We characterize Full as the minimum velocity-correction energy required to realize the completed law path, with its square root equivalently giving a sharp expectation-rate discrepancy over a normalized class of smooth observables, and derive its Full--Tangent--Hidden decomposition and matched-initial stability bound.
    \item \textbf{Differentiable design optimization.} For continuous designs, we establish local end-to-end differentiability of the finite-dimensional FIDE objective under explicit nonsingularity conditions and differentiate through both information-projection calibration and the Full correction solve.
    \item \textbf{Empirical characterization and scope.} We study criterion-dependent sensor selection, downstream effects of baseline preservation, prospective selection from aggregate predictions, and a $32$-dimensional configuration-law application.
\end{itemize}

\section{Problem Formulation} \label{sec:problem-formulation} Let $t\in[0,1]$ denote normalized experimental time and let $X_t\in\mathcal X$ have unknown population law $P_t$, where $\mathcal X$ is a domain in $\R^d$ or a flat periodic domain. A measurement design $\etavec\in\mathcal D$ specifies a time-independent observable map $\Phivec_{\etavec}$ and is paired, for design purposes, with an aggregate moment trajectory $\cvec_{\etavec}(t)\in\R^R$. At each time, these moments define the fiber
\begin{equation}
    \Phivec_{\etavec}:\mathcal X\to\R^R,\qquad \Fib_{\etavec}(\cvec):=\left\{Q\in\mathcal P(\mathcal X):\E_Q[\Phivec_{\etavec}]=\cvec\right\}. \label{eq:target-moments} 
\end{equation}

In the ideal population formulation, $\cvec_{\etavec}(t)=\E_{P_t}[\Phivec_{\etavec}(X_t)]$. In applications, $\cvec_{\etavec}(t)$ denotes the candidate-specific aggregate trajectory supplied to FIDE; Section~\ref{sec:methodology} specifies how these inputs are obtained retrospectively or prospectively. The observable coordinates may represent sensor responses, projections, localized windows, or coefficients of an observable subspace. Because finitely many expectations generally do not identify a full law, changing the design changes both what is constrained and which law directions remain unresolved. Here $\mathcal P(\mathcal X)$ denotes probability laws for which the displayed expectations exist.

Before design optimization, we fix a reference law path $\widetilde Q_t$ with density $\widetilde q_t$ and velocity $u_t$ satisfying
\begin{equation}
\partial_t\widetilde q_t+\nabla\!\cdot(\widetilde q_tu_t)=0.
\label{eq:frozen-reference-continuity}
\end{equation}

The equation is understood weakly under the boundary or decay conditions of Section~\ref{sec:theory}. The pair $(\widetilde Q_t,u_t)$ is held fixed as $\etavec$ varies and is not assumed to reproduce the unknown intermediate population dynamics. Full is defined relative to this entire law--velocity pair, since different velocity fields realizing the same reference marginals can induce different dynamical revision costs.

Scientific adequacy is specified independently by an application-dependent design-stage risk $\Risk:\mathcal D\to\R$, fixed before design optimization. This risk determines which designs are adequate; Full is used only to rank the design freedom that remains after adequacy has been enforced. Evaluating $\Risk(\etavec)$ may require additional candidate-specific summaries beyond $\cvec_{\etavec}(t)$, such as response covariances or predicted scientific quantities, which are treated as design-stage inputs. Any departure that should affect the design choice must therefore enter the scientific risk or an additional objective. Transfer of scientific adequacy to post-freeze observations is evaluated separately.

\section{Methodology}
\label{sec:methodology}
FIDE takes two conceptually distinct inputs. First, each candidate design $\etavec$ is paired with an aggregate trajectory $\cvec_{\etavec}(t)$ and the summaries needed to evaluate its scientific risk $\Risk(\etavec)$. Retrospectively these may come from existing data or simulation, while prospectively they are supplied by a predictive interface fixed before optimization. Second, the frozen pair $(\widetilde Q_t,u_t)$ supplies the common dynamical baseline against which every candidate is evaluated. The predictive interface therefore answers \emph{what would this measurement system constrain?}, whereas the frozen reference answers \emph{how much dynamical revision would those constraints require?} Given these inputs, FIDE applies the same three steps to every candidate: complete its aggregate constraints, measure the resulting dynamical revision, and minimize that revision among scientifically adequate designs.

\paragraph{Step 1: complete the moment constraints.} \label{subsec:methodology-iprojection} For each $(t,\etavec)$ admitting the projection, define \begin{equation} Q_t^{\etavec}\in\operatorname*{arg\,min}_{Q\ll\widetilde Q_t}\left\{\KL(Q\Vert\widetilde Q_t):\E_Q[\Phivec_{\etavec}]=\cvec_{\etavec}(t)\right\}. \label{eq:iprojection} \end{equation} The aggregate trajectory fixes the measured expectations, while information projection fills in the remaining law directions relative to the common reference. Thus $\widetilde Q_t$ is fixed across designs and $Q_t^{\etavec}$ is the completed law associated with design $\etavec$. The ordering of these operations is deliberate: FIDE first completes every candidate by the same rule and only then asks how dynamically costly that completed path is. This keeps the completion rule common across designs rather than allowing the dynamical score itself to choose unresolved law directions. A path-level minimum-action completion defines the corresponding coupled alternative; Appendix~\ref{app:path-completion-ablation} quantifies the difference on a controlled example. Theorem~\ref{thm:iprojection} gives the corresponding existence, uniqueness, differentiability, and exponential-tilt results.

\paragraph{Step 2: measure dynamical revision.}
\label{subsec:methodology-full}
Let $q_t^{\etavec}$ denote the projected density. The information projection fixes a law path, but the frozen reference velocity $u_t$ need not transport that path. FIDE therefore asks for the smallest velocity correction $\delta_t$ such that $u_t+\delta_t$ does. We define \begin{align} 
a_{\mathrm{full}}(t,\etavec)&:=\inf_{\delta\in L^2(Q_t^{\etavec};\R^d)}\left\{\E_{Q_t^{\etavec}}[\norm{\delta}^2]:\partial_tq_t^{\etavec}+\nabla\!\cdot\bigl(q_t^{\etavec}(u_t+\delta)\bigr)=0\right\}, \label{eq:full-action-density}\\ \Act(\etavec)&:=\int_0^1a_{\mathrm{full}}(t,\etavec)\,\rho(dt), \label{eq:full-action} 
\end{align}

where $\rho$ is a fixed probability measure on $[0,1]$, the continuity constraint is understood weakly under the prescribed boundary conditions, and $\inf\varnothing=+\infty$. We call $a_{\mathrm{full}}(t,\etavec)$ the instantaneous Full action density and $\Act(\etavec)$ the time-integrated Full action. Only the correction $\delta_t$ is optimized; the reference and completed path remain fixed. For later analysis, whenever the required density derivatives exist on the positive support of $q_t^{\etavec}$, define the normalized continuity forcing $h_{t,\etavec}:=[\partial_tq_t^{\etavec}+\nabla\!\cdot(q_t^{\etavec}u_t)]/q_t^{\etavec}$. The continuity constraint is then equivalent to $\nabla\!\cdot(q_t^{\etavec}\delta_t)=-q_t^{\etavec}h_{t,\etavec}$. Theorem~\ref{thm:poisson-correction} gives the corresponding weak and weighted-Poisson characterizations.

\paragraph{Step 3: select among scientifically adequate designs.}
\label{subsec:methodology-fide}
The observable family $\mathcal D$, scientific risk $\Risk$, tolerance $\varepsilon$, frozen reference $(\widetilde Q_t,u_t)$, information-projection rule, and Full correction geometry are fixed before optimizing $\etavec$. Let $\Risk^\star:=\inf_{\etavec\in\mathcal D}\Risk(\etavec)$ and assume $0<\Risk^\star<\infty$. For $\varepsilon\ge0$, define
\begin{equation} 
\mathcal D_\varepsilon:=\left\{\etavec\in\mathcal D:\Risk(\etavec)\le(1+\varepsilon)\Risk^\star\right\},\qquad \etavec_{\mathrm{Full}}\in\operatorname*{arg\,min}_{\etavec\in\mathcal D_\varepsilon}\Act(\etavec). \label{eq:fide-design} 
\end{equation}
Scientific risk therefore determines feasibility, and Full ranks only admissible designs. This gives FIDE a hierarchical structure: scientific adequacy is enforced first through the $\varepsilon$-constraint, and Full then resolves the remaining design freedom. The following elementary property makes the reference-relative meaning of this secondary criterion explicit.

\begin{proposition}[Reference consistency and scope]
\label{prop:reference-consistency}
Fix a design $\etavec$ for which the information projection is well defined. If $\cvec_{\etavec}(t)=\E_{\widetilde Q_t}[\Phivec_{\etavec}]$ for every $t$, then $Q_t^{\etavec}=\widetilde Q_t$ and $a_{\mathrm{full}}(t,\etavec)=0$ for every $t$, hence $\Act(\etavec)=0$. If an alternative population law $P_t\neq\widetilde Q_t$ also satisfies $\E_{P_t}[\Phivec_{\etavec}]=\cvec_{\etavec}(t)$ for every $t$, this zero action does not imply $Q_t^{\etavec}=P_t$. Thus zero Full certifies consistency with the declared reference under the supplied constraints; it does not identify among alternative laws that share those constraints. 
\end{proposition}

The first claim follows because the reference itself satisfies the moment constraints and attains zero relative entropy, so its frozen velocity realizes the projected path with zero correction. Intuitively, the candidate measurements then ask for nothing that the reference does not already predict, so neither the completed law nor its dynamics need to change. Existence qualifications for the outer minimizer and the finite-sample Law baseline are collected in Appendix~\ref{app:design-qualifications}.

\paragraph{Comparator criteria.} For empirical comparison, we also select designs using Tangent, residual $L^2$, forward KL, and static $W_2$ as secondary criteria, always on the same information-projected law path and within the same admissible set $\mathcal D_\varepsilon$. These criteria probe different properties of the completed path and their definitions are collected in Appendix~\ref{app:comparator-definitions}, while Section~\ref{subsec:theory-comparisons} summarizes how they differ from Full.

\paragraph{Numerical implementation and design gradients.}
\label{subsec:methodology-design-gradient}
In finite-sample experiments, we use \emph{Law} to denote selection by scientific risk alone. We replace the population optimum $\Risk^\star$ by the selection risk $R_{\mathrm{Law}}$ of the numerically optimized Law design and use the operational admissible set $R_{\mathrm{sel}}(\etavec)\le(1+\varepsilon)R_{\mathrm{Law}}$. This is a finite-search approximation rather than a certificate of global risk optimality. For each candidate, the numerical pipeline calibrates the empirical information projection, evaluates scientific risk and the chosen secondary criterion, and optimizes over the resulting admissible designs. All candidate-specific design-stage inputs are fixed before optimization; post-freeze target observations are used only for validation and are never fed back into candidate scoring.
For an empirical reference $\sum_{i=1}^N\bar w_i\delta_{x_i}$ with $z_i=\Phivec_{\etavec}(x_i)$, exact calibration requires $\widehat c\in\operatorname{conv}\{z_1,\ldots,z_N\}$. We distinguish this mathematical \emph{calibration validity} from \emph{support certification}: a feasible calibrated law may still be poorly supported for estimating Full action. We therefore monitor effective sample size and covariance conditioning separately. Proposition~\ref{prop:empirical-iprojection-feasibility} gives the precise feasibility result, while benchmark-specific discretizations and numerical audits are deferred to the corresponding experimental appendices.

FIDE allows both discrete and continuous designs. For continuous designs, the fixed finite-dimensional objective is locally differentiable end to end whenever the calibrated covariance and reduced Full systems remain nonsingular and vary smoothly with the design. In particular, if the reduced Full solve is $\mathsf K_{\etavec}\beta_{\etavec}=b_{\etavec}$, then for any design perturbation $\xi$ the instantaneous action satisfies $D_{\etavec}\widehat a_{\mathrm{full}}[\xi]=2\beta_{\etavec}^{\top}D_{\etavec}b_{\etavec}[\xi]-\beta_{\etavec}^{\top}D_{\etavec}\mathsf K_{\etavec}[\xi]\beta_{\etavec}$. Thus gradients propagate through the information-projection calibration, physical-time multiplier dynamics, and Full correction solve without differentiating through the iterations of the underlying solvers. Under these local conditions, continuous sensor locations or other differentiable design parameters can therefore be optimized directly with gradient-based methods. Proposition~\ref{prop:end-to-end-design-differentiability} gives the full conditions and integrated-action result.

\section{Theoretical Analysis}
\label{sec:theory}

This section develops the properties of Full needed for the residual-design interpretation. We first derive the information-projected path and its reference-relative continuity forcing. We then show that the minimum quadratic correction used in Section~\ref{sec:methodology} is exactly a sharp Sobolev expectation-rate discrepancy, identify Tangent as its restriction to the measured observable span, and finally relate Full to the remaining comparators and to matched-initial stability.

\paragraph{Information-projected path and continuity forcing}
\label{subsec:theory-iprojection}

For a fixed design, write $\phi=\Phivec_{\etavec}$, $Q_t=Q_t^{\etavec}$, $q_t=q_t^{\etavec}$, $c_t=\cvec_{\etavec}(t)$, $\lambda_t=\lambdavec_{t,\etavec}$, and $h_t=h_{t,\etavec}$, and define $A_t(\lambda)=\log\E_{\widetilde Q_t}[e^{\lambda^\top\phi}]$, $m_t=J_\phi u_t$, and $C_t=\Cov_{Q_t}(\phi,\phi)$.

\begin{theorem}[Information-projected lift and forcing]
\label{thm:iprojection}
Assume that $c_t$ lies in the relative interior of the feasible moment set, that the dual calibration objective $A_t(\lambda)-\lambda^\top c_t$ attains a minimizer $\lambda_t$ and that $A_t$ is finite in a neighborhood of $\lambda_t$. Under the conditions of Appendices~\ref{app:proof-iprojection}--\ref{app:proof-continuity-forcing}, the information projection exists, is unique at the law level, and, in a minimal observable representation with $C_t\succ0$, satisfies
\begin{align}
q_t(x)&=\widetilde q_t(x)\exp\!\left(\lambda_t^\top\phi(x)-A_t(\lambda_t)\right),\qquad \E_{Q_t}[\phi]=c_t, \label{eq:exponential-tilt}\\
C_t\dot\lambda_t&=\dot c_t-\E_{Q_t}[m_t]-\Cov_{Q_t}(\phi,\lambda_t^\top m_t), \label{eq:lambda-dot}\\
h_t&=\dot\lambda_t^\top(\phi-c_t)+\lambda_t^\top\!\left(m_t-\E_{Q_t}[m_t]\right),\qquad \E_{Q_t}[h_t]=0. \label{eq:computable-h}
\end{align}
\end{theorem}

Thus $\partial_tq_t+\nabla\!\cdot(q_tu_t)=q_th_t$, and the forcing can be evaluated from tilted expectations and an $R\times R$ covariance solve without differentiating the reference density in physical time.

\paragraph{Reference-relative expectation-rate robustness}
\label{subsec:theory-poisson}

For a sufficiently regular time-independent observable $f$, define
\begin{equation}
\mathcal E_t(f):=\frac{d}{dt}\E_{Q_t}[f]-\E_{Q_t}[\nabla f\cdot u_t]=\E_{Q_t}[h_tf]. \label{eq:weak-dynamical-discrepancy}
\end{equation}
Because $\E_{Q_t}[h_t]=0$, this functional is invariant to additive constants. Let $H_{q_t}$ denote the mean-zero weighted Sobolev energy space defined in Appendix~\ref{app:preliminaries}.

\begin{theorem}[Sharp expectation-rate characterization of instantaneous Full action]
\label{thm:poisson-correction}
Assume the weighted Dirichlet form is coercive on $H_{q_t}$ and that $f\mapsto\E_{Q_t}[h_tf]$ defines a bounded linear functional on $H_{q_t}$, under the boundary conventions of Appendix~\ref{app:preliminaries}. Then
\begin{align}
a_{\mathrm{full}}(t,\etavec)&=\sup_{\substack{f\in H_{q_t}\\ \E_{Q_t}[\|\nabla f\|^2]\le1}}|\mathcal E_t(f)|^2, \label{eq:full-weak-discrepancy}\\
\nabla\!\cdot(q_t\nabla\psi_t)&=q_th_t,\qquad \E_{Q_t}[\psi_t]=0,\qquad \delta_t^\star=-\nabla\psi_t, \label{eq:poisson-strong}\\
a_{\mathrm{full}}(t,\etavec)&=\E_{Q_t}[\|\nabla\psi_t\|^2]=\|h_t\|_{H^{-1}(q_t)}^2. \label{eq:poisson-action}
\end{align}
Here $\delta_t^\star$ is the unique minimum-energy correction in \eqref{eq:full-action-density}.
\end{theorem}

In this sense, Full asks not how far the completed law is from the reference at one time, but how difficult it is for the reference dynamics to realize its evolution. The $H^{-1}$ representation gives a useful transport interpretation of this cost. Full does not penalize continuity mismatch only by its pointwise amplitude; it weights that mismatch by the spatial effort required to correct it. Rapidly varying local discrepancies can often be repaired by moving mass over short distances and therefore have relatively small Full cost, whereas coherent low-frequency discrepancies requiring transport over longer distances are more expensive. Equivalently, in flux variables the energy weights a required correction flux by $1/q_t$, making transport of a fixed flux through poorly populated regions costly. This transport-scale sensitivity is precisely what distinguishes Full from a plain $L^2$ norm of the same continuity forcing; Appendix~\ref{app:proof-residual-spectral-comparison} gives exact Fourier examples.

The three equivalent views of Full play complementary roles. The minimum-correction form gives it a physical interpretation as transport effort, the $H^{-1}$ form exposes the spatial geometry of that effort, and the dual form gives the interpretation most directly relevant to experimental design. In particular, $\sqrt{a_{\mathrm{full}}(t,\etavec)}$ is the smallest uniform constant controlling reference-relative expectation-rate discrepancy per unit Sobolev gradient energy. Consequently, once scientific adequacy has been enforced, minimizing $\Act(\etavec)$ minimizes the time-integrated squared worst-case discrepancy over this smooth observable class rather than only over the quantities explicitly protected by the primary risk. Equivalent sharp-constant and operator formulations are given in Appendices~\ref{app:proof-hminusone} and \ref{app:comparison-operator} \citep{ambrosio2008gradientflows,mroueh2018sobolev,neklyudov2023action,chemseddine2025neural}.

Full transports the entire completed law, whereas the measured observables constrain only some of its expectation rates. This suggests separating the required correction into the part already forced by those measured rates and the additional law-level correction they leave unresolved.

\begin{proposition}[Full--Tangent--Hidden decomposition]
\label{prop:action-decomposition}
Assume the conditions of Theorem~\ref{thm:poisson-correction}, $J_\phi\in L^2(Q_t)$, $J_\phi u_t\in L^1(Q_t)$, and $\phi_j-c_{t,j}\in H_{q_t}$ for each observable coordinate, with moment differentiation and weak testing justified. Define
\begin{equation}
r_t=\E_{Q_t}[J_\phi u_t]-\dot c_t,\qquad G_t=\E_{Q_t}[J_\phi J_\phi^\top]. \label{eq:tangent-r-gram}
\end{equation}
The Tangent correction is $\delta_t^{\mathrm{tan}}=-J_\phi^\top G_t^\dagger r_t$, with $a_{\mathrm{tan}}(t,\etavec)=r_t^\top G_t^\dagger r_t$. Writing $\delta_t^{\mathrm{hid}}=\delta_t^\star-\delta_t^{\mathrm{tan}}$ and
$a_{\mathrm{hid}}(t,\etavec)=\E_{Q_t}[\|\delta_t^{\mathrm{hid}}\|^2]$, the Tangent and Hidden components are orthogonal and
\begin{equation}
a_{\mathrm{full}}(t,\etavec)
=
a_{\mathrm{tan}}(t,\etavec)
+
a_{\mathrm{hid}}(t,\etavec).
\label{eq:action-orthogonal-decomposition}
\end{equation}
Integrating against $\rho$ gives the corresponding design-level decomposition \eqref{eq:design-action-decomposition}, and hence $\Act_{\mathrm{tan}}(\etavec)\le\Act(\etavec)$.
\end{proposition}

Tangent is therefore the restriction of the Full dual problem to the finite-dimensional span of the centered measured observables, equivalently a Galerkin restriction in the weighted Sobolev energy space, while Hidden is the corresponding Galerkin energy gap \citep{mroueh2018sobolev}. For experimental design, this means that two sensor systems can be equally adequate for the scientific task while inducing very different dynamical corrections elsewhere in the completed law. Full distinguishes precisely this residual difference. The exact Tangent--Full equality condition and examples showing that Tangent does not generally control Full are given in Appendices~\ref{app:proof-tangent-full-equality} and \ref{app:proof-tangent-no-control}.

\paragraph{Criterion distinctions and matched-initial stability.}
\label{subsec:theory-comparisons}
The remaining secondary criteria probe different properties of the same information-projected law path. Residual $L^2$ measures the amplitude of the complete normalized continuity forcing, whereas Full weights that forcing through inverse transport geometry. Forward KL and static $W_2$ instead measure marginal displacement from the frozen reference and do not encode the temporal realization cost of the prescribed path. Appendix~\ref{app:analytical-comparisons} establishes these distinctions through operator identities, exact information-projected separations, and ranking reversals.

Full also yields a positive stability statement in a stronger regular-flow regime. Let $\mathsf T^u_{s,t}$ denote the characteristic flow of $u_t$ and define the matched-initial reference rollout $R_t^{\etavec}:=(\mathsf T^u_{0,t})_\#Q_0^{\etavec}$. Assuming finite second moments and Full action, classical characteristic flows for $u_t$ and $u_t+\delta_t^\star$, and an integrable spatial Lipschitz bound $\|u_t(x)-u_t(y)\|\le L(t)\|x-y\|$, Proposition~\ref{prop:matched-initial-wasserstein-control} gives, for $\rho(dt)=dt$ and $\Lambda=\int_0^1L(t)\,dt$,
\begin{equation}
W_2(Q_t^{\etavec},R_t^{\etavec})\le\int_0^t\exp\!\left(\int_s^tL(\tau)\,d\tau\right)\sqrt{a_{\mathrm{full}}(s,\etavec)}\,ds\le e^\Lambda\sqrt{\Act(\etavec)},\qquad t\in[0,1].
\label{eq:matched-initial-wasserstein-main}
\end{equation}
Hence every globally $1$-Lipschitz observable $f$ satisfies
\[
\left|\E_{Q_t^{\etavec}}[f]-\E_{R_t^{\etavec}}[f]\right|\le e^\Lambda\sqrt{\Act(\etavec)}.
\]
When $Q_0^{\etavec}=\widetilde Q_0$, one has $R_t^{\etavec}=\widetilde Q_t$ and $\Act_{W_2}(\etavec)\le e^{2\Lambda}\Act(\etavec)$. More generally, the preceding bound gives a downstream-agnostic stability interpretation: after initial-law mismatch is removed, small Full action controls the departure of every later-chosen globally $1$-Lipschitz observable from the corresponding frozen-reference rollout, even if that observable was not part of the scientific risk used to select the design. This is a one-sided reference-relative guarantee, not a guarantee of recovery of the unknown intermediate population law. Appendix~\ref{app:proof-static-wasserstein} shows that no reverse bound holds in general.
Proofs, technical conditions, and additional analytical results are collected in Appendices~\ref{app:theory-proofs} and \ref{app:analytical-comparisons}.

\section{Experiments}
\label{sec:experiments}

The experiments follow a progression from mechanism to deployment. Gaussian-mixture transport first asks whether the choice of secondary criterion actually changes the measurement design once scientific adequacy is fixed. A controlled downstream experiment then asks when the baseline-preserving preference improves prediction of a quantity that was never used for selection and when that benefit fails. Prospective double-gyre transport removes access to intermediate microscopic target states during design, and interacting skyrmions finally test the same construction on a higher-dimensional joint law. Across these studies we evaluate three distinct quantities when available: held-out task risk tests whether the scientific requirement transfers, held-out Full tests whether the selected minimum-revision preference persists on fresh data, and truth-relative downstream diagnostics test consequences outside the selection objective. Section~\ref{subsec:additional-results} collects supporting analyses of the modeling assumptions and numerical approximations behind these results. The source code is available on GitHub.\footnote{\url{https://github.com/REPOSITORY-TO-BE-RELEASED}}

\paragraph{(1) Secondary criteria induce different designs: Gaussian-mixture transport}
\label{subsec:exp-gaussian}

The Gaussian benchmark isolates the effect of the secondary criterion in a transparent two-dimensional setting: the endpoint laws are fixed, the orientation of the intermediate mixture varies, and two localized Gaussian sensors are placed on a ring. Law, Tangent, residual $L^2$, forward KL, static $W_2$, and Full select different sensor geometries despite closely matched scientific risk. Table~\ref{tab:exp-gaussian-criteria} cross-evaluates these designs using reference-relative Full action and truth-relative hidden-law discrepancy. The orderings disagree: the Full-selected geometry has the smallest held-out Full action but the largest hidden-law discrepancy. Thus the secondary criteria resolve the same scientifically admissible design freedom differently rather than serving as interchangeable approximations to a common objective. This disagreement motivates the next question: if minimum revision is not a generic truth-recovery principle, under what conditions does preserving unresolved baseline structure nevertheless improve a quantity that was never used to select the design? Complete criterion comparisons and allowance sweeps are reported in Appendices~\ref{app:gaussian-common-candidate-comparison} and~\ref{app:exp-analytic}.

\begin{table}[t]
\centering
\small
\begin{tabular}{lrrr}
\toprule
Selector & Selection risk increase (\%) & Held-out Full (cross-eval.) & Truth-relative hidden-law $\mathrm{SW}_2^2$ \\
\midrule
Law & $0$ & $29.4178$ & $0.096460$ \\
Tangent & $0.491$ & $37.3317$ & $0.094144$ \\
Residual $L^2$ & $0.441$ & $36.0548$ & $0.094173$ \\
Forward KL & $0.303$ & $33.0647$ & $0.094352$ \\
Static $W_2$ & $0.294$ & $32.9924$ & $0.094357$ \\
Full & $0.604$ & $19.5359$ & $0.102205$ \\
\bottomrule
\end{tabular}
\caption{Criterion-specific Gaussian designs at the $1\%$ allowance, selected under the same scientific-risk restriction and cross-evaluated using held-out Full action and truth-relative hidden-law squared sliced-Wasserstein distance. The two diagnostics induce different orderings; values are comparable within columns but not across columns.}
\label{tab:exp-gaussian-criteria}
\end{table}

\paragraph{(2) When does baseline preservation help and when does it hurt?}
\label{subsec:exp-baseline-scope}

We next ask when the baseline-preserving preference has a measurable consequence for a quantity never used in design. We construct a product-law problem with one task-relevant coordinate and one task-unprotected coordinate. Each candidate sensor contains the same task-relevant signal but mixes it with a different component of the unprotected coordinate, so the primary scientific risk is exactly tied across five sensor geometries. The frozen baseline is initially correct in the unprotected coordinate. As an independent downstream test, we measure truth-relative local-occupancy RMSE on $16$ smooth observables of that coordinate, none of which enters either the scientific risk or the secondary design criteria. Because Law is indifferent among all five risk-optimal geometries, we report its expected downstream error under uniform tie-breaking. When the baseline is correct, Full and static $W_2$ select the same geometry, reducing expected occupancy error by about half relative to Law, whereas the geometry selected by Tangent, residual $L^2$, and forward KL performs worse.

\begin{figure}[t]
\centering
\includegraphics[width=\columnwidth]{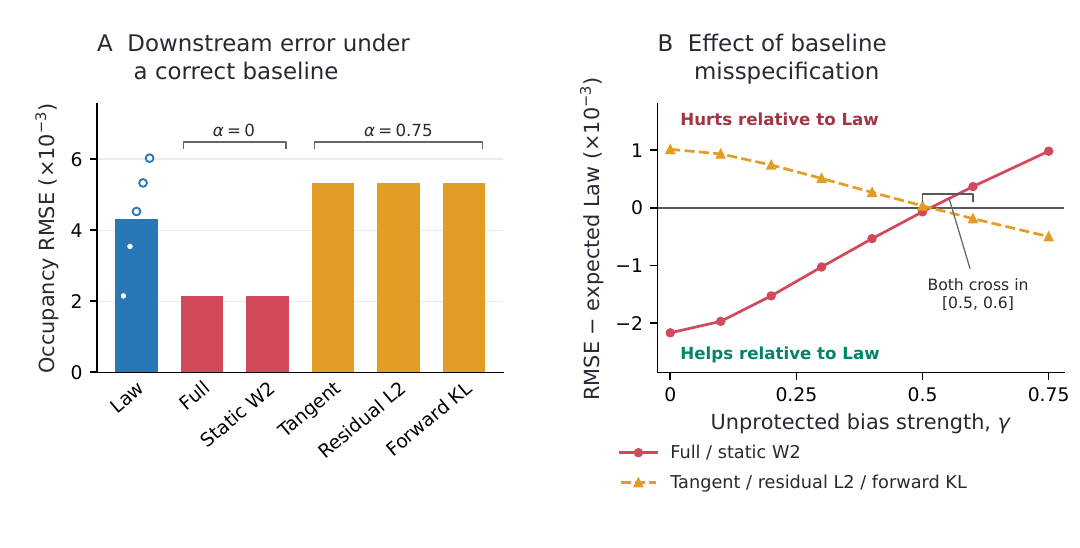}
\caption{\textbf{Downstream consequences of baseline-preserving selection.} \textbf{A:} At signal amplitude $\epsilon=0.10$ and occupancy localization scale $\kappa=4$, we evaluate truth-relative local-occupancy RMSE on $16$ smooth observables that are withheld from both the scientific risk and all secondary design criteria. Scientific risk is identical across the five candidate geometries. Full and static $W_2$ select $\alpha=0$, while Tangent, residual $L^2$, and forward KL select $\alpha=0.75$. The Law bar is the expected RMSE under uniform tie-breaking over its five risk-minimizing geometries; open circles show the corresponding geometry-specific errors. \textbf{B:} The selected geometries are then frozen while a true departure of strength $\gamma$ is introduced in the task-unprotected coordinate. The ordinate is occupancy RMSE minus the expected Law RMSE, so negative values indicate improvement over Law and positive values indicate harm. The Full/static-$W_2$ geometry changes from lower to higher error than Law as misspecification grows, while the Tangent/residual-$L^2$/forward-KL geometry reverses in the opposite direction; both crossings occur between $\gamma=0.5$ and $0.6$ in this representative setting.}
\label{fig:fide-comparator-misspecification}
\end{figure}

We then freeze these selections and the reference and introduce an increasingly strong true departure in the unprotected coordinate. Figure~\ref{fig:fide-comparator-misspecification} shows the resulting transition: the advantage of the Full/$W_2$ geometry steadily erodes and eventually reverses relative to Law. This isolates when the baseline-preserving inductive bias is useful: preserving unresolved baseline structure improves prediction of downstream quantities withheld from design when that structure is reliable, but becomes harmful once the baseline is sufficiently wrong. In this construction static $W_2$ makes the same selection, so the result identifies the value of baseline preservation rather than an advantage unique to Full. This controlled setting lets us vary baseline reliability while retaining access to ground truth; we next remove that luxury and ask whether the residual design choice can be made prospectively before intermediate microscopic target states are available. Appendix~\ref{app:baseline-correctness-ablation} gives the complete construction, robustness checks, and full misspecification sweep.

\paragraph{(3) Prospective selection without intermediate target states: double-gyre transport} The double-gyre provides a nonautonomous transport benchmark with stretching, exchange, and nontrivial passive-tracer motion \citep{shadden2005definition}. We choose four localized sensors for a heterogeneous tracer population while withholding all intermediate microscopic target states from the optimizer. Selection uses only endpoint ensembles, a frozen endpoint-trained reference, and a frozen aggregate predictive interface constructed from a separate simulation bank, which supplies the candidate moment trajectories and summaries needed to evaluate scientific risk. After the sensor geometries are frozen, we introduce a separately seeded validation reference and fresh microscopic validation trials. On these held-out data, the $2\%$--$5\%$ Full-selected designs satisfy their corresponding scientific-risk caps, while the $1\%$ point remains unresolved; across the tested frontier, the Full-selected geometries also retain lower Full action than Law. Thus both scientific adequacy, where statistically resolved, and the minimum-revision ordering transfer after design freeze without access to intermediate target states during selection. Complete results and uncertainty analyses are in Appendix~\ref{app:exp-vortices-prospective}, and Figure~\ref{fig:experimental-overview} summarizes the resulting risk--action tradeoff.

\begin{figure}[t]
    \centering
    \begin{minipage}[c]{0.50\linewidth}
        \centering
        \includegraphics[width=\linewidth]{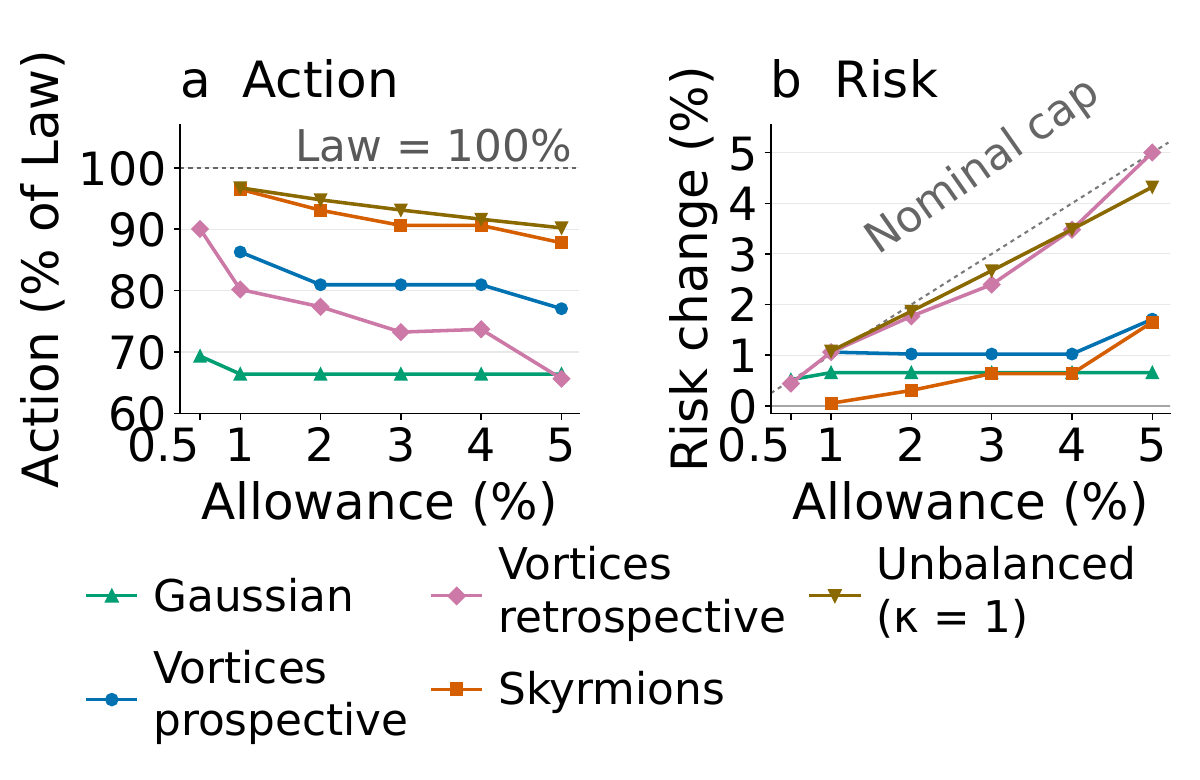}
    \end{minipage}
    \hfill
    \begin{minipage}[c]{0.49\linewidth}
        \centering
        \includegraphics[width=\linewidth]{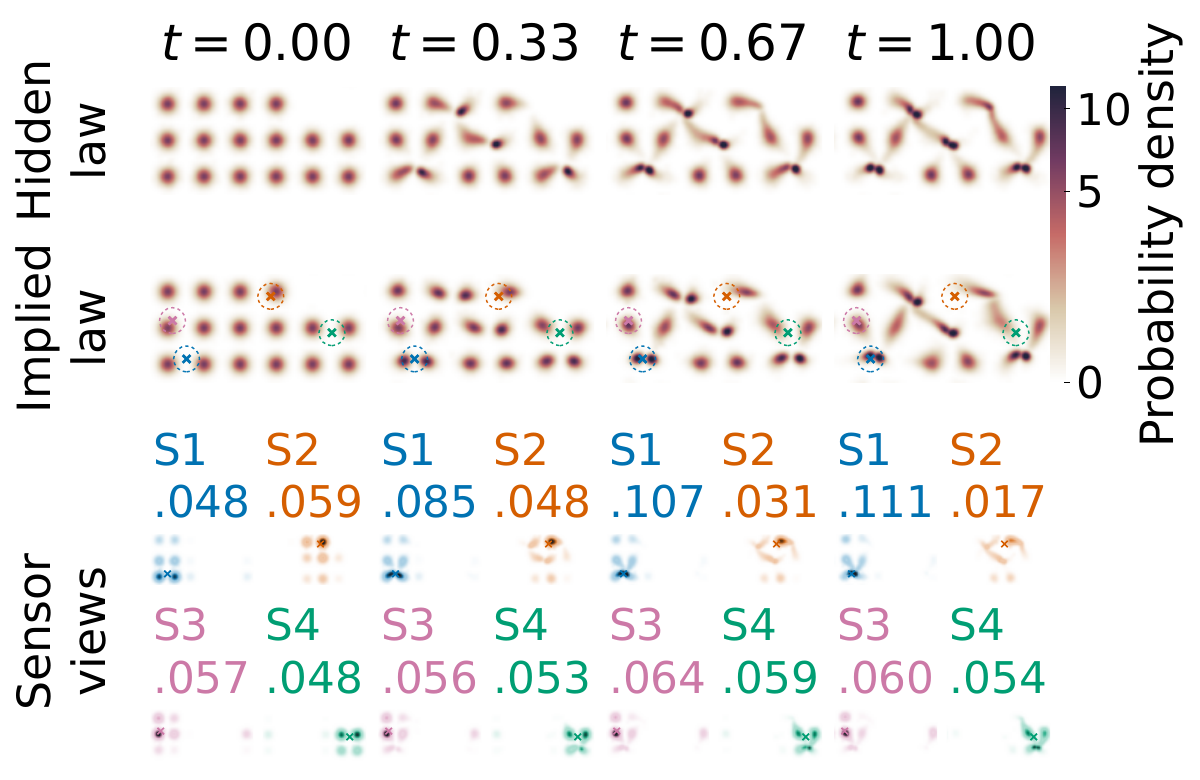}
    \end{minipage}
    \caption{\textbf{Empirical overview.} \emph{Left:} Held-out scientific risk and Full action across the tested risk allowances in the prospective double-gyre study. \emph{Right:} Interacting-skyrmion setup, where aggregate sensor views constrain a $32$-dimensional configuration law without determining it.}
    \label{fig:experimental-overview}
\end{figure}

\paragraph{(4) Configuration-space design in $32$ dimensions: interacting skyrmions} The prospective study changes the information available at design time; our final experiment instead changes the computational regime. Motivated by interacting magnetic skyrmions \citep{nagaosa2013topological,reichhardt2015collective}, we use a reduced stochastic model of $16$ interacting planar particles, giving a $32$-dimensional microscopic state. Four aggregate sensors observe averages over $512$ configurations, while scientific risk is defined through nine permutation-invariant features of pair structure and collective order. Full is approximated directly in configuration space using a fixed $K=280$ Galerkin potential dictionary, so the reported values assess the finite-dimensional construction rather than continuum convergence. On fresh paired validation trials, the Full-selected designs satisfy their scientific-risk caps, differ from the corresponding Law selections, and retain lower fitted action. The experiment therefore shows that the same residual-design construction remains operational on a $32$-dimensional joint law; complete results and numerical audits are in Appendix~\ref{app:exp-skyrmions}.

\subsection{Supporting analyses}
\label{subsec:additional-results}
Supporting studies examine the assumptions and numerical approximations behind the main experiments. Baseline-correctness and reference-perturbation analyses characterize when the minimum-revision preference helps or fails under misspecification (Appendices~\ref{app:baseline-correctness-ablation} and~\ref{app:gaussian-reference-misspecification}). Allowing an unresolved mode to adapt dynamically reduces Full action by roughly $10$--$26\%$, quantifying the effect of fixing completion before scoring, while retrospective double-gyre and prospective predictor perturbations assess repeatability and sensitivity to the aggregate interface (Appendices~\ref{app:path-completion-ablation}, \ref{app:exp-vortices}, and~\ref{app:prospective-predictor-sensitivity}). Numerical checks further test raster, smoothing, and solver choices, manufactured-problem convergence, richer skyrmion potential spaces, and leave-one-out stability (Appendices~\ref{app:prospective-numerical-stability}, \ref{app:full-manufactured-convergence}, and~\ref{app:skyrmion_ablations}). Together, these analyses distinguish sensitivity to the modeling assumptions from sensitivity to the numerical implementation.

\section{Conclusion}
\label{sec:conclusion}

FIDE separates two questions in experimental design: what must a measurement system accomplish for the scientific task, and how should the remaining law-level freedom be resolved once that task is satisfied? Scientific risk answers the first; FIDE addresses the second by completing each design's aggregate constraints relative to a common frozen reference and measuring the dynamical revision required to realize the resulting path. Full action makes this revision precise through equivalent minimum-correction and expectation-rate interpretations, while the Full--Tangent--Hidden decomposition identifies the part of the required correction that is invisible to the measured moment rates.

The experiments show that this residual choice matters. Different secondary criteria select different scientifically admissible measurements; preserving unresolved baseline structure can improve prediction of downstream quantities withheld from design when that structure is reliable and can hurt when it is sufficiently misspecified; prospective selection can transfer after design freeze without intermediate microscopic target states; and the finite-dimensional construction can be instantiated on a $32$-dimensional joint law. Minimum Full is therefore most naturally viewed as a design principle for settings in which preserving a declared baseline outside task-valued departures is itself part of the experimental specification. Extending FIDE to propagate uncertainty over that baseline is a natural next step. A complementary direction is scalable optimization with discrete or combinatorial experimental decisions, where smooth surrogate objectives or continuous relaxations  could retain gradient-based optimization while ultimately selecting feasible discrete designs.

\section*{AI Use Statement}
Generative AI tools were used to assist with code development and debugging and with language-level polishing of the manuscript. The authors reviewed and validated the resulting code and manuscript text, and take full responsibility for the final submission.

\section*{Reproducibility Statement}
The code required to reproduce all experiments will be made publicly available on GitHub upon publication. Detailed experimental protocols, information boundaries, numerical procedures, implementation details, hardware and software specifications, and benchmark-specific settings are provided in Appendix~\ref{app:experiments}. The appendix also reports numerical audits, held-out validation procedures, and sensitivity analyses used to assess reproducibility and stability of the reported results.

\bibliography{fide}
\bibliographystyle{iclr2027_conference}

\appendix

\section{Related Work}
\label{app:related-work}

FIDE sits at the intersection of experimental design under partial information, statistical completion of aggregate constraints, and dynamical transport relative to a reference model. The individual ingredients have substantial histories; the distinction here is how they are assembled at the experimental-design level. A candidate measurement system determines a finite set of observable constraints and therefore a design-dependent family of compatible laws. FIDE first completes that ambiguity relative to a common frozen reference, then asks how much the resulting completed dynamics must depart from the reference, but only after the declared scientific objective has determined which designs are acceptable.

\paragraph{Experimental design and sensor placement}
Bayesian optimal experimental design provides a decision-theoretic framework for selecting measurements by their expected utility \citep{chaloner1995bayesian}. Modern developments include variational estimators of expected information gain \citep{foster2019variational} and amortized sequential policies \citep{foster2021deepadaptive}. Goal-oriented design instead concentrates measurement effort on prescribed quantities of interest \citep{attia2018goal}, while control-oriented formulations target uncertainty that matters for subsequent control decisions \citep{madhavan2026control}. Closely related sensor-placement objectives include mutual information \citep{krause2008near}, posterior covariance \citep{alexanderian2014aoptimal}, and data-driven reconstruction quality \citep{manohar2018data}. Continuous sensor locations can also be learned from reconstruction feedback \citep{ma2025physense}, and robust OED explicitly accounts for misspecification in priors or measurement uncertainty \citep{attia2025robust}. In transport settings, placement has been studied through observability and controllability Gramians for linear advection \citep{vaidya2012actuator} and jointly with state and parameter estimation for stochastic advection--diffusion systems \citep{sharrock2022joint}.

Recent work has also used transport geometry directly in the outer design optimization. Wasserstein gradient flows can optimize probability measures over experimental designs for classical linear-model objectives \citep{jin2026optimal} and for batch expected information gain \citep{sharrock2026wasserstein}, while alternative BOED objectives replace KL-based information gain by integral probability metrics such as Wasserstein distance, maximum mean discrepancy, and energy distance \citep{wu2026beyond}. These approaches differ in the utility they optimize, but they share the usual objective of making the measurement system itself maximally informative, predictive, or useful for a specified downstream task. A particularly relevant contrast is optimal model-discrimination design, where disagreement among candidate models is itself informative and experiments are chosen to distinguish rival mechanisms or reduce uncertainty about model identity \citep{box1967discrimination,atkinson1975design,hainy2022optimal}. FIDE addresses the complementary baseline-preserving regime: the scientific risk first determines an admissible set of comparably adequate designs, and residual disagreement with the frozen reference is not assigned additional scientific utility unless it enters that risk. Only within this declared scope does FIDE use a second criterion to resolve residual law-level freedom by minimum dynamical revision. The resulting $\varepsilon$-constraint ordering is classical in multiobjective optimization \citep{haimes1971bicriterion}, the contribution here is the reference-relative dynamical criterion used to compare the admissible designs.

\paragraph{Risk-informed model credibility}
A complementary literature treats model credibility as conditional on a specified context of use rather than as a global property of a model. Risk-informed frameworks make the required credibility evidence commensurate with the consequences of relying on a computational model for that use, and organize verification, validation, and uncertainty assessment around the question the model is intended to support \citep{morrison2019assessing,viceconti2021silico}. This literature motivates the scoped-use setting considered here but does not prescribe FIDE's secondary criterion: baseline preservation is an additional design specification, and FIDE provides a dynamical rule for resolving the residual measurement-design freedom left by the primary task.

\paragraph{Aggregate information and statistical completion}
A separate literature asks what can be inferred when observations constrain only aggregate statistics rather than microscopic states.
Maximum entropy and information projection provide classical ways to complete such partial statistical information \citep{kullback1951information,jaynes1957information,csiszar1975idivergence}. Related work develops geometric views of entropy optimization under linear constraints \citep{pavon2013geometry} and generative procedures for sampling moment-constrained maximum-entropy laws \citep{lempereur2026mgd}. Particularly close in sensor-design motivation, \citet{bhattacharya2026sensor} constructs a posterior and induced likelihood by minimizing KL divergence from a dynamical prior subject to distributional-accuracy constraints, including moment constraints. FIDE also uses KL projection relative to a dynamical baseline, but the role of the projection is different: for each candidate observable map, it provides a canonical completion of that design's moment fiber before the design is scored dynamically.

There are also dynamic entropy- and projection-based approaches to incomplete information. Maximum Caliber extends entropy principles to trajectory distributions \citep{dixit2018maximumcaliber}, and Schr\"odinger-bridge formulations can incorporate distributional moments and average currents \citep{miangolarra2024maximumcaliber}. Projection filters approximate nonlinear filtering dynamics on finite-dimensional exponential families \citep{brigo1999projection}, with related Fokker--Planck approximations connecting tangent projection, moment matching, and Galerkin methods \citep{brigo2017optimal}. Best-fit statistical closure takes another route, choosing paths in quasi-equilibrium families by minimizing a time-integrated weighted squared Liouville residual with a resolved/unresolved decomposition \citep{turkington2013optimization}. FIDE does not assume that the measured moments close the dynamics or identify the underlying law. Instead, the observable map itself is a design variable, so different candidate designs induce different moment fibers and different reference-relative completed paths. The completion is fixed before its dynamical correction cost is evaluated.

\paragraph{Reference-relative probability dynamics}
The frozen reference used by FIDE is closely related to modern learned probability-flow models. Flow matching and stochastic interpolants learn probability paths together with velocity fields \citep{lipman2023flow,albergo2023building,albergo2025stochastic}, and optimal-transport couplings can be used to simplify the learned flows \citep{tong2024improving}. In our experiments, endpoint-trained flow matching is one way to instantiate the frozen pair $(\widetilde Q_t,u_t)$. The theoretical construction itself is more general: it requires only a reference law path and velocity satisfying the continuity equation, and neither is assumed to coincide with the unknown intermediate population dynamics.

Other approaches modify or infer probability dynamics relative to prescribed constraints or prior processes. Schr\"odinger bridges construct path laws close to a reference stochastic process under marginal constraints \citep{debortoli2021diffusion}; TrajectoryNet infers continuous dynamics from cross-sectional populations \citep{tong2020trajectorynet}; and constrained generative methods impose distributional \citep{khalafi2024constrained}, average \citep{hadou2026primaldual}, or exact structural constraints through constrained flow geometry \citep{cheng2025stiefel}. Dynamic optimal transport supplies the continuity-constrained kinetic-energy geometry underlying many such constructions \citep{benamou2000computational,ambrosio2008gradientflows}, with related formulations in minimum-energy distribution steering \citep{chen2016steering}, optimal transport with prior dynamics \citep{chen2017optimaltransport}, and dynamic-transport regularization of measure-valued reconstruction from indirect time-dependent observations \citep{bredies2020dynamic}.

Measure-based nudging provides another related perspective. \citet{lyu2026multiscale} map macroscopic observations to microscopic mean-field particle corrections through Wasserstein-gradient feedback, while \citet{lyu2026wasserstein} construct Wasserstein feedback from observed hydrodynamic moments for Vlasov--Poisson dynamics, where the moments leave the underlying phase-space law underdetermined. These methods seek corrections for a specified observation mechanism. In FIDE, the observation mechanism itself varies across candidate experimental designs, and the question is which scientifically adequate design induces the least reference-relative dynamical revision after its aggregate information has been canonically completed. More general Wasserstein Lagrangian formulations optimize density paths through combinations of kinetic and potential energies \citep{neklyudov2024wasserstein}; by contrast, FIDE does not reoptimize the projected path to reduce its action.

\paragraph{Minimum-energy realization and Sobolev geometry}
The dynamical criterion used by FIDE is most directly connected to work on minimum-energy realization of prescribed law paths. Action Matching learns minimum-kinetic-energy gradient velocities from temporally unpaired samples of a given marginal evolution \citep{neklyudov2023action}. For Boltzmann density curves, \citet{chemseddine2025neural} establish weighted-Poisson conditions for Wasserstein absolute continuity and minimum-norm realization and analyze velocity blow-up for particular interpolations. Related weighted-Poisson constructions connect variational Bayesian updating to feedback particle filtering \citep{laugesen2015poisson}, transport samples along geometric reference--target density interpolations in kernel Fisher--Rao flow \citep{maurais2024sampling}, and jointly optimize tilted density paths and their transport dynamics to obtain smoother velocities \citep{maurais2025learningpaths}. FIDE uses the same general realization geometry in a different outer problem: the moment-constrained path is determined by the measurement design and the common frozen reference before its realization cost is computed. Neither that path nor the reference velocity is subsequently optimized to make the action small.

The dual characterization of this correction is also related to Sobolev discrepancies. Finite-feature Sobolev discrepancy uses derivative Gram matrices and identifies the finite-feature approximation gap as an energy-norm error \citep{mroueh2018sobolev}, while Sobolev Descent uses closely related geometry for particle transport \citep{mroueh2019sobolev}. In FIDE, restricting the full expectation-rate discrepancy to the span of the measured observables produces the Tangent criterion, and the remaining energy defines the Hidden component. Wasserstein natural gradients and Wasserstein information matrices likewise pull transport geometry back to parametrized statistical families \citep{chenli2020natural,li2023wasserstein}, which is closely related to the inverse-transport Gram structure that appears in our comparisons.

Taken together, these literatures provide the main ingredients used by FIDE: information projection to complete finite statistical constraints, reference-relative transport geometry to quantify the effort required to realize a prescribed law path, and Sobolev duality to interpret that effort through observable expectation rates. Our contribution is not a new information-projection principle or a new underlying transport metric. It is to place these ingredients inside a residual experimental-design problem in which each candidate measurement map induces its own moment fiber and completed law path, scientific risk determines admissibility, and the remaining design freedom is resolved by the dynamical revision required relative to a common frozen reference.

\section{Technical statements and proofs}
\label{app:theory-proofs}

This appendix develops the technical results underlying Section~\ref{sec:theory}. We first state the notation, weak-form conventions, and functional-analytic preliminaries used throughout the proofs. We then give the assumptions and proofs for the information-projected lift, physical-time multiplier dynamics, density-free continuity forcing, the minimum-energy and weighted-$H^{-1}$ characterizations of Full, the sharp uniform certificate, Tangent, and the Full--Tangent--Hidden decomposition and equality condition. We also establish finite-reference information-projection feasibility, calibration sensitivity, and end-to-end differentiability of the fixed finite-dimensional Full objective. The finite-measure transport--reaction extension is developed alongside the balanced theory and evaluated separately in Appendix~\ref{app:unbalanced-experiment}. Operator-level comparisons and exact information-projected examples separating Full from Tangent, residual $L^2$, forward KL, and static Wasserstein displacement, together with the matched-initial stability result, are collected in Appendix~\ref{app:analytical-comparisons}.

\subsection{Preliminaries and standing conventions}
\label{app:preliminaries}

\paragraph{State space, laws, and boundary conventions.}
Whenever spatial derivatives are used, we identify the state space with a domain $\Omega\subseteq\R^d$, a flat torus, or $\R^d$. Bounded-domain statements use boundary conditions compatible with the corresponding continuity equation and variational problem; throughout the balanced theory these are periodic, sufficient-decay, or no-flux conditions chosen so that the stated integration-by-parts identities have no boundary contribution. Additional geometric hypotheses, such as connectedness or strict positivity of a density, are imposed only where explicitly required. Probability laws are denoted by capital letters and, when absolutely continuous with respect to Lebesgue measure, their densities by the corresponding lower-case letters. In particular, $\widetilde Q_t$ and $\widetilde q_t$ denote the frozen reference law and density, while $Q_t^{\etavec}$ and $q_t^{\etavec}$ denote the information-projected law and density associated with design $\etavec$. Expressions involving division by $q_t^{\etavec}$ are understood on its positive support. Expectations are written $\E_Q[\cdot]$ or $\E_q[\cdot]$ interchangeably when $q$ is the density of $Q$.

\paragraph{Weak continuity equations.}
Unless the displayed densities and vector fields are regular enough to justify a strong formulation, all continuity equations are understood in the distributional sense. Specifically, for
\begin{equation}
\partial_t q_t+\nabla\!\cdot(q_tv_t)=s_t,
\label{eq:appendix-general-continuity}
\end{equation}
the weak formulation is
\begin{equation}
\int_0^1\!\int_{\Omega}\left(q_t\,\partial_t\zeta+q_tv_t\cdot\nabla\zeta+s_t\zeta\right)\,dx\,dt=0
\label{eq:appendix-general-continuity-weak}
\end{equation}
for every admissible smooth test function $\zeta$ with compact support in physical time and satisfying the corresponding spatial boundary convention. Whenever the map $t\mapsto\int f q_t\,dx$ is absolutely continuous, this implies
\begin{equation}
\frac{d}{dt}\E_{q_t}[f]=\E_{q_t}[\nabla f\cdot v_t]+\int_{\Omega}f(x)s_t(x)\,dx
\label{eq:appendix-expectation-rate-general}
\end{equation}
for admissible time-independent scalar observables $f$. In the conservative case $s_t=0$, applying this identity coordinatewise to a sufficiently regular time-independent observable vector $\Phivec_{\etavec}:\Omega\to\R^R$ gives
\begin{equation}
\frac{d}{dt}\E_{q_t}[\Phivec_{\etavec}]=\E_{q_t}[J_{\Phivec_{\etavec}}v_t].
\label{eq:moment-rate-identity}
\end{equation}

\paragraph{Weighted $L^2$ geometry.}
For scalar or vector fields $f$ and $g$ with the required square integrability, write
\begin{equation}
\langle f,g\rangle_{L^2(q)}:=\E_q[f^\top g],\qquad \|f\|_{L^2(q)}^2:=\E_q[\|f\|^2].
\label{eq:appendix-weighted-l2}
\end{equation}
For an admissible scalar function $\varphi$, define the weighted gradient-energy seminorm
\begin{equation}
\|\varphi\|_{\dot H^1(q)}^2:=\E_q[\|\nabla\varphi\|^2].
\label{eq:appendix-weighted-h1-seminorm}
\end{equation}
Whenever the weighted Dirichlet form is coercive modulo constants, as assumed explicitly in Appendix~\ref{app:proof-poisson}, we use the mean-zero representative $\E_q[\varphi]=0$ and denote by $H_q$ the corresponding completed Hilbert energy space, equipped with the norm $\|\varphi\|_{H_q}:=\|\varphi\|_{\dot H^1(q)}$. Thus the mean-zero condition fixes the constant gauge in the balanced Poisson problem; it is not a global assumption on every function appearing in the appendix. The screened space $H_{q,\kappa}$ used in the finite-measure extension is defined separately in Corollary~\ref{cor:ub-full-correction}, where its zeroth-order term removes the constant null direction.

\paragraph{Observable and Jacobian conventions.}
For a design $\etavec$, the observable map is $\Phivec_{\etavec}:\Omega\to\R^R$. Its Jacobian is taken in the row convention
\begin{equation}
J_{\Phivec_{\etavec}}(x)\in\R^{R\times d},
\label{eq:appendix-jacobian-convention}
\end{equation}
so that $J_{\Phivec_{\etavec}}v\in\R^R$ for $v\in\R^d$ and $J_{\Phivec_{\etavec}}^\top a\in\R^d$ for $a\in\R^R$. Covariance matrices are
\begin{equation}
\Cov_Q(Y,Z):=\E_Q\!\left[(Y-\E_Q[Y])(Z-\E_Q[Z])^\top\right].
\label{eq:appendix-covariance-convention}
\end{equation}

\paragraph{Linear-algebra conventions.}
For a matrix $M$, $\Range(M)$ denotes its column space, $\ker(M)$ its nullspace, and $M^\dagger$ its Moore--Penrose pseudoinverse. All covariance and Gram matrices appearing below are symmetric positive semidefinite. Positive definiteness is imposed only when invertibility is required. In particular, positive definiteness of the projected covariance is needed for local multiplier identifiability and implicit differentiation, but not for uniqueness of the information-projected law itself.

\paragraph{Time integration and equality conventions.}
The design-level actions are integrated against the same fixed probability measure $\rho$ on $[0,1]$, unless a later comparison explicitly specializes to $\rho(dt)=dt$. Equalities between densities, scalar potentials, and vector fields are understood almost everywhere with respect to the relevant base measure. Integrated identities involving nonnegative action densities are valid in the extended-valued sense unless a finite-action assumption is explicitly needed; in particular, conclusions obtained by subtracting two integrated actions are invoked only when the corresponding quantities are finite.

\paragraph{Compact notation.}
For compactness, when $(t,\etavec)$ is fixed we write
\begin{equation}
\widetilde q=\widetilde q_t,\qquad q=q_t^{\etavec},\qquad \phi=\Phivec_{\etavec},\qquad c=\cvec_{\etavec}(t),\qquad \lambda=\lambdavec_{t,\etavec},\qquad u=u_t.
\label{eq:appendix-shorthand}
\end{equation}
Expectations without an explicit subscript in a proof are taken under the law indicated by the surrounding expression. The conventions above are deliberately weaker than the hypotheses of individual results: existence of an information projection, differentiability of the calibrated multiplier, coercivity of the weighted Poisson problem, positivity of the projected density, and rank conditions are stated locally where they are used rather than imposed globally.

\subsection{Finite-measure transport--reaction setup}
\label{app:technical-setup}

The finite-measure results below require additional notation beyond the balanced probability-law construction. We write a finite positive measure as
\begin{equation}
\mu_t(dx)=M_tq_t(x)\,dx,\qquad M_t>0,\qquad \int q_t(x)\,dx=1,
\label{eq:ub-factorization}
\end{equation}
where \(M_t\) is the total mass and \(q_t\) is its normalized shape. The normalized shape is information-projected relative to the frozen normalized reference shape \(\widetilde q_t\), while the mass is treated as a separate scalar channel. We assume that the frozen reference shape satisfies
\begin{equation}
\partial_t\widetilde q_t+\nabla\!\cdot(\widetilde q_tu_t)=0,
\label{eq:ub-reference-shape}
\end{equation}
and that the reference mass evolution has a spatially constant logarithmic source rate \(g_{\mathrm{ref}}(t)\). For a time-independent observable vector \(\phi\), define the raw and normalized moments by
\begin{equation}
\bar c_t:=\int\phi\,d\mu_t=M_t\E_{q_t}[\phi],\qquad c_t:=\frac{\bar c_t}{M_t}=\E_{q_t}[\phi].
\label{eq:ub-raw-shape-moments}
\end{equation}

A corrected finite-measure evolution is taken to satisfy
\begin{equation}
\partial_t(M_tq_t)+\nabla\!\cdot\!\left(M_tq_t(u_t+\delta_t)\right)=M_tq_t\left(g_{\mathrm{ref}}(t)+\alpha_t\right),
\label{eq:ub-corrected-continuity}
\end{equation}
where \(\delta_t\) is the transport correction and \(\alpha_t\) is the reaction-rate correction. Define
\begin{equation}
h_{\mathrm{shape}}:=\frac{\partial_tq+\nabla\!\cdot(qu)}{q},\qquad h_{\mathrm{ub}}:=h_{\mathrm{shape}}+\frac{\dot M}{M}-g_{\mathrm{ref}}.
\label{eq:ub-residual}
\end{equation}
Equation~\eqref{eq:ub-corrected-continuity} is then equivalent to
\begin{equation}
-\nabla\!\cdot(q\delta)+q\alpha=qh_{\mathrm{ub}}.
\label{eq:ub-correction-constraint}
\end{equation}
For a fixed reaction penalty \(\kappa>0\), define the instantaneous unbalanced Full action by
\begin{equation}
a_{\mathrm{full}}^{\mathrm{ub}}(t,\etavec):=M_t\inf_{\delta,\alpha}\left\{\E_{q_t}\!\left[\|\delta\|^2+\kappa\alpha^2\right]:-\nabla\!\cdot(q_t\delta)+q_t\alpha=q_th_{\mathrm{ub}}\right\}.
\label{eq:ub-full-action-definition}
\end{equation}
The separation of this forcing and action into normalized-shape and total-mass components is established in Corollary~\ref{cor:ub-mass-shape-decomposition}.

\begin{remark}[Mass is separate from the normalized-shape exponential family]
\label{rem:ub-mass-channel}
The mass \(M_t\) should not be encoded by appending the constant observable \(1\) to the normalized-shape information projection. Normalization already imposes \(\E_q[1]=1\), so doing so introduces an exactly redundant coordinate and makes the corresponding covariance matrix singular. The constant observable may, however, be included in the unbalanced Tangent observable vector after the shape has been calibrated. In that setting its spatial Jacobian vanishes, and its moment-rate equation isolates the total-mass reaction channel.
\end{remark}

\subsection{Finite-sample Law baseline}
\label{app:design-qualifications}

Equation~\eqref{eq:fide-design} defines the population ideal using the exact risk infimum. In the experiments, this value is approximated by the selection risk of a numerically optimized Law design. The resulting baseline defines a finite-search approximation to the admissible set rather than a certificate of global risk optimality. Risk tolerances apply to the selection distribution; their transfer to fresh observations is evaluated separately.

\subsection{Canonical information projection}
\label{app:proof-iprojection}

\paragraph{Technical conditions.}
Fix $(t,\etavec)$. Assume that $\cvec_{\etavec}(t)$ lies in the relative interior of the feasible moment set, that the dual objective $A_{t,\etavec}(\lambda)-\lambda^\top\cvec_{\etavec}(t)$ attains a minimizer $\lambdavec_{t,\etavec}$, and that $A_{t,\etavec}$ is finite on an open neighborhood of that minimizer. If $C_{t,\etavec}\succ0$ in a minimal observable representation, the calibrated multiplier is locally unique; without this rank condition, the projected law remains unique but its coordinates need not be.

\begin{proof}[Proof of the information-projection part of Theorem~\ref{thm:iprojection}]
Let
\begin{equation}
\mathcal Q_c:=\left\{Q\ll\widetilde Q_t:\E_Q[\phi]=c\right\}
\end{equation}
with the implicit finiteness conditions from \eqref{eq:iprojection}. For any $\lambda$ at which $A_{t,\etavec}(\lambda)<\infty$, define the exponentially tilted law
\begin{equation}
\frac{dQ_{\lambda}}{d\widetilde Q_t}(x)=\exp\!\left(\lambda^\top\phi(x)-A_{t,\etavec}(\lambda)\right).
\label{eq:appendix-tilted-law}
\end{equation}
The normalization follows directly from the definition of $A_{t,\etavec}$.

By assumption, the dual objective
\begin{equation}
D(\lambda):=A_{t,\etavec}(\lambda)-\lambda^\top c
\end{equation}
attains its minimum at $\lambda=\lambdavec_{t,\etavec}$, and $A_{t,\etavec}$ is finite on an open neighborhood of that point. On this neighborhood the standard differentiation formula for a log moment-generating function gives
\begin{equation}
\nabla_{\lambda}A_{t,\etavec}(\lambda)=\E_{Q_{\lambda}}[\phi],
\label{eq:appendix-grad-logpartition}
\end{equation}
and
\begin{equation}
\nabla_{\lambda}^2A_{t,\etavec}(\lambda)=\Cov_{Q_{\lambda}}(\phi,\phi).
\label{eq:appendix-hess-logpartition}
\end{equation}
Since $D$ is differentiable at an unconstrained minimizer,
\begin{equation}
0=\nabla D(\lambdavec_{t,\etavec})=\E_{Q_{\lambdavec_{t,\etavec}}}[\phi]-c.
\end{equation}
Hence the tilted law $Q_{\lambdavec_{t,\etavec}}$ belongs to $\mathcal Q_c$. In the shorthand of \eqref{eq:appendix-shorthand}, the calibration identity is
\begin{equation}
\E_{Q_t^{\etavec}}[\phi]=c.
\label{eq:moment-calibration}
\end{equation}

Set $Q^\star=Q_{\lambdavec_{t,\etavec}}$. For any feasible $Q\in\mathcal Q_c$ for which the displayed relative entropies are finite, the Radon--Nikodym chain rule and \eqref{eq:appendix-tilted-law} give
\begin{align}
\KL(Q\Vert\widetilde Q_t)&=\KL(Q\Vert Q^\star)+\E_Q\!\left[\log\frac{dQ^\star}{d\widetilde Q_t}\right]\nonumber\\
&=\KL(Q\Vert Q^\star)+\lambdavec_{t,\etavec}^\top\E_Q[\phi]-A_{t,\etavec}(\lambdavec_{t,\etavec})\nonumber\\
&=\KL(Q\Vert Q^\star)+\lambdavec_{t,\etavec}^\top c-A_{t,\etavec}(\lambdavec_{t,\etavec}).
\label{eq:appendix-pythagorean-kl}
\end{align}
For $Q=Q^\star$ the final two terms equal $\KL(Q^\star\Vert\widetilde Q_t)$. Therefore
\begin{equation}
\KL(Q\Vert\widetilde Q_t)-\KL(Q^\star\Vert\widetilde Q_t)=\KL(Q\Vert Q^\star)\ge0.
\label{eq:appendix-kl-gap}
\end{equation}
Equality holds only when $Q=Q^\star$ almost everywhere. Thus the I-projection is unique at the law level and has the density in \eqref{eq:exponential-tilt}.

Finally, \eqref{eq:appendix-hess-logpartition} evaluated at the calibrated multiplier gives
\begin{equation}
\nabla_{\lambda}^2A_{t,\etavec}(\lambdavec_{t,\etavec})=C_{t,\etavec}.
\end{equation}
If $C_{t,\etavec}$ is positive definite in a minimal observable representation, the dual Hessian is nonsingular at the optimizer. By continuity of the Hessian, the dual objective is strictly convex on a sufficiently small neighborhood, and the multiplier is locally unique. If the observable coordinates are redundant, different multipliers can generate the same log-density up to an additive constant on the support of $\widetilde Q_t$, so multiplier non-uniqueness does not contradict law-level uniqueness.
\end{proof}

\begin{proposition}[Finite-reference information-projection feasibility]
\label{prop:empirical-iprojection-feasibility}
Fix $(t,\etavec)$ and let the empirical reference be $\widehat{\widetilde Q}_{N}=\sum_{i=1}^{N}\bar w_i\delta_{x_i}$ with $\bar w_i>0$ and $\sum_i\bar w_i=1$. Write $z_i=\Phivec_{\etavec}(x_i)$ and $\mathcal K_N=\operatorname{conv}\{z_1,\ldots,z_N\}$. For a target moment $\widehat c$, the empirical constraint $\sum_i p_i z_i=\widehat c$, $p_i\ge0$, $\sum_i p_i=1$, is feasible if and only if $\widehat c\in\mathcal K_N$, and whenever it is feasible the empirical KL projection relative to $(\bar w_i)_{i=1}^N$ is unique. If $\widehat c\in\operatorname{relint}(\mathcal K_N)$, the minimizer has strictly positive weights of the form
\begin{equation} p_i(\lambda)=\frac{\bar w_i\exp(\lambda^\top z_i)}{\sum_{j=1}^{N}\bar w_j\exp(\lambda^\top z_j)},\qquad \sum_{i=1}^{N}p_i(\lambda)z_i=\widehat c. \label{eq:empirical-iprojection-weights} \end{equation}
After removal of affine redundancies in the observable coordinates, the corresponding finite multiplier is unique and the projected covariance $\widehat C=\sum_i p_i(\lambda)(z_i-\widehat c)(z_i-\widehat c)^\top$ is positive definite. If $\widehat c$ lies on the relative boundary of $\mathcal K_N$, the unique empirical KL projection still exists but cannot be represented by a finite full-support exponential tilt; it is obtained only as a limiting tilt with vanishing weights on part of the empirical support. If $\widehat c\notin\mathcal K_N$, no exact empirical information projection supported on the reference bank exists.
\end{proposition}

\begin{proof}
The moments attainable by probability weights on the empirical support are exactly the convex combinations of $z_1,\ldots,z_N$, which proves the feasibility statement. On the feasible simplex the objective $\sum_i p_i\log(p_i/\bar w_i)$ is strictly convex, so its minimizer is unique. For a target in $\operatorname{relint}(\mathcal K_N)$, the equality-constrained first-order conditions give the exponential weights in \eqref{eq:empirical-iprojection-weights}; conversely, every finite multiplier gives strictly positive weights and therefore a moment in $\operatorname{relint}(\mathcal K_N)$. In a minimal affine representation, zero variance of a nonzero linear combination of the features would make that combination constant on the full empirical support, contradicting minimality, so $\widehat C\succ0$. A target on a proper relative face requires the minimizing distribution to place zero mass outside that face and hence cannot be produced by finite strictly positive exponential weights. Outside $\mathcal K_N$ the moment constraint is infeasible.
\end{proof}

\begin{corollary}[Finite-measure information projection with prescribed mass]
\label{cor:ub-iprojection}
Fix $M>0$ and a raw finite-measure moment vector $\bar c$. Let $c=\bar c/M$, and suppose the conditions of Theorem~\ref{thm:iprojection} hold for this normalized moment vector. Then the finite measure $\mu^\star=MQ^\star$, where $Q^\star$ is the information projection in Theorem~\ref{thm:iprojection}, is the unique finite measure of mass $M$ whose normalized shape minimizes $\KL(Q\Vert\widetilde Q_t)$ subject to $\int\phi\,d(MQ)=\bar c$. Equivalently, $\mu^\star$ uniquely minimizes $M\KL(Q\Vert\widetilde Q_t)$ over the same fixed-mass fiber, and its normalized density has the same exponential-tilt representation \eqref{eq:exponential-tilt}.
\end{corollary}

\begin{proof}
Every finite measure $\mu$ of mass $M$ can be written uniquely as $\mu=MQ$ with $Q$ a probability law, and the raw constraint $\int\phi\,d\mu=\bar c$ is equivalent to $\E_Q[\phi]=\bar c/M=c$. Since multiplication of the objective by the positive constant $M$ does not change its minimizer, Theorem~\ref{thm:iprojection} applies directly to the normalized shape.
\end{proof}

\subsection{Physical-time multiplier dynamics}
\label{app:proof-lambda-dynamics}

\begin{proposition}[Multiplier dynamics]
\label{prop:lambda-dynamics}
Under the regularity assumptions stated below and in a minimal observable representation with $C_{t,\etavec}\succ0$, the calibrated multiplier satisfies \eqref{eq:lambda-dot}.
\end{proposition}

\paragraph{Technical conditions.}
Fix $\etavec$ and a time $t$. Assume that $\cvec_{\etavec}(s)$ is continuously differentiable in $s$ near $t$, that the reference density and velocity are regular enough for the reference continuity equation and the integrations by parts below, and that the normalized exponential family $q_{\lambda,s}$ defined below is continuously differentiable in $(\lambda,s)$ near $(\lambdavec_{t,\etavec},t)$ with differentiation and integration interchangeable. In a minimal observable representation, assume $C_{t,\etavec}\succ0$.

\begin{proof}[Proof of Proposition~\ref{prop:lambda-dynamics}]
Fix $\etavec$ and suppress it from the notation. For $\lambda$ and $t$ in the neighborhood specified above, define
\begin{equation}
q_{\lambda,t}(x):=\widetilde q_t(x)\exp\!\left(\lambda^\top\phi(x)-A_t(\lambda)\right),
\end{equation}
and the calibration residual
\begin{equation}
\mathcal F(\lambda,t):=\E_{q_{\lambda,t}}[\phi]-c_t.
\end{equation}
By the stated regularity assumptions, $\mathcal F$ is continuously differentiable near $(\lambda_t,t)$, and
\begin{equation}
\partial_{\lambda}\mathcal F(\lambda_t,t)=\Cov_{Q_t^{\etavec}}(\phi,\phi)=C_{t,\etavec}.
\end{equation}
Since $C_{t,\etavec}\succ0$, the implicit-function theorem gives a locally unique continuously differentiable calibrated path $t\mapsto\lambda_t$ satisfying $\mathcal F(\lambda_t,t)=0$. In particular, $\dot\lambda_t$ exists, so the differentiations below are justified. Let
\begin{equation}
s_t(x):=\partial_t\log\widetilde q_t(x)
\end{equation}
where the reference density is positive. From the exponential representation,
\begin{equation}
\log q_t(x)=\log\widetilde q_t(x)+\lambda_t^\top\phi(x)-A_t(\lambda_t).
\label{eq:appendix-logq}
\end{equation}
At fixed $\lambda$, differentiation of the log-partition function in physical time gives
\begin{equation}
\partial_t A_t(\lambda)=\E_{q_{\lambda,t}}[s_t].
\label{eq:appendix-partial-A-time}
\end{equation}
Along the calibrated path $t\mapsto\lambda_t$,
\begin{equation}
\frac{d}{dt}A_t(\lambda_t)=\E_{Q_t^{\etavec}}[s_t]+\dot\lambda_t^\top c_t,
\label{eq:appendix-total-A-time}
\end{equation}
where $c_t=\E_{Q_t^{\etavec}}[\phi]$.

Differentiating the calibration identity $c_t=\E_{Q_t^{\etavec}}[\phi]$ and using that $\phi$ has no explicit physical-time dependence yields
\begin{align}
\dot c_t&=\E_{Q_t^{\etavec}}\!\left[\phi\,\partial_t\log q_t\right]\nonumber\\
&=\Cov_{Q_t^{\etavec}}(\phi,s_t)+C_{t,\etavec}\dot\lambda_t.
\label{eq:appendix-calibration-derivative}
\end{align}
It remains to express the covariance with $s_t$ without differentiating the reference density explicitly.

Let
\begin{equation}
w_t(x):=\frac{q_t(x)}{\widetilde q_t(x)}=\exp\!\left(\lambda_t^\top\phi(x)-A_t(\lambda_t)\right).
\end{equation}
The reference continuity equation gives $\partial_t\widetilde q_t=-\nabla\cdot(\widetilde q_tu_t)$. Therefore, using the assumed boundary conditions,
\begin{align}
\E_{Q_t^{\etavec}}[s_t]&=\int w_t\,\partial_t\widetilde q_t\,dx\nonumber\\
&=-\int w_t\,\nabla\cdot(\widetilde q_tu_t)\,dx\nonumber\\
&=\int\widetilde q_tu_t\cdot\nabla w_t\,dx.
\label{eq:appendix-mean-score-1}
\end{align}
Since
\begin{equation}
\nabla w_t=w_tJ_\phi^\top\lambda_t,
\end{equation}
we obtain
\begin{equation}
\E_{Q_t^{\etavec}}[s_t]=\E_{Q_t^{\etavec}}\!\left[\lambda_t^\top J_\phi u_t\right]=\E_{Q_t^{\etavec}}[\lambda_t^\top m_t].
\label{eq:appendix-mean-score}
\end{equation}
Similarly, componentwise integration by parts gives
\begin{align}
\E_{Q_t^{\etavec}}[\phi s_t]&=\int w_t\phi\,\partial_t\widetilde q_t\,dx\nonumber\\
&=\int\widetilde q_tu_t\cdot\nabla(w_t\phi)\,dx\nonumber\\
&=\E_{Q_t^{\etavec}}[m_t]+\E_{Q_t^{\etavec}}[\phi\,\lambda_t^\top m_t].
\label{eq:appendix-phi-score}
\end{align}
Subtracting $\E[\phi]\E[s_t]=c_t\E[\lambda_t^\top m_t]$ yields
\begin{equation}
\Cov_{Q_t^{\etavec}}(\phi,s_t)=\E_{Q_t^{\etavec}}[m_t]+\Cov_{Q_t^{\etavec}}(\phi,\lambda_t^\top m_t).
\label{eq:appendix-score-covariance}
\end{equation}
Substitution into \eqref{eq:appendix-calibration-derivative} gives
\begin{equation}
C_{t,\etavec}\dot\lambda_t=\dot c_t-\E_{Q_t^{\etavec}}[m_t]-\Cov_{Q_t^{\etavec}}(\phi,\lambda_t^\top m_t),
\end{equation}
which is \eqref{eq:lambda-dot}. Positive definiteness of $C_{t,\etavec}$ in the stated minimal representation makes this a uniquely solvable $R\times R$ linear system. This proves the multiplier-dynamics clause of Theorem~\ref{thm:iprojection}.
\end{proof}

\begin{corollary}[Multiplier dynamics for a finite-measure path]
\label{cor:ub-lambda-dynamics}
Suppose the assumptions of Proposition~\ref{prop:lambda-dynamics} hold for the normalized shape $q_t$, and suppose $M_t>0$ and the raw moment path $\bar c_t$ in \eqref{eq:ub-raw-shape-moments} are continuously differentiable. Then the same covariance equation holds with the normalized target
\begin{equation}
c_t=\frac{\bar c_t}{M_t},\qquad \dot c_t=\frac{\dot{\bar c}_t}{M_t}-\frac{\dot M_t}{M_t}c_t,
\label{eq:ub-cdot}
\end{equation}
namely
\begin{equation}
C_{t,\etavec}\dot\lambda_{t,\etavec}=\frac{\dot{\bar c}_t}{M_t}-\frac{\dot M_t}{M_t}c_t-\E_{Q_t^{\etavec}}[m_{t,\etavec}]-\Cov_{Q_t^{\etavec}}\!\left(\Phivec_{\etavec},\lambdavec_{t,\etavec}^\top m_{t,\etavec}\right).
\label{eq:ub-lambda-dot}
\end{equation}
\end{corollary}

\begin{proof}
The information projection acts only on the normalized shape, so Proposition~\ref{prop:lambda-dynamics} applies with target $c_t=\bar c_t/M_t$. Equation \eqref{eq:ub-cdot} is the quotient rule, and substitution into \eqref{eq:lambda-dot} gives \eqref{eq:ub-lambda-dot}.
\end{proof}

\subsection{Density-free continuity forcing}
\label{app:proof-continuity-forcing}

\begin{proposition}[Density-free continuity forcing]
\label{prop:continuity-forcing}
Under the assumptions of Proposition~\ref{prop:lambda-dynamics}, the reference-relative continuity residual \(h_{t,\etavec}\) defined in Section~\ref{subsec:methodology-full} is given by \eqref{eq:computable-h} and has zero projected-law mean:
\begin{equation}
\E_{Q_t^{\etavec}}[h_{t,\etavec}]=0.
\label{eq:h-zero-mean}
\end{equation}
\end{proposition}

\begin{proof}[Proof of Proposition~\ref{prop:continuity-forcing}]
Again fix $\etavec$ and write $q_t=q_t^{\etavec}$. For any positive sufficiently regular density $q$ and vector field $u$,
\begin{equation}
\partial_tq+\nabla\cdot(qu)=q\left[(\partial_t+u\cdot\nabla)\log q+\nabla\cdot u\right].
\label{eq:appendix-material-log-density}
\end{equation}
The reference continuity equation implies
\begin{equation}
(\partial_t+u_t\cdot\nabla)\log\widetilde q_t+\nabla\cdot u_t=0.
\label{eq:appendix-reference-material}
\end{equation}
Using \eqref{eq:appendix-logq}, the time independence of $\phi$, and $m_t=J_\phi u_t$,
\begin{align}
(\partial_t+u_t\cdot\nabla)\log q_t+\nabla\cdot u_t&=\dot\lambda_t^\top\phi+\lambda_t^\top m_t-\frac{d}{dt}A_t(\lambda_t).
\label{eq:appendix-residual-step1}
\end{align}
Equations \eqref{eq:appendix-total-A-time} and \eqref{eq:appendix-mean-score} give
\begin{equation}
\frac{d}{dt}A_t(\lambda_t)=\dot\lambda_t^\top c_t+\E_{Q_t^{\etavec}}[\lambda_t^\top m_t].
\label{eq:appendix-A-dot-final}
\end{equation}
Hence the bracket in \eqref{eq:appendix-material-log-density} equals
\begin{equation}
\dot\lambda_t^\top(\phi-c_t)+\lambda_t^\top\left(m_t-\E_{Q_t^{\etavec}}[m_t]\right)=h_{t,\etavec}.
\end{equation}
Substituting into \eqref{eq:appendix-material-log-density} gives
\(\partial_tq_t+\nabla\!\cdot(q_tu_t)=q_th_{t,\etavec}\)
and proves the explicit formula \eqref{eq:computable-h}.

Finally, calibration gives $\E_{Q_t^{\etavec}}[\phi-c_t]=0$, and the second centered term also has zero expectation, proving \eqref{eq:h-zero-mean}. Equivalently, the zero-mean property follows by integrating the reference-relative continuity identity over space and using conservation of total mass. Together with Proposition~\ref{prop:lambda-dynamics} and the information-projection argument above, this completes the proof of Theorem~\ref{thm:iprojection}.
\end{proof}

\begin{corollary}[Density-free unbalanced forcing]
\label{cor:ub-continuity-forcing}
Under the assumptions of Corollary~\ref{cor:ub-lambda-dynamics}, the normalized-shape residual satisfies
\begin{equation}
h_{\mathrm{shape}}(x)=\dot\lambda_t^\top(\phi(x)-c_t)+\lambda_t^\top\left(m_t(x)-\E_{q_t}[m_t]\right),
\label{eq:ub-shape-forcing}
\end{equation}
and therefore the finite-measure reference-relative forcing in \eqref{eq:ub-residual} is
\begin{equation}
h_{\mathrm{ub}}(x)=\dot\lambda_t^\top(\phi(x)-c_t)+\lambda_t^\top\left(m_t(x)-\E_{q_t}[m_t]\right)+\frac{\dot M_t}{M_t}-g_{\mathrm{ref}}(t).
\label{eq:ub-computable-h}
\end{equation}
Moreover,
\begin{equation}
\E_{q_t}[h_{\mathrm{shape}}]=0,\qquad \E_{q_t}[h_{\mathrm{ub}}]=\frac{\dot M_t}{M_t}-g_{\mathrm{ref}}(t).
\label{eq:ub-h-mean}
\end{equation}
\end{corollary}

\begin{proof}
Equation \eqref{eq:ub-shape-forcing} is Proposition~\ref{prop:continuity-forcing} applied to the normalized shape. Adding the scalar mass-rate discrepancy from \eqref{eq:ub-residual} gives \eqref{eq:ub-computable-h}. Taking expectation and using the centering in \eqref{eq:ub-shape-forcing} gives \eqref{eq:ub-h-mean}.
\end{proof}

\subsection{Minimum-energy Full correction}
\label{app:proof-poisson}

Fix $(t,\etavec)$ and write $q=q_t^{\etavec}$ and $h=h_{t,\etavec}$. Let $H_q$ be the mean-zero weighted energy space defined in Appendix~\ref{app:preliminaries}. Assume that the weighted Dirichlet form is coercive on this space, for example through a weighted Poincar\'e inequality, and that $\varphi\mapsto\E_q[h\varphi]$ defines a bounded linear functional on $H_q$. The weak Poisson equation is
\begin{equation}
\E_q[\nabla\psi\cdot\nabla\varphi]=-\E_q[h\varphi]\qquad\text{for every }\varphi\in H_q.
\label{eq:poisson-weak}
\end{equation}

\begin{proof}[Proof of the minimum-energy and Poisson parts of Theorem~\ref{thm:poisson-correction}]
Define the bilinear form
\begin{equation}
a(\psi,\varphi):=\E_q[\nabla\psi\cdot\nabla\varphi]
\end{equation}
on the mean-zero energy space $H_q$. By construction of the energy norm,
\begin{equation}
|a(\psi,\varphi)|\le\norm{\psi}_{H_q}\norm{\varphi}_{H_q},\qquad a(\psi,\psi)=\norm{\psi}_{H_q}^2.
\label{eq:appendix-bilinear-coercive}
\end{equation}
The weighted Poincar\'e/coercivity assumption ensures that the mean-zero gauge removes the constant null direction and that this energy space is the appropriate Hilbert space for the problem. By assumption,
\begin{equation}
\ell(\varphi):=-\E_q[h\varphi]
\end{equation}
is a bounded linear functional on $H_q$. Lax--Milgram therefore gives a unique $\psi\in H_q$ satisfying
\begin{equation}
a(\psi,\varphi)=\ell(\varphi)\qquad\text{for every }\varphi\in H_q,
\end{equation}
which is exactly \eqref{eq:poisson-weak}. Since $\E_q[h]=0$, the restriction to mean-zero test functions causes no loss: for any admissible $\varphi$, setting $\varphi_0:=\varphi-\E_q[\varphi]$ gives $\nabla\varphi_0=\nabla\varphi$ and $\E_q[h\varphi_0]=\E_q[h\varphi]$, so the same weak identity holds for arbitrary admissible test functions. Under the stated periodic, decay, or no-flux boundary conditions, integration by parts converts this weak identity into
\begin{equation}
\nabla\cdot(q\nabla\psi)=qh
\end{equation}
in the distributional sense, together with the gauge $\E_q[\psi]=0$.

Set
\begin{equation}
\delta^\star:=-\nabla\psi.
\end{equation}
Then $\delta^\star\in L^2(q;\R^d)$. The weak Poisson equation gives, for every $\varphi\in H_q$,
\begin{equation}
\E_q[\delta^\star\cdot\nabla\varphi]=\E_q[h\varphi].
\label{eq:appendix-delta-star-feasible}
\end{equation}
This is precisely the weak form of
\begin{equation}
\nabla\cdot(q\delta^\star)=-qh,
\end{equation}
so $u_t+\delta^\star$ satisfies the projected-law continuity equation.

Now let $\delta\in L^2(q;\R^d)$ be any other feasible correction and define $w:=\delta-\delta^\star$. Subtracting the two weak constraints gives the identity first on the admissible test class. Because $w\in L^2(q;\R^d)$, the map $\varphi\mapsto\E_q[w\cdot\nabla\varphi]$ is continuous in the $H_q$ energy norm; density therefore extends the identity to all of $H_q$:
\begin{equation}
\E_q[w\cdot\nabla\varphi]=0\qquad\text{for every }\varphi\in H_q.
\label{eq:appendix-weighted-div-free}
\end{equation}
Taking $\varphi=\psi$ yields
\begin{equation}
\E_q[w\cdot\nabla\psi]=0,
\end{equation}
and therefore, because $\delta^\star=-\nabla\psi$,
\begin{equation}
\E_q[\delta^\star\cdot w]=0.
\label{eq:appendix-orthogonality}
\end{equation}
Consequently
\begin{align}
\E_q[\norm{\delta}^2]&=\E_q[\norm{\delta^\star+w}^2]\nonumber\\
&=\E_q[\norm{\delta^\star}^2]+\E_q[\norm{w}^2]\nonumber\\
&\ge\E_q[\norm{\delta^\star}^2].
\label{eq:appendix-pythagorean-energy}
\end{align}
Equality holds if and only if $w=0$ $q$-almost everywhere. Thus $\delta^\star$ is the unique minimum-energy feasible correction, and
\begin{equation}
a_{\mathrm{full}}(t,\etavec)=\E_q[\norm{\delta^\star}^2]=\E_q[\norm{\nabla\psi}^2].
\end{equation}
This proves the minimum-energy and Poisson assertions in Theorem~\ref{thm:poisson-correction}. The weighted $H^{-1}$ identity in \eqref{eq:poisson-action} is proved in Appendix~\ref{app:proof-hminusone}.
\end{proof}

The screened operator below is the reference-relative analogue of the transport--reaction operator in Sobolev--Fisher discrepancy and Unbalanced Sobolev Descent \citep{mroueh2020unbalanced}. Its gradient and zeroth-order terms correspond to transport and reaction costs, respectively; the present forcing comes from the prescribed finite-measure path, with the boundary conditions stated in this appendix.

\begin{corollary}[Minimum-energy unbalanced move--reaction correction]
\label{cor:ub-full-correction}
Fix $(t,\etavec)$, write $q=q_t^{\etavec}$, $M=M_t$, and $h_{\mathrm{ub}}=h_{\mathrm{ub},t,\etavec}$, and let $\kappa>0$. Let $H_{q,\kappa}$ be the completion of the admissible test functions under
\begin{equation}
\norm{\psi}_{H_{q,\kappa}}^2:=\E_q\!\left[\norm{\nabla\psi}^2+\frac{\psi^2}{\kappa}\right].
\label{eq:ub-screened-energy-norm}
\end{equation}
Assume that $\varphi\mapsto\E_q[h_{\mathrm{ub}}\varphi]$ is a bounded linear functional on $H_{q,\kappa}$. Then there exists a unique $\psi\in H_{q,\kappa}$ satisfying
\begin{equation}
\E_q\!\left[\nabla\psi\cdot\nabla\varphi+\frac{1}{\kappa}\psi\varphi\right]=\E_q[h_{\mathrm{ub}}\varphi]\qquad\text{for every }\varphi\in H_{q,\kappa}.
\label{eq:ub-screened-poisson-weak}
\end{equation}
Under the stated boundary and regularity assumptions, this is the weak form of
\begin{equation}
-\nabla\!\cdot(q\nabla\psi)+\frac{q}{\kappa}\psi=qh_{\mathrm{ub}}.
\label{eq:ub-screened-poisson}
\end{equation}
The unique minimum-energy correction in \eqref{eq:ub-full-action-definition} is
\begin{equation}
\delta^\star=\nabla\psi,\qquad \alpha^\star=\frac{\psi}{\kappa},
\label{eq:ub-optimal-correction}
\end{equation}
and
\begin{equation}
a_{\mathrm{full}}^{\mathrm{ub}}(t,\etavec)=M\E_q\!\left[\norm{\nabla\psi}^2+\frac{\psi^2}{\kappa}\right].
\label{eq:ub-full-action-energy}
\end{equation}
\end{corollary}

\begin{proof}
Define
\begin{equation}
a_\kappa(\psi,\varphi):=\E_q\!\left[\nabla\psi\cdot\nabla\varphi+\frac{1}{\kappa}\psi\varphi\right].
\end{equation}
This bilinear form is continuous and coercive on $H_{q,\kappa}$ because $a_\kappa(\psi,\psi)=\norm{\psi}_{H_{q,\kappa}}^2$. Lax--Milgram therefore gives a unique solution of \eqref{eq:ub-screened-poisson-weak}. Integration by parts yields \eqref{eq:ub-screened-poisson} whenever the strong form is justified.

Set $(\delta^\star,\alpha^\star)=(\nabla\psi,\psi/\kappa)$. Equation \eqref{eq:ub-screened-poisson-weak} gives
\begin{equation}
\E_q[\delta^\star\cdot\nabla\varphi+\alpha^\star\varphi]=\E_q[h_{\mathrm{ub}}\varphi],
\end{equation}
which is the weak form of \eqref{eq:ub-correction-constraint}, so $(\delta^\star,\alpha^\star)$ is feasible. Let $(\delta,\alpha)$ be any other feasible pair and write $w=\delta-\delta^\star$ and $\gamma=\alpha-\alpha^\star$. Subtracting the two weak constraints gives
\begin{equation}
\E_q[w\cdot\nabla\varphi+\gamma\varphi]=0
\end{equation}
for every $\varphi\in H_{q,\kappa}$. Taking $\varphi=\psi$ yields
\begin{equation}
\E_q\!\left[w\cdot\delta^\star+\kappa\gamma\alpha^\star\right]=\E_q\!\left[w\cdot\nabla\psi+\gamma\psi\right]=0.
\end{equation}
Hence, in the product norm induced by $\E_q[\|\delta\|^2+\kappa\alpha^2]$,
\begin{align}
\E_q[\|\delta\|^2+\kappa\alpha^2]&=\E_q[\|\delta^\star\|^2+\kappa(\alpha^\star)^2]+\E_q[\|w\|^2+\kappa\gamma^2]\nonumber\\
&\ge\E_q[\|\delta^\star\|^2+\kappa(\alpha^\star)^2].
\end{align}
Equality holds only when $w=0$ and $\gamma=0$ in the corresponding $L^2(q)$ spaces. Multiplication by $M>0$ proves both uniqueness and \eqref{eq:ub-full-action-energy}.
\end{proof}

\begin{corollary}[Mass--shape decomposition of unbalanced Full action]
\label{cor:ub-mass-shape-decomposition}
Under the assumptions of Corollary~\ref{cor:ub-full-correction}, define the
spatially constant logarithmic mass-rate discrepancy
\begin{equation}
b_t:=\frac{\dot M_t}{M_t}-g_{\mathrm{ref}}(t),
\label{eq:ub-mass-rate-discrepancy}
\end{equation}
so that
\begin{equation}
h_{\mathrm{ub}}=h_{\mathrm{shape}}+b_t,
\qquad
\E_{q_t}[h_{\mathrm{shape}}]=0.
\label{eq:ub-shape-mass-forcing-split}
\end{equation}
Let $\psi_{\mathrm{shape}}\in H_{q_t,\kappa}$ denote the unique solution of
\begin{equation}
\E_{q_t}\!\left[
\nabla\psi_{\mathrm{shape}}\cdot\nabla\varphi
+\frac{1}{\kappa}\psi_{\mathrm{shape}}\varphi
\right]
=
\E_{q_t}[h_{\mathrm{shape}}\varphi]
\qquad
\text{for every }\varphi\in H_{q_t,\kappa}.
\label{eq:ub-shape-screened-weak}
\end{equation}
Then
\begin{equation}
\E_{q_t}[\psi_{\mathrm{shape}}]=0,
\qquad
\psi=\psi_{\mathrm{shape}}+\kappa b_t,
\label{eq:ub-potential-mass-shape-split}
\end{equation}
where $\psi$ is the Full screened potential from
Corollary~\ref{cor:ub-full-correction}. Consequently,
\begin{equation}
\delta^\star=\nabla\psi_{\mathrm{shape}},
\qquad
\alpha^\star=b_t+\frac{\psi_{\mathrm{shape}}}{\kappa},
\label{eq:ub-correction-mass-shape-split}
\end{equation}
and the instantaneous unbalanced Full action decomposes orthogonally as
\begin{equation}
a_{\mathrm{full}}^{\mathrm{ub}}(t,\etavec)
=
M_t\kappa b_t^2
+
M_t\E_{q_t}\!\left[
\|\nabla\psi_{\mathrm{shape}}\|^2
+\frac{\psi_{\mathrm{shape}}^2}{\kappa}
\right].
\label{eq:ub-mass-shape-action-decomposition}
\end{equation}
The first term is the cost of the spatially uniform reaction required to
correct the total-mass rate, while the second is the screened correction
cost of the mean-zero shape forcing.
\end{corollary}

\begin{proof}
Because $\E_{q_t}[h_{\mathrm{shape}}]=0$, testing
\eqref{eq:ub-shape-screened-weak} with $\varphi=1$ gives
\begin{equation}
\frac{1}{\kappa}\E_{q_t}[\psi_{\mathrm{shape}}]=0,
\end{equation}
and hence $\E_{q_t}[\psi_{\mathrm{shape}}]=0$.

Since constants have zero spatial gradient,
\begin{equation}
-\nabla\!\cdot\!\left(q_t\nabla(\kappa b_t)\right)
+\frac{q_t}{\kappa}(\kappa b_t)
=
q_tb_t.
\end{equation}
Adding this identity to the screened equation for
$\psi_{\mathrm{shape}}$ shows that
\begin{equation}
\psi=\psi_{\mathrm{shape}}+\kappa b_t
\end{equation}
solves
\begin{equation}
-\nabla\!\cdot(q_t\nabla\psi)
+\frac{q_t}{\kappa}\psi
=
q_t(h_{\mathrm{shape}}+b_t)
=
q_th_{\mathrm{ub}}.
\end{equation}
Uniqueness in Corollary~\ref{cor:ub-full-correction} therefore identifies
this sum with the Full screened potential. Equation
\eqref{eq:ub-correction-mass-shape-split} follows immediately from
\eqref{eq:ub-optimal-correction}.

Finally,
\begin{align}
\E_{q_t}\!\left[
\|\nabla\psi\|^2+\frac{\psi^2}{\kappa}
\right]
&=
\E_{q_t}\!\left[
\|\nabla\psi_{\mathrm{shape}}\|^2
+
\frac{(\psi_{\mathrm{shape}}+\kappa b_t)^2}{\kappa}
\right]
\nonumber\\
&=
\E_{q_t}\!\left[
\|\nabla\psi_{\mathrm{shape}}\|^2
+\frac{\psi_{\mathrm{shape}}^2}{\kappa}
\right]
+
2b_t\E_{q_t}[\psi_{\mathrm{shape}}]
+
\kappa b_t^2
\nonumber\\
&=
\E_{q_t}\!\left[
\|\nabla\psi_{\mathrm{shape}}\|^2
+\frac{\psi_{\mathrm{shape}}^2}{\kappa}
\right]
+
\kappa b_t^2.
\end{align}
Multiplying by $M_t$ and using
\eqref{eq:ub-full-action-energy} proves
\eqref{eq:ub-mass-shape-action-decomposition}.
\end{proof}

\begin{remark}[Operator form of the mass--shape decomposition]
\label{rem:ub-mass-shape-operator}
Suppose additionally that the weighted Dirichlet form admits the
self-adjoint operator representation
\begin{equation}
\mathsf L_{q_t}
=
-q_t^{-1}\nabla\!\cdot(q_t\nabla)
\end{equation}
on $L^2(q_t)$, so that
$\mathsf L_{q_t}+\kappa^{-1}I$ is invertible. Then
\begin{equation}
\psi_{\mathrm{shape}}
=
\left(\mathsf L_{q_t}+\kappa^{-1}I\right)^{-1}
h_{\mathrm{shape}},
\end{equation}
and Corollary~\ref{cor:ub-mass-shape-decomposition} becomes
\begin{equation}
a_{\mathrm{full}}^{\mathrm{ub}}(t,\etavec)
=
M_t\kappa
\left(
\frac{\dot M_t}{M_t}-g_{\mathrm{ref}}(t)
\right)^2
+
M_t
\left\langle
h_{\mathrm{shape}},
\left(\mathsf L_{q_t}+\kappa^{-1}I\right)^{-1}
h_{\mathrm{shape}}
\right\rangle_{L^2(q_t)}.
\label{eq:ub-mass-shape-operator-decomposition}
\end{equation}
Thus the spatially constant mass-rate mismatch and the mean-zero shape
forcing are orthogonal in the screened energy geometry.
\end{remark}

\begin{remark}[Interpretation of the reaction penalty]
\label{rem:ub-kappa-interpretation}
The parameter $\kappa>0$ fixes the relative cost of reaction versus transport and is chosen before measurement-design optimization. Since $\delta$ has units of state-distance per unit time and $\alpha$ has units of inverse time, $\kappa$ has units of squared state distance, or is dimensionless after the corresponding nondimensionalization. Larger $\kappa$ makes reaction more expensive, while smaller $\kappa$ allows the forcing to be realized more cheaply through reaction. On the mean-zero shape subspace, $(\mathsf L_q+\kappa^{-1}I)^{-1}\to\mathsf L_q^{-1}$ as $\kappa\to\infty$ whenever the balanced inverse is well defined. Thus, if $b_t=\dot M_t/M_t-g_{\mathrm{ref}}(t)=0$, the unbalanced shape cost approaches the balanced transport cost; if $b_t\neq0$, the mass term $M_t\kappa b_t^2$ diverges, so no finite balanced limit exists. We therefore treat $\kappa$ as a fixed modeling choice rather than tuning it separately across competing designs.
\end{remark}

\begin{remark}[Screening removes the balanced compatibility and gauge conditions]
\label{rem:ub-screening-coercivity}
The reaction term removes the constant null direction present in the balanced Dirichlet seminorm. In the balanced problem one therefore imposes a mean-zero gauge and requires $\E_q[h]=0$. For finite $\kappa$, the term $\kappa^{-1}\E_q[\psi^2]$ controls constants directly, so neither a gauge condition nor zero-mean forcing is required. Testing \eqref{eq:ub-screened-poisson-weak} with $\varphi=1$ gives $\E_q[\psi]/\kappa=\E_q[h_{\mathrm{ub}}]$. By Corollary~\ref{cor:ub-mass-shape-decomposition}, this mean component corresponds exactly to the spatially uniform reaction $b_t=\dot M_t/M_t-g_{\mathrm{ref}}(t)$, while the remaining screened potential has zero $q_t$-mean.
\end{remark}

\begin{remark}[Scope of regularized density--source evaluations]
\label{rem:regularized-density-source-scope}
At a fixed time, the weighted-Poisson construction applies to a probability density \(q\) and a density-level source \(s=qh\) satisfying \(\E_q[h]=0\), whether or not this pair has been derived from a prescribed law path. Under the assumptions of Appendix~\ref{app:proof-poisson} for \(h=s/q\), define
\begin{equation} \mathfrak a(q,s):=\inf_{\delta\in L^2(q;\R^d)}\left\{\E_q[\|\delta\|^2]:\nabla\!\cdot(q\delta)=-s\right\}=\|s/q\|_{H^{-1}(q)}^2. \label{eq:general-density-source-action} \end{equation}
This recovers the Full action of a specified path \(q_t\) relative to \(u_t\) when \(s_t=\partial_tq_t+\nabla\!\cdot(q_tu_t)\). The Poisson variational identity alone does not establish this source identity.

For example, let \(S\) be a fixed convolution with a probability kernel on \(\R^d\) or a flat torus, and assume the derivatives below commute with convolution. If \(s_t=\partial_tq_t+\nabla\!\cdot(q_tu_t)\), then
\begin{equation} \partial_t(Sq_t)+\nabla\!\cdot((Sq_t)u_t)=Ss_t+\nabla\!\cdot\bigl((Sq_t)u_t-S(q_tu_t)\bigr). \label{eq:smoothing-continuity-commutator} \end{equation}
Thus separately smoothing the density and source generally produces the represented cost \(\mathfrak a(Sq_t,Ss_t)\), not automatically the Full action of \(Sq_t\) under the unchanged velocity \(u_t\). Moreover, with \(S^\ast\) denoting the adjoint and the integrals understood coordinatewise,
\[ \int\Phivec_{\etavec}(x)(Sq_t)(x)\,dx=\int(S^\ast\Phivec_{\etavec})(x)q_t(x)\,dx, \]
so smoothing need not preserve calibration against the original observables. The continuum theorems therefore do not by themselves establish consistency or convergence of an empirical or spatially regularized evaluator. Discrete minimum-norm and orthogonality identities concern the specified discrete source, operators, and inner product; identifying their value with the population Full action requires a separate consistency argument.
\end{remark}

\subsection{Weighted negative-Sobolev interpretation}
\label{app:proof-hminusone}

Define the weighted negative-Sobolev norm
\begin{equation}
\norm{h}_{H^{-1}(q)}:=\sup_{\varphi\in H_q:\,\E_q[\norm{\nabla\varphi}^2]\le1}\left|\E_q[h\varphi]\right|.
\label{eq:hminusone-norm}
\end{equation}

\begin{lemma}[Weighted negative-Sobolev identity]
\label{lem:hminusone}

Under the coercivity and regularity assumptions of Appendix~\ref{app:proof-poisson},
\begin{equation}
a_{\mathrm{full}}(t,\etavec)=\norm{h_{t,\etavec}}_{H^{-1}(q_t^{\etavec})}^2.
\label{eq:hminusone-action}
\end{equation}
\end{lemma}

\begin{proof}[Proof of Lemma~\ref{lem:hminusone}]
On $H_q$ use the inner product
\begin{equation}
\langle\psi,\varphi\rangle_{H_q}:=\E_q[\nabla\psi\cdot\nabla\varphi].
\end{equation}
Theorem~\ref{thm:poisson-correction} gives
\begin{equation}
\E_q[h\varphi]=-\langle\psi,\varphi\rangle_{H_q}
\end{equation}
for every $\varphi\in H_q$. Hence, by Cauchy--Schwarz,
\begin{equation}
|\E_q[h\varphi]|\le\norm{\psi}_{H_q}\norm{\varphi}_{H_q},
\end{equation}
so $\norm{h}_{H^{-1}(q)}\le\norm{\psi}_{H_q}$. If $\psi\neq0$, choosing $\varphi=-\psi/\norm{\psi}_{H_q}$ attains equality; if $\psi=0$, both sides vanish. Therefore
\begin{equation}
\norm{h}_{H^{-1}(q)}=\norm{\psi}_{H_q}=\left(\E_q[\norm{\nabla\psi}^2]\right)^{1/2}.
\end{equation}
Squaring and using the minimum-energy identity proved in Appendix~\ref{app:proof-poisson} gives \eqref{eq:hminusone-action}, which is the second equality in \eqref{eq:poisson-action} and completes the proof of Theorem~\ref{thm:poisson-correction}.
\end{proof}

\begin{corollary}[Sharp uniform certificate]
\label{cor:full-sharp-certificate}
Under the assumptions of Theorem~\ref{thm:poisson-correction},
\begin{equation} a_{\mathrm{full}}(t,\etavec)=\inf\left\{B\ge0:\ |\mathcal E_t(f)|^2\le B\,\E_{Q_t}[\|\nabla f\|^2]\ \text{for every }f\in H_{q_t}\right\}. \label{eq:full-sharp-certificate} \end{equation}
Hence Full is the smallest uniform constant controlling squared reference-relative expectation-rate discrepancy over the admissible Sobolev observable class.
\end{corollary}

\begin{proof}
By Theorem~\ref{thm:poisson-correction} and Lemma~\ref{lem:hminusone}, $a_{\mathrm{full}}(t,\etavec)$ is the squared operator norm of the linear functional $f\mapsto\mathcal E_t(f)$ on $H_{q_t}$ equipped with the gradient-energy norm. The displayed infimum is precisely that squared norm.
\end{proof}

\begin{proposition}[Galerkin lower bound, monotonicity, and convergence]
\label{prop:galerkin-monotonicity}
Fix $(t,\etavec)$ under the assumptions of Theorem~\ref{thm:poisson-correction}, write $q=q_t^{\etavec}$ and $h=h_{t,\etavec}$, and let $V\subset H_q$ be a finite-dimensional linear subspace. Define
\begin{equation}
a_V:=\sup_{f\in V}\left\{2\E_q[hf]-\E_q[\|\nabla f\|^2]\right\}.
\label{eq:galerkin-restricted-action}
\end{equation}
If $\psi=\psi_{t,\etavec}$ is the Full Poisson potential, then
\begin{equation}
a_V=a_{\mathrm{full}}-\operatorname{dist}_{H_q}^2(-\psi,V),\qquad 0\le a_V\le a_{\mathrm{full}}.
\label{eq:galerkin-energy-gap}
\end{equation}
Hence $V_1\subseteq V_2$ implies $a_{V_1}\le a_{V_2}\le a_{\mathrm{full}}$. If $V_1\subseteq V_2\subseteq\cdots$ and $\overline{\bigcup_n V_n}^{\,H_q}=H_q$, then $a_{V_n}\uparrow a_{\mathrm{full}}$.
\end{proposition}

\begin{proof}
The weak Poisson identity gives $\E_q[hf]=-\langle\psi,f\rangle_{H_q}$, and therefore
\[
2\E_q[hf]-\|f\|_{H_q}^2=\|\psi\|_{H_q}^2-\|f+\psi\|_{H_q}^2.
\]
Since $\|\psi\|_{H_q}^2=a_{\mathrm{full}}$, taking the supremum over $f\in V$ gives \eqref{eq:galerkin-energy-gap}. Enlarging $V$ can only decrease the distance to $-\psi$, which proves monotonicity; density of the nested union makes this distance converge to zero.
\end{proof}

\begin{remark}[Tangent as a Galerkin restriction]
\label{rem:tangent-galerkin}
Fix $(t,\etavec)$, write $r=r_{t,\etavec}$, $G=G_{t,\etavec}$, and $\overline\phi=\Phivec_{\etavec}-\cvec_{\etavec}(t)$, and assume that each coordinate of $\overline\phi$ belongs to $H_q$. Testing the reference-relative continuity identity gives $\E_q[h\overline\phi]=-r$, while $\|\beta^\top\overline\phi\|_{H_q}^2=\beta^\top G\beta$. Taking $V=\operatorname{span}\{\overline\phi_1,\ldots,\overline\phi_R\}$ in Proposition~\ref{prop:galerkin-monotonicity} therefore gives $a_V=r^\top G^\dagger r=a_{\mathrm{tan}}$ and
\[
a_{\mathrm{full}}-a_{\mathrm{tan}}=\operatorname{dist}_{H_q}^2(-\psi,V)=a_{\mathrm{hid}}.
\]
Thus Tangent is the measured-observable Galerkin restriction of Full, while Proposition~\ref{prop:galerkin-monotonicity} applies to arbitrary finite potential dictionaries such as those used in the skyrmion evaluator. This is the same finite-feature energy-gap viewpoint underlying Sobolev discrepancy \citep{mroueh2018sobolev}.
\end{remark}

\begin{corollary}[Screened negative-Sobolev interpretation]
\label{cor:ub-hminusone}
Define
\begin{equation}
\norm{h}_{H^{-1}_{\kappa}(q)}:=\sup_{\varphi\in H_{q,\kappa}:\,\E_q[\|\nabla\varphi\|^2+\kappa^{-1}\varphi^2]\le1}\left|\E_q[h\varphi]\right|.
\label{eq:ub-hminusone-norm}
\end{equation}
Under the conditions of Corollary~\ref{cor:ub-full-correction},
\begin{equation}
a_{\mathrm{full}}^{\mathrm{ub}}(t,\etavec)=M_t\norm{h_{\mathrm{ub},t,\etavec}}_{H^{-1}_{\kappa}(q_t^{\etavec})}^2.
\label{eq:ub-hminusone-action}
\end{equation}
\end{corollary}

\begin{proof}
Equation \eqref{eq:ub-screened-poisson-weak} states that the functional $\varphi\mapsto\E_q[h_{\mathrm{ub}}\varphi]$ is represented in the Hilbert space $H_{q,\kappa}$ by $\psi^\star$. Cauchy--Schwarz gives an upper bound by $\norm{\psi^\star}_{H_{q,\kappa}}$, and choosing $\varphi=\psi^\star/\norm{\psi^\star}_{H_{q,\kappa}}$ when $\psi^\star\neq0$ attains equality. Corollary~\ref{cor:ub-full-correction} then gives \eqref{eq:ub-hminusone-action}.
\end{proof}

\subsection{Minimum-energy tangent correction}
\label{app:proof-tangent-solution}

For the minimum-energy Tangent problem, exact feasibility is equivalent to
\begin{equation}
r_{t,\etavec}\in\Range(G_{t,\etavec}).
\label{eq:tangent-feasibility}
\end{equation}

\begin{proposition}[General minimum-energy tangent correction]
\label{prop:tangent-solution}
Assume $J_{\Phivec_{\etavec}}\in L^2(Q_t^{\etavec};\R^{R\times d})$. Exact moment-rate matching is feasible if and only if \eqref{eq:tangent-feasibility} holds. In that case the unique minimum-energy correction is
\begin{equation}
\delta^{\mathrm{tan}}_{t,\etavec}(x)
=
-J_{\Phivec_{\etavec}}(x)^\top G_{t,\etavec}^{\dagger}r_{t,\etavec},
\label{eq:tangent-correction}
\end{equation}
and
\begin{equation}
a_{\mathrm{tan}}(t,\etavec)
=
r_{t,\etavec}^\top G_{t,\etavec}^{\dagger}r_{t,\etavec}.
\label{eq:tangent-action-closed}
\end{equation}
If \eqref{eq:tangent-feasibility} fails, the exact moment-rate constraint is infeasible; the pseudoinverse then has only a least-squares interpretation.
\end{proposition}

\begin{proof}[Proof of Proposition~\ref{prop:tangent-solution}]
Fix $(t,\etavec)$ and let
\begin{equation}
\mathcal H:=L^2(Q_t^{\etavec};\R^d)
\end{equation}
with its usual inner product. Define the bounded linear operator
\begin{equation}
L:\mathcal H\to\R^R,\qquad L\delta:=\E_{Q_t^{\etavec}}[J_{\Phivec_{\etavec}}\delta].
\label{eq:appendix-tangent-operator}
\end{equation}
Boundedness follows from Cauchy--Schwarz and the assumption $J_{\Phivec_{\etavec}}\in L^2(Q_t^{\etavec})$. Its adjoint is
\begin{equation}
L^*a=J_{\Phivec_{\etavec}}^\top a,\qquad a\in\R^R,
\label{eq:appendix-tangent-adjoint}
\end{equation}
because
\begin{equation}
\langle L\delta,a\rangle_{\R^R}=\E_{Q_t^{\etavec}}\!\left[\delta^\top J_{\Phivec_{\etavec}}^\top a\right].
\end{equation}
Moreover,
\begin{equation}
LL^*a=G_{t,\etavec}a.
\label{eq:appendix-LLstar}
\end{equation}
Since the codomain is finite dimensional,
\begin{equation}
\Range(L)=\Range(LL^*)=\Range(G_{t,\etavec}).
\label{eq:appendix-range-identity}
\end{equation}
Thus the condition $r_{t,\etavec}\in\Range(G_{t,\etavec})$ is exactly the feasibility condition for $L\delta=-r_{t,\etavec}$.

The feasible set is a closed affine subspace of the Hilbert space $\mathcal H$, so it contains a unique element of minimum norm. This minimum-norm element is orthogonal to $\ker L$ and therefore belongs to
\begin{equation}
(\ker L)^\perp=\overline{\Range(L^*)}.
\end{equation}
Because $L^*$ has finite-dimensional domain, its range is closed, so the minimizer has the form $\delta=L^*a$. The constraint becomes
\begin{equation}
G_{t,\etavec}a=-r_{t,\etavec}.
\end{equation}
The minimum-norm coefficient may be chosen as
\begin{equation}
a=-G_{t,\etavec}^{\dagger}r_{t,\etavec},
\end{equation}
which gives \eqref{eq:tangent-correction}. Its squared norm is
\begin{align}
\E_{Q_t^{\etavec}}[\norm{\delta^{\mathrm{tan}}_{t,\etavec}}^2]&=(G_{t,\etavec}^{\dagger}r_{t,\etavec})^\top G_{t,\etavec}(G_{t,\etavec}^{\dagger}r_{t,\etavec})\nonumber\\
&=r_{t,\etavec}^\top G_{t,\etavec}^{\dagger}r_{t,\etavec},
\end{align}
where the last equality uses symmetry of $G_{t,\etavec}$ and $r_{t,\etavec}\in\Range(G_{t,\etavec})$. This proves \eqref{eq:tangent-action-closed}. If $r_{t,\etavec}\notin\Range(G_{t,\etavec})=\Range(L)$, no exact solution exists, establishing the remaining feasibility statement.
\end{proof}

\begin{corollary}[Minimum-energy unbalanced tangent correction]
\label{cor:ub-tangent-solution}
Let $\bar c_t=\int\phi\,d\mu_t$ denote the raw finite-measure moment vector, and assume $\phi\in L^2(q_t;\R^R)$ and $J_\phi\in L^2(q_t;\R^{R\times d})$. Define
\begin{equation}
r^{\mathrm{ub}}_{t,\etavec}:=\E_{q_t}\!\left[J_\phi u_t+g_{\mathrm{ref}}\phi\right]-\frac{\dot{\bar c}_t}{M_t}
\label{eq:ub-tangent-r}
\end{equation}
and
\begin{equation}
G^{\mathrm{ub}}_{t,\etavec}:=\E_{q_t}\!\left[J_\phi J_\phi^\top+\frac{1}{\kappa}\phi\phi^\top\right].
\label{eq:ub-tangent-gram}
\end{equation}
If $r^{\mathrm{ub}}_{t,\etavec}\in\Range(G^{\mathrm{ub}}_{t,\etavec})$, then the minimum of
\begin{equation}
M_t\E_{q_t}[\|\delta\|^2+\kappa\alpha^2]
\end{equation}
subject to exact raw-moment-rate matching
\begin{equation}
\E_{q_t}[J_\phi\delta+\phi\alpha]=-r^{\mathrm{ub}}_{t,\etavec}
\label{eq:ub-tangent-constraint}
\end{equation}
is attained uniquely by
\begin{equation}
\delta^{\mathrm{tan}}_{\mathrm{ub}}=-J_\phi^\top(G^{\mathrm{ub}})^\dagger r^{\mathrm{ub}},\qquad \alpha^{\mathrm{tan}}_{\mathrm{ub}}=-\frac{1}{\kappa}\phi^\top(G^{\mathrm{ub}})^\dagger r^{\mathrm{ub}},
\label{eq:ub-tangent-correction}
\end{equation}
and
\begin{equation}
a_{\mathrm{tan}}^{\mathrm{ub}}(t,\etavec)=M_t(r^{\mathrm{ub}}_{t,\etavec})^\top(G^{\mathrm{ub}}_{t,\etavec})^\dagger r^{\mathrm{ub}}_{t,\etavec}.
\label{eq:ub-tangent-action}
\end{equation}
If $r^{\mathrm{ub}}\notin\Range(G^{\mathrm{ub}})$, exact raw-moment-rate matching is infeasible.
\end{corollary}

\begin{proof}
For any sufficiently regular finite-measure path satisfying
\begin{equation}
\partial_t\mu+\nabla\!\cdot(\mu v)=\mu g,
\end{equation}
integration by parts gives
\begin{equation}
\frac{d}{dt}\int\phi\,d\mu=\int(J_\phi v+\phi g)\,d\mu.
\label{eq:ub-moment-rate-identity}
\end{equation}
Applying this with $v=u+\delta$, $g=g_{\mathrm{ref}}+\alpha$, and $\mu=Mq$ yields \eqref{eq:ub-tangent-constraint}. Now equip
\begin{equation}
\mathcal Z:=L^2(q;\R^d)\times L^2(q)
\end{equation}
with
\begin{equation}
\langle(\delta,\alpha),(\delta',\alpha')\rangle_{\mathcal Z}:=\E_q[\delta\cdot\delta'+\kappa\alpha\alpha'].
\label{eq:ub-product-inner-product}
\end{equation}
Define $L_{\mathrm{ub}}:\mathcal Z\to\R^R$ by $L_{\mathrm{ub}}(\delta,\alpha)=\E_q[J_\phi\delta+\phi\alpha]$. Its adjoint is
\begin{equation}
L_{\mathrm{ub}}^*a=\left(J_\phi^\top a,\frac{1}{\kappa}\phi^\top a\right),
\end{equation}
since
\begin{equation}
\langle(\delta,\alpha),L_{\mathrm{ub}}^*a\rangle_{\mathcal Z}=a^\top\E_q[J_\phi\delta+\phi\alpha].
\end{equation}
Consequently,
\begin{equation}
L_{\mathrm{ub}}L_{\mathrm{ub}}^*a=G^{\mathrm{ub}}a.
\end{equation}
The Hilbert-space minimum-norm argument from Proposition~\ref{prop:tangent-solution} therefore applies verbatim, giving $(\delta^{\mathrm{tan}}_{\mathrm{ub}},\alpha^{\mathrm{tan}}_{\mathrm{ub}})=-L_{\mathrm{ub}}^*(G^{\mathrm{ub}})^\dagger r^{\mathrm{ub}}$ and squared product norm $(r^{\mathrm{ub}})^\top(G^{\mathrm{ub}})^\dagger r^{\mathrm{ub}}$. Multiplication by $M_t$ gives \eqref{eq:ub-tangent-action}.
\end{proof}

\begin{remark}[Making total-mass dynamics tangent-visible]
If the raw tangent observable vector is augmented by the constant coordinate $\phi_0\equiv1$, then its transport Jacobian vanishes and its tangent equation reduces to the mass-rate condition. The corresponding Gram entry is $\kappa^{-1}$, so total-mass mismatch is represented entirely through reaction. This augmentation is appropriate for the tangent criterion even though, by Remark~\ref{rem:ub-mass-channel}, the constant coordinate should not be inserted into the normalized-shape information-projection covariance.
\end{remark}

\subsection{Full--Tangent--Hidden decomposition}
\label{app:proof-action-decomposition}

Fix $(t,\etavec)$ and assume the hypotheses of Theorem~\ref{thm:poisson-correction}. Assume additionally that $J_{\Phivec_{\etavec}}\in L^2(Q_t^{\etavec})$, $J_{\Phivec_{\etavec}}u_t\in L^1(Q_t^{\etavec})$, and $\Phivec_{\etavec,j}-\cvec_{\etavec,j}(t)\in H_{q_t^{\etavec}}$ for every observable coordinate. We further assume that differentiation of the calibrated moments and weak testing with these coordinates are justified and that the relevant boundary terms vanish. These conditions ensure that the Full correction satisfies the Tangent moment-rate constraints and that the Hilbert-space projections below are well defined.

The pointwise identities underlying the balanced decomposition are
\begin{equation}
\E_{Q_t^{\etavec}}\!\left[J_{\Phivec_{\etavec}}\delta_{t,\etavec}^{\mathrm{hid}}\right]=0,
\label{eq:hidden-tangent-null}
\end{equation}
and
\begin{equation}
\E_{Q_t^{\etavec}}\!\left[\delta_{t,\etavec}^{\mathrm{tan}}\cdot\delta_{t,\etavec}^{\mathrm{hid}}\right]=0.
\label{eq:tangent-hidden-orthogonality}
\end{equation}
The hidden action density is
\begin{equation}
a_{\mathrm{hid}}(t,\etavec):=\E_{Q_t^{\etavec}}\!\left[\norm{\delta_{t,\etavec}^{\mathrm{hid}}}^2\right]\ge0.
\label{eq:hidden-action-density}
\end{equation}
For measurable action densities, define $\Act_{\mathrm{hid}}(\etavec):=\int_0^1a_{\mathrm{hid}}(t,\etavec)\,\rho(dt)$. The design-level counterpart of \eqref{eq:action-orthogonal-decomposition} is
\begin{equation}
\Act(\etavec)
=
\Act_{\mathrm{tan}}(\etavec)
+
\Act_{\mathrm{hid}}(\etavec),
\qquad
\Act_{\mathrm{tan}}(\etavec)\le\Act(\etavec).
\label{eq:design-action-decomposition}
\end{equation}

\begin{proof}[Proof of Proposition~\ref{prop:action-decomposition}]
Fix $(t,\etavec)$ and write $Q=Q_t^{\etavec}$. Let
\begin{equation}
L\delta:=\E_Q[J_{\Phivec_{\etavec}}\delta]
\end{equation}
be the moment-rate operator introduced implicitly in the proof of Proposition~\ref{prop:tangent-solution}. Let $\delta_t^\star$ denote the minimum-energy Full correction. Because $u_t+\delta_t^\star$ transports the projected law and the calibration identity gives $\E_Q[\Phivec_{\etavec}]=\cvec_{\etavec}(t)$, the moment-rate identity \eqref{eq:moment-rate-identity} yields
\begin{equation}
L\delta_t^\star=-\left(\E_Q[J_{\Phivec_{\etavec}}u_t]-\dot\cvec_{\etavec}(t)\right)=-r_{t,\etavec}.
\label{eq:appendix-full-implies-tangent}
\end{equation}
Hence the tangent constraint is feasible. By Proposition~\ref{prop:tangent-solution}, its unique minimum-norm solution is $\delta_{t,\etavec}^{\mathrm{tan}}$.

Define
\begin{equation}
\delta_{t,\etavec}^{\mathrm{hid}}:=\delta_t^\star-\delta_{t,\etavec}^{\mathrm{tan}}.
\end{equation}
Both $\delta_t^\star$ and $\delta_{t,\etavec}^{\mathrm{tan}}$ satisfy $L\delta=-r_{t,\etavec}$, so
\begin{equation}
L\delta_{t,\etavec}^{\mathrm{hid}}=0.
\label{eq:appendix-hidden-kernel}
\end{equation}
Thus $\delta_{t,\etavec}^{\mathrm{hid}}\in\ker L$, which proves \eqref{eq:hidden-tangent-null}.

The proof of Proposition~\ref{prop:tangent-solution} also shows that the minimum-norm tangent solution belongs to $(\ker L)^\perp$. Therefore
\begin{equation}
\E_Q\!\left[\delta_{t,\etavec}^{\mathrm{tan}}\cdot\delta_{t,\etavec}^{\mathrm{hid}}\right]=0,
\end{equation}
which is \eqref{eq:tangent-hidden-orthogonality}. Applying the Pythagorean identity in $L^2(Q;\R^d)$ gives
\begin{align}
a_{\mathrm{full}}(t,\etavec)&=\E_Q[\norm{\delta_t^\star}^2]\nonumber\\
&=\E_Q\!\left[\norm{\delta_{t,\etavec}^{\mathrm{tan}}+\delta_{t,\etavec}^{\mathrm{hid}}}^2\right]\nonumber\\
&=\E_Q[\norm{\delta_{t,\etavec}^{\mathrm{tan}}}^2]+\E_Q[\norm{\delta_{t,\etavec}^{\mathrm{hid}}}^2]\nonumber\\
&=a_{\mathrm{tan}}(t,\etavec)+a_{\mathrm{hid}}(t,\etavec),
\end{align}
proving \eqref{eq:action-orthogonal-decomposition}. Integrating the pointwise identity against the common nonnegative measure $\rho$ gives \eqref{eq:design-action-decomposition} whenever the terms are measurable; the stated pointwise and integrated inequalities follow from nonnegativity of the hidden action. At fixed $t$, equality $a_{\mathrm{tan}}(t,\etavec)=a_{\mathrm{full}}(t,\etavec)$ holds if and only if $\delta_{t,\etavec}^{\mathrm{hid}}=0$ in $L^2(Q;\R^d)$, equivalently $\delta_t^\star=\delta_{t,\etavec}^{\mathrm{tan}}$ $Q$-almost everywhere. For the integrated actions, if \(\Act(\etavec)<\infty\), then \(\Act_{\mathrm{tan}}(\etavec)=\Act(\etavec)\) holds if and only if \(a_{\mathrm{hid}}(t,\etavec)=0\) for \(\rho\)-almost every \(t\), equivalently \(\delta_{t,\etavec}^{\mathrm{hid}}=0\) in \(L^2(Q_t^{\etavec};\R^d)\) for \(\rho\)-almost every \(t\). The decomposition remains valid for measurable nonnegative extended-valued actions, but equality of two infinite actions does not imply that the hidden action vanishes.
\end{proof}

\begin{corollary}[Unbalanced Full--Tangent--Hidden decomposition]
\label{cor:ub-action-decomposition}
Under the assumptions of Corollaries~\ref{cor:ub-full-correction} and \ref{cor:ub-tangent-solution}, assume additionally that each coordinate of \(\phi=\Phivec_{\etavec}\) belongs to \(H_{q_t,\kappa}\), that \(J_\phi u_t\in L^1(q_t;\R^R)\), and that raw-moment differentiation and weak testing with these coordinates are justified, with vanishing boundary terms. Let
\begin{equation}
z^\star_{\mathrm{ub}}:=(\delta^\star,\alpha^\star),\qquad z^{\mathrm{tan}}_{\mathrm{ub}}:=(\delta^{\mathrm{tan}}_{\mathrm{ub}},\alpha^{\mathrm{tan}}_{\mathrm{ub}}),\qquad z^{\mathrm{hid}}_{\mathrm{ub}}:=z^\star_{\mathrm{ub}}-z^{\mathrm{tan}}_{\mathrm{ub}}.
\label{eq:ub-hidden-pair}
\end{equation}
Then
\begin{equation}
L_{\mathrm{ub}}z^{\mathrm{hid}}_{\mathrm{ub}}=0
\label{eq:ub-hidden-null}
\end{equation}
and
\begin{equation}
\E_q\!\left[\delta^{\mathrm{tan}}_{\mathrm{ub}}\cdot\delta^{\mathrm{hid}}_{\mathrm{ub}}+\kappa\alpha^{\mathrm{tan}}_{\mathrm{ub}}\alpha^{\mathrm{hid}}_{\mathrm{ub}}\right]=0.
\label{eq:ub-tangent-hidden-orthogonality}
\end{equation}
Defining
\begin{equation}
a_{\mathrm{hid}}^{\mathrm{ub}}(t,\etavec):=M_t\E_q\!\left[\|\delta^{\mathrm{hid}}_{\mathrm{ub}}\|^2+\kappa(\alpha^{\mathrm{hid}}_{\mathrm{ub}})^2\right],
\label{eq:ub-hidden-action}
\end{equation}
one has
\begin{equation}
a_{\mathrm{full}}^{\mathrm{ub}}(t,\etavec)=a_{\mathrm{tan}}^{\mathrm{ub}}(t,\etavec)+a_{\mathrm{hid}}^{\mathrm{ub}}(t,\etavec),
\qquad
a_{\mathrm{tan}}^{\mathrm{ub}}(t,\etavec)\le a_{\mathrm{full}}^{\mathrm{ub}}(t,\etavec).
\label{eq:ub-action-decomposition}
\end{equation}
Whenever the three terms are integrable against the common nonnegative measure $\rho$, the same decomposition and inequality hold after integration in time.
\end{corollary}

\begin{proof}
The Full correction satisfies the complete finite-measure continuity equation, so the raw moment identity \eqref{eq:ub-moment-rate-identity} implies that $z^\star_{\mathrm{ub}}$ satisfies the tangent constraint. The tangent correction is the unique minimum-norm point of that affine constraint in the product Hilbert space \eqref{eq:ub-product-inner-product}; therefore $z^{\mathrm{tan}}_{\mathrm{ub}}\in(\ker L_{\mathrm{ub}})^\perp$. Since both $z^\star_{\mathrm{ub}}$ and $z^{\mathrm{tan}}_{\mathrm{ub}}$ have the same image under $L_{\mathrm{ub}}$, their difference lies in $\ker L_{\mathrm{ub}}$. This proves \eqref{eq:ub-hidden-null} and \eqref{eq:ub-tangent-hidden-orthogonality}. Pythagoras in the product Hilbert space, followed by multiplication by $M_t$, gives \eqref{eq:ub-action-decomposition}; integration against $\rho$ preserves the identity and inequality.
\end{proof}

When \(0<\Act(\etavec)<\infty\), we summarize the relative size of the tangent-invisible component by the hidden-action fraction
\begin{equation}
\Gamma(\etavec):=\frac{\Act_{\mathrm{hid}}(\etavec)}{\Act(\etavec)}=1-\frac{\Act_{\mathrm{tan}}(\etavec)}{\Act(\etavec)}\in[0,1].
\label{eq:hidden-action-fraction}
\end{equation}
Thus \(\Gamma(\etavec)\) near one indicates that most of the minimum law-level correction lies outside the directions visible through the measured moment rates. In the discretized experimental audits we denote the corresponding common-raster quantity by \(\Gamma_h\).

\subsection{Exact condition for Tangent--Full equality}
\label{app:proof-tangent-full-equality}

\begin{corollary}[When Tangent and Full corrections coincide]
\label{cor:tangent-full-equality}
Under the conditions of Proposition~\ref{prop:action-decomposition}, suppose additionally that $Q_t^{\etavec}$ has a density positive almost everywhere on a connected open domain $\Omega$, and that $\psi_{t,\etavec}$ and the coordinates of $\Phivec_{\etavec}$ belong to $H^1_{\mathrm{loc}}(\Omega)$. Define
\begin{equation}
\beta_{t,\etavec}:=G_{t,\etavec}^{\dagger}r_{t,\etavec}.
\label{eq:tangent-beta}
\end{equation}
Then
\begin{equation}
a_{\mathrm{tan}}(t,\etavec)=a_{\mathrm{full}}(t,\etavec)
\label{eq:tangent-full-equality}
\end{equation}
if and only if
\begin{equation}
\psi_{t,\etavec}(x)=\beta_{t,\etavec}^\top\!\left(\Phivec_{\etavec}(x)-\cvec_{\etavec}(t)\right)\qquad Q_t^{\etavec}\text{-a.e.}
\label{eq:observable-span-equality}
\end{equation}
\end{corollary}

\begin{proof}[Proof of Corollary~\ref{cor:tangent-full-equality}]
Let
\begin{equation}
\beta_{t,\etavec}=G_{t,\etavec}^{\dagger}r_{t,\etavec}.
\end{equation}
By Proposition~\ref{prop:tangent-solution},
\begin{equation}
\delta_{t,\etavec}^{\mathrm{tan}}=-J_{\Phivec_{\etavec}}^\top\beta_{t,\etavec}=-\nabla\!\left(\beta_{t,\etavec}^\top\Phivec_{\etavec}\right)
\end{equation}
where the final identity is understood in the weak sense. Theorem~\ref{thm:poisson-correction} gives
\begin{equation}
\delta_t^\star=-\nabla\psi_{t,\etavec}.
\end{equation}
By Proposition~\ref{prop:action-decomposition}, equality of tangent and full action is equivalent to $\delta_t^\star=\delta_{t,\etavec}^{\mathrm{tan}}$ $Q_t^{\etavec}$-almost everywhere. Because the density of $Q_t^{\etavec}$ is positive almost everywhere on $\Omega$, this is equivalent to
\begin{equation}
\nabla\left(\psi_{t,\etavec}-\beta_{t,\etavec}^\top\Phivec_{\etavec}\right)=0
\end{equation}
almost everywhere on $\Omega$. A Sobolev function with zero weak gradient on a connected open domain is almost everywhere constant. Hence
\begin{equation}
\psi_{t,\etavec}(x)=\beta_{t,\etavec}^\top\Phivec_{\etavec}(x)+C
\end{equation}
for some constant $C$. The mean-zero gauge for the Poisson potential and the calibration identity imply
\begin{equation}
0=\E_{Q_t^{\etavec}}[\psi_{t,\etavec}]=\beta_{t,\etavec}^\top\cvec_{\etavec}(t)+C,
\end{equation}
so $C=-\beta_{t,\etavec}^\top\cvec_{\etavec}(t)$, giving \eqref{eq:observable-span-equality}.

Conversely, if \eqref{eq:observable-span-equality} holds, then
\begin{equation}
-\nabla\psi_{t,\etavec}=-J_{\Phivec_{\etavec}}^\top\beta_{t,\etavec}=\delta_{t,\etavec}^{\mathrm{tan}}
\end{equation}
$Q_t^{\etavec}$-almost everywhere. Thus the hidden component vanishes, and Proposition~\ref{prop:action-decomposition} gives $a_{\mathrm{full}}(t,\etavec)=a_{\mathrm{tan}}(t,\etavec)$.
\end{proof}

\begin{corollary}[Exact unbalanced tangent--Full equality condition]
\label{cor:ub-tangent-full-equality}
Under the assumptions of Corollary~\ref{cor:ub-action-decomposition}, assume that the screened-Poisson potential $\psi$ and the coordinates of $\phi$ belong to $H_{q,\kappa}$, and define
\begin{equation}
\beta^{\mathrm{ub}}:=(G^{\mathrm{ub}})^\dagger r^{\mathrm{ub}}.
\label{eq:ub-tangent-beta}
\end{equation}
Then
\begin{equation}
a_{\mathrm{tan}}^{\mathrm{ub}}(t,\etavec)=a_{\mathrm{full}}^{\mathrm{ub}}(t,\etavec)
\label{eq:ub-tangent-full-equality}
\end{equation}
if and only if
\begin{equation}
\psi=-({\beta^{\mathrm{ub}}})^\top\phi
\qquad\text{in }H_{q,\kappa},
\label{eq:ub-observable-span-equality}
\end{equation}
equivalently, the screened Full potential belongs to the uncentered linear span of the tangent observables with the coefficient fixed by the tangent normal equations.
\end{corollary}

\begin{proof}
Corollary~\ref{cor:ub-action-decomposition} gives equality if and only if the Full and Tangent correction pairs coincide. By Corollaries~\ref{cor:ub-full-correction} and \ref{cor:ub-tangent-solution},
\begin{equation}
z^\star_{\mathrm{ub}}=\left(\nabla\psi,\frac{\psi}{\kappa}\right),\qquad z^{\mathrm{tan}}_{\mathrm{ub}}=\left(-J_\phi^\top\beta^{\mathrm{ub}},-\frac{1}{\kappa}\phi^\top\beta^{\mathrm{ub}}\right).
\end{equation}
Equality of the reaction components gives $\psi=-({\beta^{\mathrm{ub}}})^\top\phi$ in $L^2(q)$; because both sides lie in $H_{q,\kappa}$, the corresponding weak gradients then agree as well, giving equality of the transport components. Conversely, \eqref{eq:ub-observable-span-equality} implies equality of both components and hence zero hidden action. Unlike the balanced case, no additive constant remains undetermined because the reaction component contains the potential itself.
\end{proof}

\subsection{Differentiability with respect to the measurement design}
\label{app:proof-design-sensitivity}

For continuous designs, the numerical objective depends on the design through the calibrated information projection, its physical-time multiplier derivative, the resulting continuity forcing, and finally the finite-dimensional Full correction solve. We first record the sensitivity of the calibration multiplier and then use it to establish local differentiability of the complete fixed numerical objective.

Define the calibration residual
\begin{equation}
F(\lambdavec,\etavec,t):=\E_{q_{\lambdavec,\etavec,t}}[\Phivec_{\etavec}]-\cvec_{\etavec}(t),
\label{eq:calibration-residual}
\end{equation}
where
\begin{equation}
q_{\lambdavec,\etavec,t}(x)\propto\widetilde q_t(x)\exp\!\left(\lambdavec^\top\Phivec_{\etavec}(x)\right).
\label{eq:design-tilted-family}
\end{equation}

\begin{proposition}[Design sensitivity of the information projection]
\label{prop:design-sensitivity}
Fix \(t\) and express the design in local coordinates on an open set. Suppose \(F\) is continuously differentiable in \((\lambdavec,\etavec)\) on an open neighborhood of a calibrated point satisfying \(F(\lambdavec_{t,\etavec},\etavec,t)=0\). Assume that the observable and target maps are continuously differentiable in the design coordinates and that the differentiations of the tilted expectations below can be interchanged with integration throughout this neighborhood. If \(C_{t,\etavec}=\partial_{\lambdavec}F(\lambdavec_{t,\etavec},\etavec,t)\) is positive definite, then the calibrated multiplier defines a locally unique continuously differentiable map of the design. For every design perturbation \(\xi\) in these coordinates,
\begin{equation}
D_{\etavec}\lambdavec_{t,\etavec}[\xi]=-C_{t,\etavec}^{-1}D_{\etavec}F[\xi],
\label{eq:implicit-design-gradient}
\end{equation}
where
\begin{align}
D_{\etavec}F[\xi]&=\E_{Q_t^{\etavec}}\!\left[D_{\etavec}\Phivec_{\etavec}[\xi]\right]+\Cov_{Q_t^{\etavec}}\!\left(\Phivec_{\etavec},\lambdavec_{t,\etavec}^\top D_{\etavec}\Phivec_{\etavec}[\xi]\right)\nonumber\\
&\quad-D_{\etavec}\cvec_{\etavec}(t)[\xi].
\label{eq:implicit-design-rhs}
\end{align}
\end{proposition}

\begin{proof}[Proof of Proposition~\ref{prop:design-sensitivity}]
Fix $t$ and abbreviate
\begin{equation}
q_{\lambda,\etavec}(x)=\widetilde q_t(x)\exp\!\left(\lambda^\top\Phivec_{\etavec}(x)-A_t(\lambda,\etavec)\right),
\end{equation}
where
\begin{equation}
A_t(\lambda,\etavec)=\log\E_{\widetilde Q_t}\!\left[\exp(\lambda^\top\Phivec_{\etavec}(X))\right].
\end{equation}
The reference law is frozen with respect to $\etavec$. At a calibrated solution,
\begin{equation}
F(\lambda_{t,\etavec},\etavec,t)=0.
\end{equation}
Differentiation with respect to $\lambda$ gives
\begin{equation}
\partial_{\lambda}F=\Cov_{q_{\lambda,\etavec}}(\Phivec_{\etavec},\Phivec_{\etavec}),
\end{equation}
so at the calibrated point
\begin{equation}
\partial_{\lambda}F=C_{t,\etavec}.
\label{eq:appendix-design-dlambdaF}
\end{equation}
By assumption this matrix is positive definite and hence invertible. The implicit-function theorem therefore gives a locally differentiable multiplier map $\etavec\mapsto\lambda_{t,\etavec}$ and, for any perturbation direction $\xi$,
\begin{equation}
D_{\etavec}\lambda_{t,\etavec}[\xi]=-C_{t,\etavec}^{-1}D_{\etavec}F[\xi].
\label{eq:appendix-ift}
\end{equation}
It remains to compute the directional derivative of $F$ while holding $\lambda$ fixed.

For any scalar or vector test function $f_{\etavec}$, differentiation of a normalized exponential tilt gives the identity
\begin{equation}
D_{\etavec}\E_{q_{\lambda,\etavec}}[f_{\etavec}][\xi]=\E_{q_{\lambda,\etavec}}[D_{\etavec}f_{\etavec}[\xi]]+\Cov_{q_{\lambda,\etavec}}\!\left(f_{\etavec},\lambda^\top D_{\etavec}\Phivec_{\etavec}[\xi]\right),
\label{eq:appendix-tilt-directional-identity}
\end{equation}
provided differentiation and integration can be interchanged. Applying \eqref{eq:appendix-tilt-directional-identity} with $f_{\etavec}=\Phivec_{\etavec}$ and subtracting $D_{\etavec}\cvec_{\etavec}(t)[\xi]$ gives
\begin{align}
D_{\etavec}F[\xi]&=\E_{Q_t^{\etavec}}\!\left[D_{\etavec}\Phivec_{\etavec}[\xi]\right]+\Cov_{Q_t^{\etavec}}\!\left(\Phivec_{\etavec},\lambdavec_{t,\etavec}^\top D_{\etavec}\Phivec_{\etavec}[\xi]\right)\nonumber\\
&\quad-D_{\etavec}\cvec_{\etavec}(t)[\xi],
\end{align}
which is \eqref{eq:implicit-design-rhs}. Substitution into \eqref{eq:appendix-ift} proves \eqref{eq:implicit-design-gradient}.
\end{proof}

\begin{proposition}[End-to-end differentiability of the finite-dimensional Full objective]
\label{prop:end-to-end-design-differentiability}
Let the continuous design $\etavec$ vary in local coordinates on an open set, and fix numerical time nodes $t_1,\ldots,t_M$ with nonnegative quadrature weights $\omega_1,\ldots,\omega_M$. At each $t_j$, suppose the assumptions of Proposition~\ref{prop:design-sensitivity} hold, and write the physical-time multiplier equation from \eqref{eq:lambda-dot} as
\begin{equation}
C_{j,\etavec}\dot\lambdavec_{j,\etavec}=g_{j,\etavec},
\qquad
g_{j,\etavec}:=\dot\cvec_{\etavec}(t_j)-\E_{Q_{t_j}^{\etavec}}[m_{j,\etavec}]-\Cov_{Q_{t_j}^{\etavec}}\!\left(\Phivec_{\etavec},\lambdavec_{j,\etavec}^{\top}m_{j,\etavec}\right).
\label{eq:design-lambda-dot-reduced}
\end{equation}
Assume that $C_{j,\etavec}$ and $g_{j,\etavec}$ are continuously differentiable in $\etavec$ and that $C_{j,\etavec}$ remains nonsingular. Suppose further that, after the fixed gauge choice or basis restriction used by the numerical evaluator, the Full correction at $t_j$ is represented by a reduced linear system
\begin{equation}
\mathsf K_{j,\etavec}\beta_{j,\etavec}=b_{j,\etavec},
\label{eq:design-full-reduced-system}
\end{equation}
where $\mathsf K_{j,\etavec}$ is symmetric positive definite and both $\mathsf K_{j,\etavec}$ and $b_{j,\etavec}$ are continuously differentiable in $\etavec$. Define the corresponding numerical action by
\begin{equation}
\widehat a_{\mathrm{full}}(t_j,\etavec):=\beta_{j,\etavec}^{\top}\mathsf K_{j,\etavec}\beta_{j,\etavec}=b_{j,\etavec}^{\top}\mathsf K_{j,\etavec}^{-1}b_{j,\etavec},
\qquad
\widehat\Act(\etavec):=\sum_{j=1}^{M}\omega_j\widehat a_{\mathrm{full}}(t_j,\etavec).
\label{eq:design-discrete-full-action}
\end{equation}
Then $\lambdavec_{j,\etavec}$, $\dot\lambdavec_{j,\etavec}$, $\beta_{j,\etavec}$, $\widehat a_{\mathrm{full}}(t_j,\etavec)$, and $\widehat\Act(\etavec)$ are locally continuously differentiable in $\etavec$. For every design perturbation $\xi$,
\begin{equation}
D_{\etavec}\dot\lambdavec_{j,\etavec}[\xi]
=
C_{j,\etavec}^{-1}\!\left(
D_{\etavec}g_{j,\etavec}[\xi]
-
D_{\etavec}C_{j,\etavec}[\xi]\dot\lambdavec_{j,\etavec}
\right),
\label{eq:implicit-design-lambda-dot-gradient}
\end{equation}
and the instantaneous action derivative is
\begin{equation}
D_{\etavec}\widehat a_{\mathrm{full}}(t_j,\etavec)[\xi]
=
2\beta_{j,\etavec}^{\top}D_{\etavec}b_{j,\etavec}[\xi]
-
\beta_{j,\etavec}^{\top}D_{\etavec}\mathsf K_{j,\etavec}[\xi]\beta_{j,\etavec}.
\label{eq:implicit-full-action-gradient}
\end{equation}
Consequently,
\begin{equation}
D_{\etavec}\widehat\Act(\etavec)[\xi]
=
\sum_{j=1}^{M}\omega_jD_{\etavec}\widehat a_{\mathrm{full}}(t_j,\etavec)[\xi].
\label{eq:implicit-integrated-full-gradient}
\end{equation}
\end{proposition}

\begin{proof}
Proposition~\ref{prop:design-sensitivity} gives local continuous differentiability of $\lambdavec_{j,\etavec}$. Since $C_{j,\etavec}$ remains nonsingular, \eqref{eq:design-lambda-dot-reduced} gives $\dot\lambdavec_{j,\etavec}=C_{j,\etavec}^{-1}g_{j,\etavec}$, which is locally continuously differentiable, and differentiation of this identity yields \eqref{eq:implicit-design-lambda-dot-gradient}. The assumed smooth dependence of the remaining numerical quantities therefore makes $\mathsf K_{j,\etavec}$ and $b_{j,\etavec}$ continuously differentiable functions of the design. Positive definiteness of $\mathsf K_{j,\etavec}$ gives $\beta_{j,\etavec}=\mathsf K_{j,\etavec}^{-1}b_{j,\etavec}$ and hence local continuous differentiability of the reduced correction. Finally,
\[
\widehat a_{\mathrm{full}}=b^{\top}\mathsf K^{-1}b.
\]
Using $D(\mathsf K^{-1})[\xi]=-\mathsf K^{-1}D\mathsf K[\xi]\mathsf K^{-1}$ and $\beta=\mathsf K^{-1}b$ gives
\[
D\widehat a_{\mathrm{full}}[\xi]
=
2\beta^{\top}Db[\xi]
-
\beta^{\top}D\mathsf K[\xi]\beta,
\]
which proves \eqref{eq:implicit-full-action-gradient}. The finite weighted sum in \eqref{eq:design-discrete-full-action} then gives \eqref{eq:implicit-integrated-full-gradient}.
\end{proof}

\begin{remark}[Scope of the differentiability claim]
\label{rem:end-to-end-design-differentiability-scope}
Proposition~\ref{prop:end-to-end-design-differentiability} concerns the fixed finite-dimensional objective differentiated during continuous-design optimization. It does not assert differentiability of the continuum action under arbitrary perturbations of the weighted Sobolev problem, nor of discrete decisions such as calibration-feasibility rejection, support-certification thresholds, changes of numerical basis or gauge, or membership in the risk-constrained admissible set. Within a fixed numerical representation and away from such transitions, the proposition gives the end-to-end derivative through calibration, physical-time multiplier dynamics, and the Full correction solve.
\end{remark}

\begin{corollary}[Design sensitivity in the finite-measure unbalanced extension]
\label{cor:ub-design-sensitivity}
Suppose the normalized-shape information projection in Corollary~\ref{cor:ub-iprojection} is calibrated to
\begin{equation}
c_{\etavec}(t)=\frac{\bar c_{\etavec}(t)}{M_{\etavec}(t)},
\end{equation}
and suppose the assumptions of Proposition~\ref{prop:design-sensitivity} hold. Then
\begin{equation}
D_{\etavec}\lambda_{t,\etavec}[\xi]=-C_{t,\etavec}^{-1}D_{\etavec}F[\xi]
\label{eq:ub-design-lambda}
\end{equation}
with the same expression \eqref{eq:implicit-design-rhs}, but with
\begin{equation}
D_{\etavec}c_{\etavec}(t)[\xi]=\frac{1}{M_{\etavec}(t)}D_{\etavec}\bar c_{\etavec}(t)[\xi]-\frac{c_{\etavec}(t)}{M_{\etavec}(t)}D_{\etavec}M_{\etavec}(t)[\xi].
\label{eq:ub-design-c}
\end{equation}
In particular, if the mass schedule is fixed with respect to the measurement design, only the first term in \eqref{eq:ub-design-c} remains.

\end{corollary}

\section{Analytical Comparisons and Exact Information-Projected Examples}
\label{app:analytical-comparisons}

The criterion distinctions summarized in Section~\ref{subsec:theory-comparisons} are developed here at the operator level and through exact information-projected examples. We first collect the comparator definitions used in the experiments, then interpret Full through the inverse weighted transport operator, characterize the Tangent-invisible component, compare residual $L^2$ with Full through the transport spectrum, and establish separations from forward KL and static $W_2$. The final subsection proves the matched-initial stability result used in the main text. The analytical comparisons need not assume equal scientific risk. Unless stated otherwise, $\mathbb T=\R/(2\pi\mathbb Z)$, spatial boundary conditions are periodic, Wasserstein distance uses the geodesic ground metric, and integrated comparisons use $\rho(dt)=dt$ on $[0,1]$.

\subsection{Comparator definitions}
\label{app:comparator-definitions}

In the experiments, all comparator selectors use the same information-projected path $Q_t^{\etavec}$ and the same admissible set $\mathcal D_\varepsilon$ as Full; only the secondary score changes. With $J_{\Phivec_{\etavec}}$ denoting the observable Jacobian, we define
\begin{align}
a_{\mathrm{tan}}(t,\etavec)&:=\inf_{\delta\in L^2(Q_t^{\etavec};\R^d)}\left\{\E_{Q_t^{\etavec}}[\|\delta\|^2]:\E_{Q_t^{\etavec}}[J_{\Phivec_{\etavec}}(u_t+\delta)]=\dot{\cvec}_{\etavec}(t)\right\}, \label{eq:tangent-action-variational}\\
a_{\mathrm{res}}(t,\etavec)&:=\E_{Q_t^{\etavec}}[h_{t,\etavec}^2], \label{eq:residual-l2-action}\\
d_{\mathrm{KL}}(t,\etavec)&:=\KL(Q_t^{\etavec}\Vert\widetilde Q_t), \label{eq:kl-marginal-objective}\\
\Act_{W_2}(\etavec)&:=\int_0^1W_2^2(Q_t^{\etavec},\widetilde Q_t)\,\rho(dt). \label{eq:static-wasserstein-action}
\end{align}
The corresponding integrated scores are $\Act_{\mathrm{tan}}(\etavec):=\int_0^1a_{\mathrm{tan}}(t,\etavec)\,\rho(dt)$, $\Act_{\mathrm{res}}(\etavec):=\int_0^1a_{\mathrm{res}}(t,\etavec)\,\rho(dt)$, and $J_{\mathrm{KL}}(\etavec):=\int_0^1d_{\mathrm{KL}}(t,\etavec)\,\rho(dt)$. Residual $L^2$ is used when $h_{t,\etavec}$ is square integrable, and static $W_2$ when the marginals have finite second moments.

\subsection{Weighted transport operator}
\label{app:comparison-operator}

Fix $(t,\etavec)$ and write $Q=Q_t^{\etavec}$, $q=q_t^{\etavec}$, and $h=h_{t,\etavec}$. Let $L_0^2(Q)=\{f\in L^2(Q):\E_Q[f]=0\}$. For the $L^2$ operator representation, assume $h\in L_0^2(Q)$, a densely defined closed weighted Dirichlet form, and the weighted Poincar\'e inequality of Appendix~\ref{app:proof-poisson}. The associated self-adjoint operator $\mathsf L_q$ on $L_0^2(Q)$ is characterized by
\begin{equation}
\langle\varphi,\mathsf L_q\psi\rangle_{L^2(Q)}=\E_Q[\nabla\varphi\cdot\nabla\psi],\qquad \psi\in\operatorname{Dom}(\mathsf L_q),\quad \varphi\in H_q.
\label{eq:weighted-transport-operator-weak}
\end{equation}
When the strong representation is justified,
\begin{equation}
\mathsf L_q\psi=-q^{-1}\nabla\!\cdot(q\nabla\psi).
\label{eq:weighted-transport-operator-strong}
\end{equation}
The mean-zero Poisson potential satisfies $\psi=-\mathsf L_q^{-1}h$. Testing the weak Poisson equation with $\psi$ gives
\begin{equation} a_{\mathrm{full}}(t,\etavec)=\left\langle h_t,\mathsf L_{q_t}^{-1}h_t\right\rangle_{L^2(Q_t)}. \label{eq:full-operator-representation} \end{equation}
Thus Full applies the inverse weighted transport operator to the continuity forcing before measuring its quadratic size. This representation requires the stated $L^2$ operator assumptions; the more general variational characterization of Theorem~\ref{thm:poisson-correction} only requires the forcing to define a bounded functional on $H_{q_t}$, in which case the inverse is understood from $H_{q_t}^\ast$ to $H_{q_t}$ and the pairing is dual rather than an $L^2$ inner product.

\subsection{Tangent-invisible correction and lack of universal control}
\label{app:proof-tangent-no-control}

\begin{remark}[Finite-rank origin of hidden action]
\label{rem:tangent-finite-rank}
Let $\mathcal H=L^2(Q;\R^d)$ and $L\delta=\E_Q[J_{\Phivec_{\etavec}}\delta]$. Then $\operatorname{rank}(L)\le R$ and $\operatorname{Range}(L^\ast)=\{J_{\Phivec_{\etavec}}^\top b:b\in\R^R\}$. Proposition~\ref{prop:action-decomposition} identifies Tangent as the orthogonal projection of the Full correction onto this range and places Hidden in $\ker L$. If $\mathcal H$ is infinite dimensional, finite $R$ leaves an infinite-dimensional kernel; with $D$ discrete velocity degrees of freedom, $\dim\ker L\ge D-R$. This dimension count identifies unresolved directions but does not imply nonzero hidden action in every instance. The Galerkin interpretation and exact equality conditions are given in Remark~\ref{rem:tangent-galerkin} and Corollary~\ref{cor:tangent-full-equality}.
\end{remark}

\begin{theorem}[No universal tangent control]
\label{thm:tangent-no-control}
There is no universal finite $K$ such that $a_{\mathrm{full}}(t,\etavec)\le K a_{\mathrm{tan}}(t,\etavec)$ over all regular instances of the information-projected construction. There are smooth instances with $a_{\mathrm{tan}}=0<a_{\mathrm{full}}$, and smooth families with positive Tangent action for which $a_{\mathrm{full}}/a_{\mathrm{tan}}\to\infty$. The zero-versus-positive construction can be stationary, so no such universal bound holds for the integrated actions either.
\end{theorem}

\begin{example}[A stationary law with entirely hidden correction]
\label{ex:smooth-zero-tangent-positive-full}
Take $\Phi(x)=\cos x$, $a>0$, $\vartheta\ne0$, and
\begin{equation}
I_0(z)=\frac{1}{2\pi}\int_0^{2\pi}e^{z\cos x}\,dx,\qquad \widetilde q(x)=\frac{e^{-\vartheta\cos x}}{2\pi I_0(\vartheta)},\qquad u(x)=a e^{\vartheta\cos x}.
\label{eq:tangent-counterexample-reference}
\end{equation}
The reference flux $\widetilde q u=a/(2\pi I_0(\vartheta))$ is constant. Information projection onto $\E[\cos X]=0$ gives the uniform law $q=(2\pi)^{-1}$, with
\begin{equation}
a_{\mathrm{tan}}=0,\qquad a_{\mathrm{full}}=a^2\!\left[I_0(2\vartheta)-I_0(\vartheta)^2\right]>0.
\label{eq:zero-tangent-positive-full}
\end{equation}
\end{example}

\begin{proof}[Verification and proof of Theorem~\ref{thm:tangent-no-control}]
To include the positive-Tangent perturbation, fix $t_0\in(0,1)$ and set $\lambda_t=\vartheta+2\varepsilon(t-t_0)$. With $A(\alpha)=\log\E_{\widetilde Q}[e^{\alpha\cos X}]$, define
\begin{equation}
q_t(x)=\widetilde q(x)\exp\!\left(\lambda_t\cos x-A(\lambda_t)\right),\qquad c(t)=\E_{q_t}[\cos X].
\label{eq:tangent-counterexample-path}
\end{equation}
This is the exact information projection for its target moment. All densities are smooth and positive, and the scalar observable covariance is positive. At $t_0$, $A(\vartheta)=-\log I_0(\vartheta)$, so $q_{t_0}=(2\pi)^{-1}$, $c(t_0)=0$, $C_{t_0}=G_{t_0}=1/2$, and $\dot c(t_0)=\varepsilon$. Since $\E_{q_{t_0}}[-\sin X\,u(X)]=0$, one has $r_{t_0}=-\varepsilon$ and $\delta_{t_0}^{\mathrm{tan}}=-2\varepsilon\sin x$. The continuity residual and Poisson equation are
\begin{equation}
h_{t_0}(x)=2\varepsilon\cos x+u'(x),\qquad \psi''(x)=2\varepsilon\cos x+u'(x).
\label{eq:tangent-counterexample-h}
\end{equation}
Periodicity fixes the integration constant in
\begin{equation}
\psi'(x)=2\varepsilon\sin x+u(x)-\E_{q_{t_0}}[u],\qquad \delta_{t_0}^\star=-\psi'(x).
\label{eq:tangent-counterexample-psi-prime}
\end{equation}
The sine term is odd and $u-\E[u]$ is even under $x\mapsto2\pi-x$, so their inner product vanishes. Consequently,
\begin{equation}
a_{\mathrm{tan}}(t_0)=2\varepsilon^2,\qquad a_{\mathrm{full}}(t_0)=2\varepsilon^2+a^2\!\left[I_0(2\vartheta)-I_0(\vartheta)^2\right].
\label{eq:tangent-counterexample-actions}
\end{equation}
The bracket is the uniform-law variance of $e^{\vartheta\cos X}$ and is strictly positive because $\vartheta\ne0$. Setting $\varepsilon=0$ gives Example~\ref{ex:smooth-zero-tangent-positive-full} at every time. For $\varepsilon\ne0$,
\begin{equation}
\frac{a_{\mathrm{full}}(t_0)}{a_{\mathrm{tan}}(t_0)}=1+\frac{a^2[I_0(2\vartheta)-I_0(\vartheta)^2]}{2\varepsilon^2}\longrightarrow\infty\qquad\text{as }\varepsilon\to0.
\label{eq:unbounded-full-tangent-ratio}
\end{equation}
\end{proof}

\subsubsection{No universal tangent control in the unbalanced geometry}
\label{app:proof-unbalanced-tangent-no-control}

\begin{corollary}[No universal tangent control in the unbalanced geometry]
\label{cor:ub-no-tangent-control}
Fix any $\kappa>0$. There is no universal finite constant $K$ such that
\begin{equation}
a_{\mathrm{full}}^{\mathrm{ub}}(t,\etavec)\le K\,a_{\mathrm{tan}}^{\mathrm{ub}}(t,\etavec)
\label{eq:ub-no-tangent-control}
\end{equation}
for all regular finite-measure instances of the unbalanced construction. In fact, there are smooth instances with
\begin{equation}
a_{\mathrm{tan}}^{\mathrm{ub}}(t,\etavec)=0\qquad\text{and}\qquad a_{\mathrm{full}}^{\mathrm{ub}}(t,\etavec)>0.
\label{eq:ub-zero-tangent-positive-full}
\end{equation}
\end{corollary}

\begin{proof}
Use the construction in the proof of Theorem~\ref{thm:tangent-no-control}, set $M_t\equiv1$ and $g_{\mathrm{ref}}\equiv0$, and evaluate at $t=t_0$, where $q_{t_0}$ is uniform and
\begin{equation}
h_{\mathrm{ub}}(x)=2\varepsilon\cos x+u'(x).
\end{equation}
For the single observable $\Phi(x)=\cos x$, Corollary~\ref{cor:ub-tangent-solution} gives
\begin{equation}
G^{\mathrm{ub}}=\E[\sin^2X]+\frac{1}{\kappa}\E[\cos^2X]=\frac12\left(1+\frac1\kappa\right),
\end{equation}
and therefore
\begin{equation}
a_{\mathrm{tan}}^{\mathrm{ub}}(t_0)=\frac{2\kappa}{\kappa+1}\varepsilon^2.
\label{eq:ub-counterexample-tangent}
\end{equation}
For the Full problem, the screened operator on the uniform torus is
\begin{equation}
\mathcal L_\kappa:=-\partial_{xx}+\kappa^{-1}.
\end{equation}
The forcing $2\varepsilon\cos x$ is even, whereas $u'(x)$ is odd, and $\mathcal L_\kappa^{-1}$ preserves parity. Hence the two forcing components are orthogonal in the screened energy geometry. Since
\begin{equation}
\mathcal L_\kappa(\cos x)=\left(1+\frac1\kappa\right)\cos x,
\end{equation}
the even component contributes exactly
\begin{equation}
\frac{2\kappa}{\kappa+1}\varepsilon^2.
\end{equation}
The odd component contributes
\begin{equation}
c_\kappa:=\E_{q_{t_0}}\!\left[u'\,\mathcal L_\kappa^{-1}u'\right]>0,
\end{equation}
because $\mathcal L_\kappa$ is strictly positive and $u'\not\equiv0$. Thus
\begin{equation}
a_{\mathrm{full}}^{\mathrm{ub}}(t_0)=\frac{2\kappa}{\kappa+1}\varepsilon^2+c_\kappa.
\label{eq:ub-counterexample-full}
\end{equation}
At $\varepsilon=0$, \eqref{eq:ub-counterexample-tangent}--\eqref{eq:ub-counterexample-full} give \eqref{eq:ub-zero-tangent-positive-full}. For $\varepsilon\neq0$ their ratio diverges as $\varepsilon\to0$, ruling out any universal finite $K$.
\end{proof}

\subsection{Residual amplitude versus transport effort}
\label{app:proof-residual-spectral-comparison}

The residual comparator uses the full normalized residual, $a_{\mathrm{res}}=\|h\|_{L^2(Q)}^2$, rather than a finite-dimensional projection. Its difference from Full is the missing inverse transport operator. Weighted squared Liouville residuals also appear in best-fit statistical closure \citep{turkington2013optimization}; the present comparator is the plain $L^2(Q)$ norm on an observation-prescribed path, not a weighted objective that selects the path itself.

\begin{proposition}[Spectral comparison]
\label{prop:residual-spectral-comparison}
Under Appendix~\ref{app:comparison-operator}, suppose $\mathsf L_q$ admits a complete orthonormal eigenbasis $(e_j)_{j\ge1}$ on $L_0^2(Q)$, with $\mathsf L_qe_j=\nu_je_j$ and $0<\nu_1\le\nu_2\le\cdots$. If $h=\sum_jc_je_j$, then
\begin{equation}
a_{\mathrm{res}}=\sum_jc_j^2,\qquad a_{\mathrm{full}}=\sum_j\frac{c_j^2}{\nu_j},\qquad a_{\mathrm{full}}\le\nu_1^{-1}a_{\mathrm{res}}.
\label{eq:residual-full-spectral-identities}
\end{equation}
For residuals supported in a spectral band $\nu_j\le\Lambda$,
\begin{equation}
\nu_1a_{\mathrm{full}}\le a_{\mathrm{res}}\le\Lambda a_{\mathrm{full}}.
\label{eq:finite-band-equivalence}
\end{equation}
If $\nu_j\to\infty$, no finite $C$ satisfies $a_{\mathrm{res}}\le C a_{\mathrm{full}}$ for every $h\in L_0^2(Q)$.
\end{proposition}

\begin{proof}
Parseval's identity gives the residual norm. Expanding $\mathsf L_q^{-1}h$ in the same eigenbasis gives the Full quadratic form. The inequalities follow by bounding the active eigenvalues above and below. A universal reverse bound would imply $1\le C/\nu_j$ upon taking $h=e_j$, contradicting $\nu_j\to\infty$.
\end{proof}

\begin{corollary}[Pointwise ranking reversal]
\label{cor:residual-full-ranking-reversal}
If $\nu_j>\nu_\ell$ and $1<a_A/a_B<\sqrt{\nu_j/\nu_\ell}$, then $h_A=a_Ae_j$ and $h_B=a_Be_\ell$ satisfy $a_{\mathrm{res}}(h_A)>a_{\mathrm{res}}(h_B)$ but $a_{\mathrm{full}}(h_A)<a_{\mathrm{full}}(h_B)$.
\end{corollary}

\begin{proof}
The two comparisons are respectively $a_A^2>a_B^2$ and $a_A^2/\nu_j<a_B^2/\nu_\ell$.
\end{proof}

\begin{example}[Exact information-projected Fourier family]
\label{ex:mfsi-fourier-spectral-scaling}
Let $\widetilde q=q_0=(2\pi)^{-1}$ and $u=0$. For an integer $k\ge1$, define
\begin{equation}
\Phi_k(x)=\cos(kx),\qquad q_{\lambda,k}(x)=\frac{e^{\lambda\cos(kx)}}{2\pi I_0(\lambda)},\qquad c(\lambda)=\frac{d}{d\lambda}\log I_0(\lambda).
\label{eq:appendix-fourier-family}
\end{equation}
Periodicity makes the normalization and the moment $\E_{q_{\lambda,k}}[\Phi_k]=c(\lambda)$ independent of $k$. Hence $q_{\lambda,k}$ is exactly the information projection onto this moment fiber. Along $q_{k,t}=q_{\lambda_k(t),k}$,
\begin{equation}
h_{k,t}(x)=\dot\lambda_k(t)\bigl(\cos(kx)-c(\lambda_k(t))\bigr).
\label{eq:appendix-fourier-forcing}
\end{equation}
At $\lambda_k(t_0)=0$ and $\dot\lambda_k(t_0)=a$, the density is uniform, $h_{k,t_0}=a\cos(kx)$, and the mean-zero Poisson potential is $\psi(x)=-a\cos(kx)/k^2$. Therefore
\begin{equation}
a_{\mathrm{res}}(t_0,k)=\frac{a^2}{2},
\qquad
a_{\mathrm{full}}(t_0,k)=\frac{a^2}{2k^2}.
\label{eq:torus-residual-full-scaling}
\end{equation}
The pair $h_A=2\cos(3x)$ and $h_B=\cos x$ gives the exact instantaneous ranking reversal $2>1/2$ for residual $L^2$ and $2/9<1/2$ for Full.
\end{example}

\paragraph{Transport metric along the Fourier family.}
Write $q_\lambda=q_{\lambda,1}$ and define
\begin{equation}
g(\lambda)=\|\cos x-c(\lambda)\|_{H^{-1}(q_\lambda)}^2.
\label{eq:appendix-fourier-transport-metric}
\end{equation}
This function is continuous, strictly positive, and satisfies $g(0)=1/2$. An explicit verification avoids any assumption about a limiting discretization. Set $f_\lambda(x)=\cos x-c(\lambda)$ and
\begin{equation}
F_\lambda(x)=\int_0^xq_\lambda(y)f_\lambda(y)\,dy,\qquad b_\lambda=\frac{\int_0^{2\pi}F_\lambda(x)/q_\lambda(x)\,dx}{\int_0^{2\pi}1/q_\lambda(x)\,dx}.
\label{eq:appendix-fourier-flux}
\end{equation}
For a unit multiplier rate, $v_\lambda=(b_\lambda-F_\lambda)/q_\lambda$ satisfies $(q_\lambda v_\lambda)'=-q_\lambda f_\lambda$. Its spatial integral vanishes by the definition of $b_\lambda$, so it is a periodic gradient and the minimum-energy correction. Consequently,
\begin{equation}
g(\lambda)=\int_0^{2\pi}\frac{(b_\lambda-F_\lambda(x))^2}{q_\lambda(x)}\,dx.
\label{eq:appendix-fourier-metric-integral}
\end{equation}
On every compact multiplier interval, $q_\lambda$ is uniformly positive and all terms vary smoothly with $\lambda$, proving continuity. The energy cannot vanish because $f_\lambda$ is not identically zero. At $\lambda=0$, $v_0=-\sin x$, giving $g(0)=1/2$. Finally, if $\chi_\lambda$ solves $\mathsf L_{q_{\lambda,1}}\chi_\lambda=f_\lambda$, then $k^{-2}\chi_\lambda(kx)$ solves the frequency-$k$ problem. Thus the exact action along any differentiable multiplier path is
\begin{equation}
a_{\mathrm{full}}(t,k)=\frac{\dot\lambda_k(t)^2}{k^2}g(\lambda_k(t)).
\label{eq:appendix-full-frequency-scaling}
\end{equation}

\subsection{Forward KL: local geometry and exact separations}
\label{app:proof-forward-kl}

\begin{proposition}[Exact dual form and local information geometry]
\label{prop:kl-dual-local}
At fixed $(t,\etavec)$, abbreviate $A=A_{t,\etavec}$ and $\phi=\Phivec_{\etavec}$, and let $Q_\lambda$ be the exponential tilt of $\widetilde Q_t$. Write $\widetilde c=\E_{\widetilde Q_t}[\phi]$ and $\widetilde C=\Cov_{\widetilde Q_t}(\phi,\phi)$. Then
\begin{equation}
\KL(Q_\lambda\Vert\widetilde Q_t)=\lambda^\top\nabla A(\lambda)-A(\lambda).
\label{eq:kl-dual-population}
\end{equation}
At the calibrated multiplier, $\nabla A(\lambda)=\cvec_{\etavec}(t)$. If $A$ is $C^3$ near zero, then
\begin{equation}
\KL(Q_\lambda\Vert\widetilde Q_t)=\frac12\lambda^\top\widetilde C\lambda+O(\|\lambda\|^3).
\label{eq:kl-local-lambda}
\end{equation}
If also $\widetilde C\succ0$, then, for $\Delta c=\E_{Q_\lambda}[\phi]-\widetilde c$,
\begin{equation}
\KL(Q_\lambda\Vert\widetilde Q_t)=\frac12\Delta c^\top\widetilde C^{-1}\Delta c+O(\|\Delta c\|^3).
\label{eq:kl-local-moment}
\end{equation}
\end{proposition}

\begin{proof}
Integrating $\log(dQ_\lambda/d\widetilde Q_t)=\lambda^\top\phi-A(\lambda)$ gives the dual identity. Since $A(0)=0$, Taylor expansion yields $A(\lambda)=\widetilde c^\top\lambda+\tfrac12\lambda^\top\widetilde C\lambda+O(\|\lambda\|^3)$ and $\nabla A(\lambda)=\widetilde c+\widetilde C\lambda+O(\|\lambda\|^2)$. Substitution cancels the linear term. When $\widetilde C\succ0$, the inverse-function theorem gives $\lambda=\widetilde C^{-1}\Delta c+O(\|\Delta c\|^2)$, proving the moment-coordinate expansion.
\end{proof}

\paragraph{Marginal tilt versus multiplier velocity.}
For a stationary reference with $u=0$, let $\overline\phi_\lambda=\phi-\E_{Q_\lambda}[\phi]$. When the centered coordinates and inverse operator are well defined, set
\begin{equation}
[\mathsf H(\lambda)]_{ab}=\langle\overline\phi_{\lambda,a},\mathsf  L_{q_\lambda}^{-1}\overline\phi_{\lambda,b}\rangle_{L^2(Q_\lambda)}.
\label{eq:kl-full-transport-gram}
\end{equation}
The identity $h_t=\dot\lambda_t^\top\overline\phi_{\lambda_t}$ implies
\begin{equation}
a_{\mathrm{full}}(t)=\dot\lambda_t^\top\mathsf H(\lambda_t)\dot\lambda_t.
\label{eq:kl-full-stationary-family}
\end{equation}
In a regular parametrization admitting a Wasserstein pullback, $\mathsf H$ is the transport metric in natural-parameter coordinates \citep{chenli2020natural,li2023wasserstein}. It differs from both the covariance/Fisher matrix and the observable derivative Gram matrix. Marginal KL acts on tilt displacement, whereas Full acts on the velocity of the prescribed path in transport geometry.

\begin{proposition}[No universal comparison between forward KL and Full]
\label{prop:kl-full-non-equivalence}
Even with the state space $\mathbb T$ and time interval $[0,1]$ fixed, neither $d_{\mathrm{KL}}\le C a_{\mathrm{full}}$ nor $a_{\mathrm{full}}\le C d_{\mathrm{KL}}$ holds with an instance-independent finite $C$ over all regular instances of the information-projected construction. Likewise, for $\rho(dt)=dt$, neither $J_{\mathrm{KL}}\le C\Act$ nor $\Act\le C J_{\mathrm{KL}}$ holds universally.
\end{proposition}

\begin{example}[Pointwise separations]
\label{ex:kl-full-pointwise-separation}
Under the uniform reference with $u=0$, take the Fourier family in \eqref{eq:appendix-fourier-family}. A constant nonzero multiplier $\lambda_0$ gives a stationary projected law distinct from the reference: $h_t=0$, so $a_{\mathrm{full}}=0$, while $d_{\mathrm{KL}}>0$. Conversely, at $\lambda_k(t_0)=0$ and $\dot\lambda_k(t_0)=a\ne0$, the two marginals coincide but their time derivatives need not, and
\begin{equation}
d_{\mathrm{KL}}(t_0,k)=0,\qquad a_{\mathrm{full}}(t_0,k)=\frac{a^2}{2k^2}>0.
\label{eq:kl-zero-full-positive}
\end{equation}
The constant-tilt construction also gives $J_{\mathrm{KL}}>0=\Act$ after integration.
\end{example}

\begin{example}[Rapid marginal oscillation]
\label{ex:rapid-kl-full-separation}
Fix $\epsilon>0$, set $\lambda_n(t)=\epsilon\sin(2\pi nt)$, and take
\begin{equation}
q_{n,t}(x)=\frac{\exp(\lambda_n(t)\cos x)}{2\pi I_0(\lambda_n(t))},\qquad n\in\mathbb N,\quad n\ge1.
\label{eq:rapid-von-mises-path}
\end{equation}
Every path is an exact information projection for its $\cos x$ moment trajectory and satisfies $q_{n,0}=q_{n,1}=q_0$. For integer $n$,
\begin{equation}
J_{\mathrm{KL}}(n)=J_{\mathrm{KL}}(1),\qquad \Act(n)=n^2\Act(1)>0.
\label{eq:rapid-kl-full-exact-scaling}
\end{equation}
There exist constants $C_\epsilon<\infty$ and $m_\epsilon>0$, independent of $n$, such that
\begin{equation}
J_{\mathrm{KL}}(n)\le C_\epsilon\epsilon^2,\qquad \Act(n)\ge2\pi^2m_\epsilon\epsilon^2n^2.
\label{eq:rapid-kl-full-scaling}
\end{equation}
As a tie-breaking thought experiment, suppose the primary task does not distinguish these paths beyond constraints that they all satisfy, such as their common endpoints. Their time-averaged forward KL displacement from the reference is identical for every integer $n$, yet their required dynamical revision grows as $n^2$. Under the minimum-additional-commitment preference specified in the main text, Full therefore distinguishes a path that makes a modest reference-relative excursion from one that repeatedly moves back and forth through the same range, even though a static marginal criterion and the declared endpoint task do not.
\end{example}

\begin{proof}[Verification and completion of Proposition~\ref{prop:kl-full-non-equivalence}]
Let $D(\lambda)$ be the forward KL of $q_{\lambda,1}$ relative to $q_0$. This function is smooth and vanishes to second order at zero, so $D(\lambda)\le C_\epsilon\lambda^2$ on $[-\epsilon,\epsilon]$. The function $g$ in \eqref{eq:appendix-fourier-transport-metric} has a positive minimum $m_\epsilon$ on this interval. Thus
\begin{equation}
J_{\mathrm{KL}}(n)=\int_0^1D(\epsilon\sin(2\pi nt))\,dt,\qquad \Act(n)=(2\pi n\epsilon)^2\int_0^1\cos^2(2\pi nt)g(\epsilon\sin(2\pi nt))\,dt.
\label{eq:rapid-kl-full-integrals}
\end{equation}
The substitution $\theta=2\pi nt$ integrates $n$ identical periods, proving \eqref{eq:rapid-kl-full-exact-scaling}. Bounding $D$ and $g$ as above and using $\int_0^1\cos^2(2\pi nt)\,dt=1/2$ proves \eqref{eq:rapid-kl-full-scaling}. Constants may be chosen uniformly for $0<\epsilon\le\epsilon_0$ by taking bounds on a fixed compact interval $[-\epsilon_0,\epsilon_0]$. The marginal score is positive for $\epsilon>0$, so $\Act(n)/J_{\mathrm{KL}}(n)\to\infty$. Together with Example~\ref{ex:kl-full-pointwise-separation}, this proves all claims of Proposition~\ref{prop:kl-full-non-equivalence}.
\end{proof}

\begin{corollary}[Local KL/Full ranking reversal]
\label{cor:kl-full-ranking-reversal}
Let $s\in C^1([0,1])$ be nonconstant and compare $q_{A,t}=q_{2\epsilon s(t),3}$ with $q_{B,t}=q_{\epsilon s(t),1}$. Then $J_{\mathrm{KL}}(A)>J_{\mathrm{KL}}(B)$ for every $\epsilon>0$, whereas $\Act(A)<\Act(B)$ for all sufficiently small $\epsilon>0$. More precisely,
\begin{equation}
J_{\mathrm{KL}}(A)=\epsilon^2\int_0^1s(t)^2\,dt+o(\epsilon^2),\qquad J_{\mathrm{KL}}(B)=\frac{\epsilon^2}{4}\int_0^1s(t)^2\,dt+o(\epsilon^2),
\label{eq:kl-ranking-asymptotics}
\end{equation}
while
\begin{equation}
\Act(A)=\frac{2\epsilon^2}{9}\int_0^1\dot s(t)^2\,dt+o(\epsilon^2),\qquad \Act(B)=\frac{\epsilon^2}{2}\int_0^1\dot s(t)^2\,dt+o(\epsilon^2).
\label{eq:full-ranking-asymptotics}
\end{equation}
\end{corollary}

\begin{proof}
For $A(\lambda)=\log I_0(\lambda)$, $D(\lambda)=\lambda A'(\lambda)-A(\lambda)$ is even and $D'(\lambda)=\lambda\Var_{q_{\lambda,1}}(\cos X)$. Hence $D$ is strictly increasing in $|\lambda|$ away from zero. Since $s$ is nonconstant, it is nonzero on a set of positive measure, proving the strict KL ordering. The expansion $D(\lambda)=\lambda^2/4+o(\lambda^2)$ gives \eqref{eq:kl-ranking-asymptotics}. Equation~\eqref{eq:appendix-full-frequency-scaling} gives the exact actions $4\epsilon^2\dot s(t)^2g(2\epsilon s(t))/9$ and $\epsilon^2\dot s(t)^2g(\epsilon s(t))$. Continuity of $g$ at zero, $g(0)=1/2$, and boundedness of $s$ give \eqref{eq:full-ranking-asymptotics} after integration. Because $\int_0^1\dot s(t)^2\,dt>0$, the leading coefficients yield the opposite strict Full ranking for sufficiently small $\epsilon$.
\end{proof}

\subsection{Static Wasserstein displacement and initial matching}
\label{app:proof-static-wasserstein}

The comparator $\Act_{W_2}$ in \eqref{eq:static-wasserstein-action} averages static distances between projected and reference marginals. It is not a dynamic endpoint-transport objective and does not optimize or directly assess the time derivative of the prescribed projected path.

\begin{proposition}[Static Wasserstein displacement and Full are non-equivalent]
\label{prop:static-wasserstein-non-equivalence}
Over regular instances of the information-projected construction on $\mathbb T$, neither $\Act_{W_2}\le C\Act$ nor $\Act\le C\Act_{W_2}$ holds with an instance-independent finite $C$ when $\rho(dt)=dt$. The analogous pointwise inequalities for $W_2^2(Q_t^{\etavec},\widetilde Q_t)$ and $a_{\mathrm{full}}(t,\etavec)$ also fail. In the rapid-oscillation family of Example~\ref{ex:rapid-kl-full-separation},
\begin{equation}
\Act_{W_2}(n)=\Act_{W_2}(1),\qquad \Act(n)=n^2\Act(1),
\label{eq:rapid-wasserstein-full-exact-scaling}
\end{equation}
and there is $M_\epsilon<\infty$, independent of $n$, such that
\begin{equation}
\Act_{W_2}(n)\le M_\epsilon\epsilon^2,\qquad \Act(n)\ge2\pi^2m_\epsilon\epsilon^2n^2.
\label{eq:rapid-wasserstein-full-scaling}
\end{equation}
\end{proposition}

\begin{proof}
The constant nonzero tilt in Example~\ref{ex:kl-full-pointwise-separation} has zero Full action but positive Wasserstein distance from the reference, both pointwise and after integration. At a zero multiplier with nonzero multiplier derivative, the Wasserstein distance is zero while Full is positive. These give the pointwise failures and rule out $\Act_{W_2}\le C\Act$ without initial matching. For the reverse integrated bound, the time average of $W_2^2(q_{\epsilon\sin(2\pi nt),1},q_0)$ is independent of integer $n$ by periodicity, whereas Full has the exact $n^2$ scaling already proved.

For the small-amplitude bound, set $M_\epsilon=\max_{|\lambda|\le\epsilon}g(\lambda)$. Given $|\lambda|\le\epsilon$, the path $r\mapsto q_{r\lambda,1}$, $r\in[0,1]$, has minimum kinetic energy density $\lambda^2g(r\lambda)$. Using this path as an admissible competitor in dynamic optimal transport gives
\begin{equation}
W_2^2(q_{\lambda,1},q_0)\le\int_0^1\lambda^2g(r\lambda)\,dr\le M_\epsilon\lambda^2.
\label{eq:appendix-static-w2-bound}
\end{equation}
Substitution of $\lambda=\lambda_n(t)$ and integration prove \eqref{eq:rapid-wasserstein-full-scaling}. The constant can be chosen uniformly for $0<\epsilon\le\epsilon_0$ by using the maximum over $[-\epsilon_0,\epsilon_0]$.
\end{proof}

\begin{proposition}[Wasserstein stability relative to a matched-initial reference rollout]
\label{prop:matched-initial-wasserstein-control}

Let $\mathcal X=\R^d$ or a flat torus, take $\rho(dt)=dt$, and assume $Q_t^{\etavec},R_t^{\etavec}\in\mathcal P_2(\mathcal X)$ and $\Act(\etavec)<\infty$. Let $\mathsf T^u_{0,t}$ denote the characteristic flow generated by the frozen reference velocity $u_t$, and define the matched-initial reference rollout
\begin{equation}
R_t^{\etavec}:=(\mathsf T^u_{0,t})_\#Q_0^{\etavec}.
\end{equation}
Suppose $u_t$ and $u_t+\delta_t^\star$ admit classical characteristic flows realizing $R_t^{\etavec}$ and $Q_t^{\etavec}$, respectively, and assume $\|u_t(x)-u_t(y)\|\le L(t)\|x-y\|$ with $L\ge0$ integrable, imposed on periodic lifts for a torus. With $\Lambda=\int_0^1L(t)\,dt$,
\begin{equation} W_2(Q_t^{\etavec},R_t^{\etavec})\le\int_0^t\exp\!\left(\int_s^tL(\tau)\,d\tau\right)\sqrt{a_{\mathrm{full}}(s,\etavec)}\,ds, \label{eq:matched-initial-wasserstein-pointwise} \end{equation}
and consequently
\begin{equation} \int_0^1W_2^2(Q_t^{\etavec},R_t^{\etavec})\,dt\le e^{2\Lambda}\Act(\etavec). \label{eq:matched-initial-wasserstein-action} \end{equation}

If additionally $Q_0^{\etavec}=\widetilde Q_0$, then $R_t^{\etavec}=\widetilde Q_t$ and the left-hand side equals $\Act_{W_2}(\etavec)$ as defined in \eqref{eq:static-wasserstein-action}.

Moreover, for every globally $1$-Lipschitz observable $f$,
\begin{equation} \left|\E_{Q_t^{\etavec}}[f]-\E_{R_t^{\etavec}}[f]\right|\le W_1(Q_t^{\etavec},R_t^{\etavec})\le W_2(Q_t^{\etavec},R_t^{\etavec}). \label{eq:matched-initial-lipschitz-observable-control} \end{equation}
Since $\int_0^t a_{\mathrm{full}}(s,\etavec)\,ds\le\Act(\etavec)$ and $\int_s^tL(\tau)\,d\tau\le\Lambda$, the pointwise bound also gives
\begin{equation} \sup_{\operatorname{Lip}(f)\le1}\left|\E_{Q_t^{\etavec}}[f]-\E_{R_t^{\etavec}}[f]\right|\le e^\Lambda\sqrt{\Act(\etavec)},\qquad t\in[0,1]. \label{eq:appendix-matched-initial-observable-stability} \end{equation}
Thus accumulated Full action controls a one-sided reference-relative stability envelope for Lipschitz observables after removing initial-law mismatch.
\end{proposition}

\begin{proof}
Couple the two characteristic flows from a common initial random state $X_0=Y_0\sim Q_0^{\etavec}$, with $\dot X_t=u_t(X_t)+\delta_t^\star(X_t)$ and $\dot Y_t=u_t(Y_t)$. Then $\mathcal L(X_t)=Q_t^{\etavec}$ and $\mathcal L(Y_t)=R_t^{\etavec}$. On a torus, use lifts with the same initial representative; lifted Euclidean distance bounds geodesic distance. Gronwall's inequality gives
\begin{equation} \|X_t-Y_t\|\le\int_0^t\exp\!\left(\int_s^tL(\tau)\,d\tau\right)\|\delta_s^\star(X_s)\|\,ds. \end{equation}
The $L^2$ integral Minkowski inequality and $\E[\|\delta_s^\star(X_s)\|^2]=a_{\mathrm{full}}(s,\etavec)$ give \eqref{eq:matched-initial-wasserstein-pointwise}. Cauchy--Schwarz yields $W_2^2(Q_t^{\etavec},R_t^{\etavec})\le e^{2\Lambda}t\int_0^ta_{\mathrm{full}}(s,\etavec)\,ds\le e^{2\Lambda}\Act(\etavec)$, and integration in $t$ proves \eqref{eq:matched-initial-wasserstein-action}. Finally, the Kantorovich--Rubinstein bound gives the first inequality in \eqref{eq:matched-initial-lipschitz-observable-control}, while $W_1\le W_2$ follows from the finite second-moment assumption.
\end{proof}

The constant-tilt example has $Q_0^{\etavec}\neq\widetilde Q_0$, so it does not contradict the special-case bound relative to the original frozen reference marginals. Relative to the matched-initial rollout $R_t^{\etavec}$, the constant-tilt path has zero Full action and zero matched-initial Wasserstein departure, consistently with Proposition~\ref{prop:matched-initial-wasserstein-control}. When $Q_0^{\etavec}=\widetilde Q_0$, one has $R_t^{\etavec}=\widetilde Q_t$, and the proposition specializes to a bound on static Wasserstein displacement from the original reference path. The oscillatory family satisfies this initial equality and still rules out a reverse estimate. Likewise, the spectral equivalence in \eqref{eq:finite-band-equivalence} requires common positive lower and finite upper spectral bounds to obtain constants uniform across designs. These statements are independent of the scientific-risk constraint.

\section{Additional Experiments and Ablations}
\label{app:validation-ablations}

Shared experimental conventions and reproducibility details are provided in Appendix~\ref{app:experiments}. Supporting-study-specific protocols are given below where needed.

\subsection{Additional Experiments}
\label{app:additional-experiments}

We report two additional supporting experiments. The retrospective double-gyre study tests repeatability of the Full-versus-Law contrast across fresh observations and independently trained fits within the same declared reference-learning regime. The finite-measure study extends FIDE beyond mass-preserving probability laws by allowing reference-relative correction through both transport and reaction.

\subsubsection{Retrospective double-gyre repeatability study}
\label{app:exp-vortices}

\paragraph{System and principal results.} The study uses the double-gyre dynamics defined in Appendix~\ref{app:exp-vortices-prospective}, with four Gaussian sensors observing aggregate particle responses at nine acquisition times. Selection uses a frozen microscopic target simulation bank together with three independently seeded endpoint-trained references fitted from the same physical endpoint dataset. The study therefore asks a narrow question: does the Full-versus-Law action reduction repeat across fresh observations and across independent fits within the same declared reference-learning regime? On $64$ shared fresh trials per reference, the mean reference-wise reduction ranges from $9.96\%$ at the $0.5\%$ allowance to $34.34\%$ at $5\%$, and simultaneous $95\%$ paired-bootstrap intervals remain above zero across all $18$ reference-by-allowance contrasts. Held-out scientific-risk point estimates remain close to their nominal allowances, although matching simultaneous risk-feasibility inference is unavailable because the retained evaluation summaries do not contain paired per-trial Law and Full scientific losses. These results support finite-sample and within-regime reference-fit repeatability; they do not establish transfer to an unseen reference family, arbitrary reference misspecification, or independent physical endpoint datasets.

\paragraph{Protocol and numerical evaluation.} The truth bank contains $50{,}000$ particles. Each of the three endpoint-trained flow-matching references uses four width-$128$ hidden layers, $12{,}000$ Adam steps, and $32{,}768$ rollout particles. Moment reconstruction uses endpoint-anchored bounded cubic penalized least squares with three internal knots, smoothing $10^{-4}$, and relative ridge $10^{-10}$. Scientific risk averages four MMD bandwidths across the three references; the frozen Law baseline is $0.0386$ and the population-risk ceiling is $0.0634$. The shared selection bank contains $128$ trials, and the final numerical Full evaluator uses a $256\times128$ raster at all $21$ scientific times. After the six-point frontier is frozen, evaluation uses $64$ shared fresh trials per reference. Simultaneous $95\%$ action intervals use $100{,}000$ paired bootstrap resamples over all $18$ reference-by-allowance contrasts.

\begin{table}[t]
\centering
\small
\begin{tabular}{rrrr}
\toprule
Allowance & Full holdout risk & Risk increase (\%) & Excess over allowance (pp) \\
\midrule
$0.5\%$ & 0.038777 & 0.4404 & $-0.0596$ \\
$1\%$ & 0.039016 & 1.0598 & $+0.0598$ \\
$2\%$ & 0.039289 & 1.7686 & $-0.2314$ \\
$3\%$ & 0.039531 & 2.3945 & $-0.6055$ \\
$4\%$ & 0.039950 & 3.4794 & $-0.5206$ \\
$5\%$ & 0.040538 & 5.0023 & $+0.0023$ \\
\bottomrule
\end{tabular}
\caption{\textbf{Held-out scientific-risk transfer for the retrospective frontier.} The holdout Law risk is $0.038607$. Negative excess indicates a point estimate below the nominal allowance. Values are descriptive because the retained summaries do not contain paired per-trial scientific losses needed for matching simultaneous risk-feasibility inference.}
\label{tab:retrospective-heldout-risk}
\end{table}

\subsubsection{Finite-measure transport--reaction experiment}
\label{app:unbalanced-experiment}

We consider an unbalanced Gaussian transport problem in which the evolving object is a finite measure rather than a probability law. FIDE acts on the normalized shape through the same information-projection construction used in the balanced experiments, while discrepancies in total mass are represented through the reaction channel of the screened transport--reaction correction. As in the balanced setting, Law provides the scientific-risk reference design and Tangent optimizes only the correction visible through the measured raw moment rates. Full instead optimizes the unrestricted transport--reaction correction in \eqref{eq:ub-full-action-definition}. We additionally introduce GlobalReaction as an intermediate structural comparator. It retains an unrestricted transport correction but restricts the reaction field to a single spatially constant rate at each time:
\begin{equation}
a_{\mathrm{GR}}^{\mathrm{ub}}(t,\etavec)
:=
M_t\inf_{\delta,\;a\in\mathbb R}
\left\{
\E_{q_t}\!\left[\|\delta\|^2+\kappa a^2\right]:
-\nabla\!\cdot(q_t\delta)+q_ta=q_th_{\mathrm{ub}}
\right\}.
\label{eq:ub-global-reaction-action}
\end{equation}
Integrating the constraint in \eqref{eq:ub-global-reaction-action} over space fixes \(a=\E_{q_t}[h_{\mathrm{ub}}]=\dot M_t/M_t-g_{\mathrm{ref}}(t)\). GlobalReaction therefore permits reaction to correct the global mass-rate discrepancy, while any spatially heterogeneous mean-zero forcing must be realized through transport. Full removes this restriction by allowing the optimal reaction field to vary over space, whereas Tangent imposes the distinct restriction of matching only the measured raw moment rates.

The primary experiment fixes $\kappa=1$ before validation and evaluates the frozen Law, Tangent, GlobalReaction, and Full geometries on the same $128$ fresh paired trials across the five prespecified risk allowances. Table~\ref{tab:unbalanced-primary-comparison} cross-evaluates every selected geometry using the common Full-action functional. Tangent and GlobalReaction both reduce the common Full cross-score relative to Law, while the Full-selected geometry yields the lowest Full cross-score at every operating point. The Full-versus-Law reduction increases from $3.26\%$ at the $1\%$ allowance to $9.79\%$ at the $5\%$ allowance, with positive lower endpoints of the paired action intervals throughout. At the $5\%$ operating point, the Tangent- and GlobalReaction-selected geometries reduce the same Full cross-score by $3.72\%$ and $4.03\%$, respectively, compared with $9.79\%$ for the Full-selected geometry. These cross-scores quantify the unrestricted transport--reaction robustness consequence of the different structural restrictions; they do not rank Tangent or GlobalReaction on their native objectives. Tangent and GlobalReaction are used here as descriptive structural comparators, and the reported action inference is for the prespecified Full-versus-Law contrast.

\begin{table}[t]
\centering
\small
\setlength{\tabcolsep}{3.8pt}
\resizebox{\linewidth}{!}{%
\begin{tabular}{rrrrrrrrr}
\toprule
& \multicolumn{4}{c}{Mean Full action} & \multicolumn{3}{c}{Reduction vs.\ Law (\%)} & \\
\cmidrule(lr){2-5}\cmidrule(lr){6-8}
Allowance & Law & Tangent & GlobalReaction & Full & Tangent & GlobalReaction & Full & Full interval (\%) \\
\midrule
$1\%$ & 0.77523 & 0.76973 & 0.76652 & 0.74994 & 0.71 & 1.12 & 3.26 & $[1.50,\ 5.03]$ \\
$2\%$ & 0.77523 & 0.76502 & 0.76205 & 0.73475 & 1.32 & 1.70 & 5.22 & $[3.46,\ 6.99]$ \\
$3\%$ & 0.77523 & 0.75927 & 0.75660 & 0.72193 & 2.06 & 2.40 & 6.88 & $[5.11,\ 8.64]$ \\
$4\%$ & 0.77523 & 0.75296 & 0.75043 & 0.71029 & 2.87 & 3.20 & 8.38 & $[6.61,\ 10.14]$ \\
$5\%$ & 0.77523 & 0.74635 & 0.74397 & 0.69936 & 3.72 & 4.03 & 9.79 & $[8.02,\ 11.55]$ \\
\bottomrule
\end{tabular}%
}
\caption{\textbf{Fresh finite-measure comparison at $\kappa=1$.} Each frozen selector is evaluated using the same numerical Full-action functional on $128$ paired validation trials. Reductions are computed relative to the Law action at the same $\kappa$. Tangent and GlobalReaction are descriptive structural comparators; the final column reports the paired interval for the prespecified Full-versus-Law action reduction.}
\label{tab:unbalanced-primary-comparison}
\end{table}

The mechanism ablations below examine the structural restrictions imposed by Tangent and GlobalReaction. Tangent retains only correction visible through the measured raw moment rates, whereas GlobalReaction allows unrestricted transport but restricts reaction to the global mass-rate discrepancy. Their common-Full cross-scores exceed those of the Full-selected geometries in Table~\ref{tab:unbalanced-primary-comparison}. Section~\ref{app:unbalanced-mechanism-ablation} examines the corresponding Move--Reaction and Tangent--Hidden decompositions directly.

\subsection{Ablations}
\label{app:ablations}

We begin with two method-level ablations. The first examines FIDE's separation of completion from scoring by comparing independent per-time information projection with a restricted path-level completion at exactly fixed measured moments. The second asks when the baseline-preserving preference itself has a truth-relative downstream benefit: a controlled product-law construction makes the reference exactly correct in an unprotected factor, evaluates a held-out local-occupancy loss that is independent of Full's Sobolev geometry, and then deliberately violates baseline correctness. The remaining studies are benchmark-specific. The Gaussian common-pool analysis compares all secondary criteria on the same promoted candidates and separately perturbs the learned reference; held-out risk analyses assess scientific-risk transfer; prospective studies test the numerical evaluator and aggregate predictive interface; a manufactured solution tests continuum convergence of the periodic Full evaluator; skyrmion analyses examine richer potential spaces and support sensitivity; and the finite-measure analyses examine the transport--reaction extension. Unless stated otherwise, the relevant sensor geometries remain frozen throughout these diagnostics.

\subsubsection{Static information projection versus a restricted path-level completion}
\label{app:path-completion-ablation}

FIDE deliberately separates completion from scoring: each moment fiber is first completed by information projection, after which Full measures the dynamical revision required by the resulting path. A natural alternative is to let unresolved law directions vary jointly over time so as to reduce Full itself. These constructions answer different questions, but the distinction raises a quantitative issue: how much of the Full action assigned to an information-projected path is attributable to the measured moment trajectory, and how much can be changed by choosing a different completion of the same fibers? We examine this in a controlled one-dimensional example in which the measured trajectory is held exactly fixed while one otherwise unmeasured law direction is allowed to adapt dynamically.

Let the state space be the flat torus $\mathbb T=\mathbb R/(2\pi\mathbb Z)$, with uniform frozen reference $\widetilde q=(2\pi)^{-1}$ and reference velocity $u=0$. The measured observable is $\phi(x)=\cos x$, and we prescribe
\begin{equation}
c_n(t)=0.4\sin(2\pi nt),\qquad n\in\{1,2,4\}.
\label{eq:path-completion-target}
\end{equation}
The standard FIDE completion is the pointwise information projection
\begin{equation}
q_t^{\mathrm{IP}}(x)=\frac{\exp(\lambda_t\cos x)}{\int_{\mathbb T}\exp(\lambda_t\cos y)\,dy},
\qquad
\E_{q_t^{\mathrm{IP}}}[\cos X]=c_n(t).
\label{eq:path-completion-ip}
\end{equation}
To construct a restricted path-level alternative, we enlarge the exponential family by one unmeasured Fourier coordinate,
\begin{equation}
q_t^{\mathrm{dyn}}(x)=\frac{\exp(\lambda_t\cos x+\gamma_t\cos 2x)}{\int_{\mathbb T}\exp(\lambda_t\cos y+\gamma_t\cos 2y)\,dy},
\qquad
\E_{q_t^{\mathrm{dyn}}}[\cos X]=c_n(t),
\label{eq:path-completion-dynamic-family}
\end{equation}
and parameterize $\gamma_t=\sum_{k=1}^{5}a_k\sin(k\pi t)$. Thus $\gamma_0=\gamma_1=0$, and both completions have the same measured moment trajectory and the same endpoint densities. For every candidate $\gamma_t$, the multiplier $\lambda_t$ is recalibrated so that \eqref{eq:path-completion-target} holds exactly. We then optimize $a_1,\ldots,a_5\in[-2,2]$ to minimize the time-integrated Full action of this enlarged path. The static information projection is the feasible special case $a_1=\cdots=a_5=0$, so this ablation changes only the completion of the unmeasured law direction; it does not change the observed moment trajectory.

Because $u=0$, differentiation of the calibrated exponential family gives
\begin{equation}
\frac{\partial_tq_t}{q_t}
=
\dot\lambda_t\bigl(\cos x-c_n(t)\bigr)
+
\dot\gamma_t\bigl(\cos 2x-\E_{q_t}[\cos 2X]\bigr),
\qquad
\dot\lambda_t=
\frac{\dot c_n(t)-\Cov_{q_t}(\cos X,\cos 2X)\dot\gamma_t}
{\Var_{q_t}(\cos X)}.
\label{eq:path-completion-density-rate}
\end{equation}
For a periodic one-dimensional density path, let $F_t(x)=\int_0^x\partial_tq_t(y)\,dy$. The minimum-energy continuity flux has $q_t(x)\delta_t(x)=b_t-F_t(x)$, where
\begin{equation}
b_t=
\frac{\int_{\mathbb T}F_t(x)/q_t(x)\,dx}
{\int_{\mathbb T}1/q_t(x)\,dx},
\qquad
a_{\mathrm{full}}(t)=
\int_{\mathbb T}\frac{(b_t-F_t(x))^2}{q_t(x)}\,dx.
\label{eq:path-completion-one-dimensional-action}
\end{equation}
This gives the Full action for both paths without finite-differencing the density in physical time.

The measured trajectory itself imposes a nonzero action independently of the completion rule. Indeed,
\[
\dot c_n(t)=\E_{q_t}[-\sin(X)\delta_t(X)],
\]
so Cauchy--Schwarz and $\E[\sin^2X]\le1-c_n(t)^2$ imply
\begin{equation}
a_{\mathrm{full}}(t)\ge\frac{\dot c_n(t)^2}{1-c_n(t)^2}.
\label{eq:path-completion-lower-bound}
\end{equation}
We use the integral of \eqref{eq:path-completion-lower-bound} only as a universal moment-imposed lower bound; it need not be attainable by a feasible law path.

\begin{table}[t]
\centering
\small
\begin{tabular}{rrrrr}
\toprule
Frequency $n$ & Moment lower bound & Static I-projection & Restricted dynamic & Reduction (\%) \\
\midrule
$1$ & $3.295850$ & $6.521546$ & $5.890663$ & $9.67$ \\
$2$ & $13.183401$ & $26.086185$ & $20.867915$ & $20.00$ \\
$4$ & $52.733603$ & $104.344741$ & $77.063457$ & $26.15$ \\
\bottomrule
\end{tabular}
\caption{\textbf{Effect of the completion rule at fixed measured moments.} The static column evaluates Full on the independent per-time information projections in \eqref{eq:path-completion-ip}. The restricted dynamic column optimizes one additional unmeasured Fourier coordinate over time while preserving exactly the same trajectory $c_n(t)$. ``Reduction'' is $100(\Act_{\mathrm{IP}}-\Act_{\mathrm{dyn}})/\Act_{\mathrm{IP}}$. The moment lower bound is the time integral of \eqref{eq:path-completion-lower-bound} and is not asserted to be attainable.}
\label{tab:path-completion-ablation}
\end{table}

The completion choice is therefore consequential but does not account for all of the dynamical cost. At $n=1$, allowing the additional unmeasured mode reduces action from $6.52$ to $5.89$, a $9.67\%$ reduction, while both values remain well above the universal moment-imposed lower bound of $3.30$. The reduction grows to $20.00\%$ and $26.15\%$ over the two more rapidly varying trajectories tested. This shows that part of the Full action assigned to the static information-projected path depends on how the unresolved fiber directions are completed rather than being forced solely by the measured moment trajectory.

This result does not make the path-level construction a replacement for the FIDE completion rule. The restricted dynamic path is optimized using the same Full criterion by which it is subsequently evaluated, whereas FIDE intentionally fixes the completion independently of Full and then asks how much dynamical revision that prescribed completion requires. The two procedures therefore encode different modeling choices: the former uses unmeasured law directions to reduce dynamical action, while the latter uses a common local minimum-information-revision rule to resolve those directions before dynamical scoring. The ablation quantifies the price of that separation in this toy family rather than establishing that one completion is scientifically superior.

The numerical calculation uses $512$ periodic spatial points and $121$ physical-time nodes. Moment calibration, mass compatibility, and differentiated-moment compatibility hold to below $6\times10^{-15}$ in the reported runs, and setting all dynamic coefficients to zero reproduces the independently evaluated static action to machine precision. Re-evaluation of the frozen solutions on successively finer spatial and temporal grids changes the reported completion gaps by less than $0.006$ percentage points. For $n=2$ and $n=4$, the first coefficient reaches the prespecified lower bound $-2$; the increase in relative gap across the three tested frequencies should therefore be read descriptively rather than as a converged or universal frequency law. The dynamic construction is restricted to the family in \eqref{eq:path-completion-dynamic-family} and is not claimed to attain the globally minimum-action path over all laws satisfying the moment constraints.

\subsubsection{When does baseline preservation help? A controlled downstream ablation}
\label{app:baseline-correctness-ablation}

The baseline-preserving specification is useful only if preserving the reference in directions left outside the scientific task has a meaningful consequence. We therefore construct a population-level ablation in which this assumption is exactly true and then violate it in a controlled way. The purpose is not to show that Full recovers truth in general (Proposition~\ref{prop:reference-consistency} rules out such a claim), but to ask two narrower questions: when the reference is correct in an unprotected part of the state, which secondary selectors protect a downstream quantity that is independent of their native objectives, and how do those consequences change as the same unprotected direction becomes genuinely misspecified?

\paragraph{A baseline-correct residual design problem.} Let $(Y,Z)\in\mathbb T^2$ with uniform stationary reference $\widetilde q=(2\pi)^{-2}$ and $u=0$, and write $e_k(x)=\sqrt{2}\cos(kx)$. At baseline correctness the true population is
\begin{equation}
p_{t,0}(y,z)=\frac{1+\epsilon s(t)e_1(y)}{(2\pi)^2},\qquad s(t)=\sin(2\pi t),
\label{eq:baseline-correct-truth}
\end{equation}
so the scientific signal evolves only in $Y$ while the unprotected $Z$ marginal is exactly the stationary reference. Candidate sensors deliberately mix the protected signal with different unresolved $Z$ directions:
\begin{equation}
\phi_\alpha(y,z)=e_1(y)+\psi_\alpha(z),\qquad \psi_\alpha(z)=\sum_{k\in\{1,3,10\}}\sqrt{w_{\alpha,k}}\,e_k(z),\qquad \alpha\in\{0,\tfrac14,\tfrac12,\tfrac34,1\},
\label{eq:baseline-correct-sensors}
\end{equation}
where $w_\alpha=(1-\alpha)(0,90/91,1/91)+\alpha(10/11,0,1/11)$. Every $\psi_\alpha$ has the same $L^2$ energy and the same gradient energy,
\[
\sum_k w_{\alpha,k}=1,\qquad \sum_k k^2w_{\alpha,k}=10,
\]
but different inverse-transport energy
\begin{equation}
H_\alpha:=\sum_{k\in\{1,3,10\}}\frac{w_{\alpha,k}}{k^2}\in\{0.11,0.31,0.51,0.71,0.91\}.
\label{eq:baseline-correct-H}
\end{equation}
At $\gamma=0$, every sensor has the same target moment $\E_{P_t}[\phi_\alpha]=\epsilon s(t)$ and exactly the same integrated sampling risk, $2-\epsilon^2/2$. Scientific risk therefore leaves all five designs tied by construction; the ablation isolates only how a secondary criterion resolves that residual freedom.

The information projection factorizes as $Q_t^\alpha=Q_{Y,t}^\alpha\otimes Q_{Z,t}^\alpha$. For $C_\epsilon:=2\pi^2\epsilon^2$, the small-amplitude expansion gives
\begin{equation}
\Act(\alpha)=\frac{C_\epsilon}{4}(1+H_\alpha)+o(\epsilon^2).
\label{eq:baseline-correct-full-local}
\end{equation}
To expose the corresponding hidden mechanism directly, we also evaluate a fixed-reference Sobolev rate error on the unprotected factor,
\begin{equation}
D_Z(\alpha):=\int_0^1\sum_{m\ge1}\frac{|\dot m_{m,c}^\alpha(t)|^2+|\dot m_{m,s}^\alpha(t)|^2}{m^2}\,dt,\qquad D_Z(\alpha)=\frac{C_\epsilon}{4}H_\alpha+o(\epsilon^2),
\label{eq:baseline-correct-Dz}
\end{equation}
where $m_{m,c}^\alpha$ and $m_{m,s}^\alpha$ are the cosine and sine moments of $Q_{Z,t}^\alpha$. Because the true $Z$ marginal is stationary at baseline correctness, these are truth-relative spurious rates. Full and $D_Z$ therefore have the same predicted strict ordering, with $\alpha=0$ preferred. Tangent, residual $L^2$, and forward KL are independent of $\alpha$ at leading order because the corresponding $L^2$ and $H^1$ quantities were matched; static $W_2$ shares the local inverse-transport ordering and is therefore expected to agree with Full in this construction.

The numerical experiment uses the prespecified amplitudes $\epsilon\in\{0.10,0.05,0.025\}$ without parameter search. At every amplitude, Full and $D_Z$ are strictly increasing in $\alpha$ and have Spearman rank correlation one. The maximum relative errors of the leading Full and $D_Z$ expansions are respectively $2.88\times10^{-4}$ and $9.83\times10^{-4}$ at $\epsilon=0.10$, and decrease to $1.8\times10^{-5}$ and $6.14\times10^{-5}$ at $\epsilon=0.025$. Tangent, residual $L^2$, and KL have small finite-amplitude preferences for $\alpha=0.75$, but their relative score ranges shrink with $\epsilon$, consistently with their leading-order indifference. Static $W_2$ selects the same $\alpha=0$ geometry as Full.

\paragraph{Held-out local occupancy: an independent downstream loss.} The preceding $D_Z$ diagnostic is intentionally aligned with the inverse-transport mechanism underlying Full, so it is not by itself an independent downstream test. We therefore freeze the five sensors and evaluate ordinary local population predictions that never enter scientific risk, information projection, or any secondary criterion. For $16$ evenly spaced centers $\theta_j=2\pi j/16$ and three prespecified localization scales $\kappa\in\{2,4,8\}$, define
\begin{equation}
g_{\kappa,j}(z)=\exp\!\bigl(\kappa[\cos(z-\theta_j)-1]\bigr),
\label{eq:occupancy-windows}
\end{equation}
and measure
\begin{equation}
\operatorname{OccRMSE}_{\kappa}(\alpha):=\left[\frac1{16}\sum_{j=0}^{15}\int_0^1\left(\E_{Q_{Z,t}^\alpha}[g_{\kappa,j}]-\E_{P_{Z,t}}[g_{\kappa,j}]\right)^2dt\right]^{1/2}.
\label{eq:occupancy-rmse}
\end{equation}
This loss is evaluated directly against the known true $Z$ marginal. It contains no reference-relative rate discrepancy, $H^{-1}$ norm, inverse Laplacian, correction energy, gradient normalization, or design-dependent weighting.

\begin{figure}[t]
\centering
\includegraphics[width=0.78\linewidth]{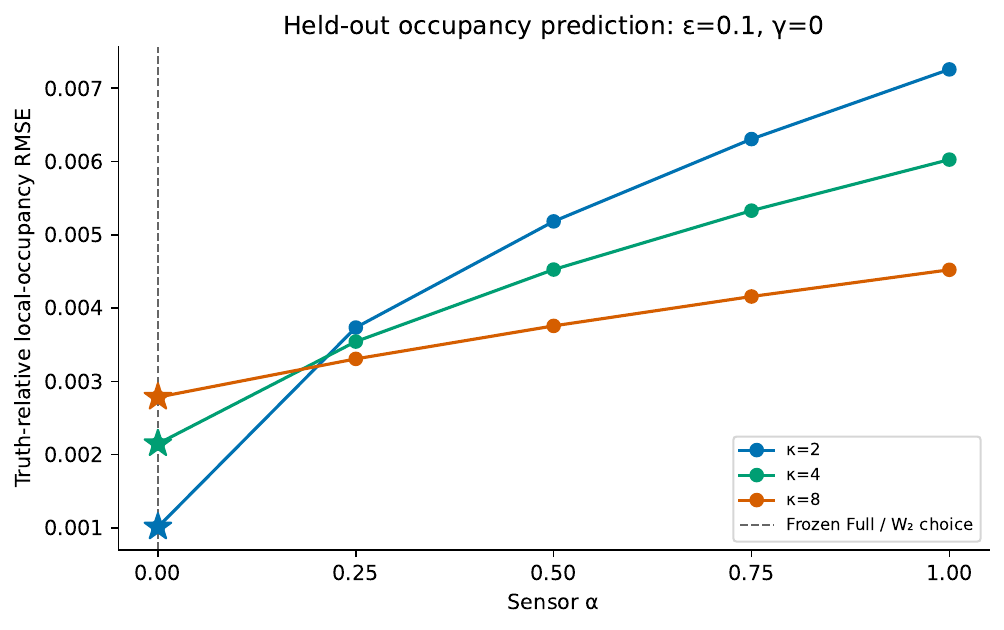}
\caption{\textbf{Candidate-family occupancy error under baseline correctness.} At $\epsilon=0.10$ and $\gamma=0$, truth-relative local-occupancy RMSE is shown for all five frozen geometries and all three prespecified localization scales $\kappa\in\{2,4,8\}$. Scientific risk is identical across the five designs. Full and static $W_2$ select $\alpha=0$, Tangent, residual $L^2$, and forward KL select $\alpha=0.75$, and Law is indifferent among all five geometries. The occupancy loss is withheld from design and is not used by any selector.}
\label{fig:baseline-occupancy}
\end{figure}

At baseline correctness, Full and static $W_2$ select $\alpha=0$, while Tangent, residual $L^2$, and forward KL select $\alpha=0.75$ in the finite-amplitude calculation. Law remains exactly indifferent among all five geometries. For the selector-level comparison, we therefore use an outcome-independent Law policy that draws uniformly from the complete set of five risk minimizers and then freezes the sampled geometry; its expected occupancy RMSE is the arithmetic mean of the five per-geometry RMSEs, not the RMSE of a mixture distribution and not a tie broken using the downstream loss. At $\epsilon=0.10$ and $\kappa=4$, the resulting RMSEs are $2.1492\times10^{-3}$ for Full/static $W_2$, $4.3147\times10^{-3}$ for the expected Law policy, and $5.3296\times10^{-3}$ for Tangent/residual $L^2$/forward KL, corresponding to differences of $-50.19\%$ and $+23.52\%$ relative to expected Law. Across all three tested amplitudes and all three localization scales, the correct-baseline ordering remains Full/static $W_2$ $<$ expected Law $<$ Tangent/residual $L^2$/forward KL. Figure~\ref{fig:baseline-occupancy} retains the more detailed candidate-family view: occupancy error increases monotonically across the five frozen geometries, with $\alpha=0<0.25<0.5<0.75<1$ at all three localization scales.

The occupancy result is also predicted independently of Full. Writing $I_k$ for the modified Bessel function,
\begin{equation}
\operatorname{OccMSE}_{\kappa}(\alpha)=\frac{\epsilon^2e^{-2\kappa}}{8}\sum_{k\in\{1,3,10\}}w_{\alpha,k}I_k(\kappa)^2+o(\epsilon^2).
\label{eq:occupancy-local-expansion}
\end{equation}
This is a different spectral weighting from $H_\alpha=\sum_kw_{\alpha,k}/k^2$. The predicted squared-error ratios between $\alpha=1$ and $\alpha=0$ are $51.3866$, $7.8609$, and $2.6372$ for $\kappa=2,4,8$; at $\epsilon=0.025$ the numerical ratios are $51.3868$, $7.8610$, and $2.6373$. Thus the downstream agreement is not obtained by evaluating the Full objective under another name.

\paragraph{Deliberately violating baseline correctness.} We next add a genuine unprotected departure to the true $Z$ marginal while leaving the reference and sensor family unchanged:
\begin{equation}
p_{t,\gamma}(y,z)=\frac{1+\epsilon s(t)e_1(y)+\gamma\epsilon r(t)e_1(z)}{(2\pi)^2},\qquad r(t)=\sin(4\pi t).
\label{eq:baseline-misspecified-truth}
\end{equation}
The $\alpha=0$ sensor has no $k=1$ component in $Z$ and is therefore blind to this new departure, whereas $\alpha=1$ is strongly sensitive to it. No design is reselected as $\gamma$ changes. The local expansion predicts that the $\alpha=0$ versus $\alpha=1$ ordering in $D_Z$ reverses at $\gamma_\star=\sqrt{22/309}\approx0.2668$; the numerical crossover estimates converge to this value as $\epsilon$ decreases.

\begin{figure}[t]
\centering
\includegraphics[width=\linewidth]{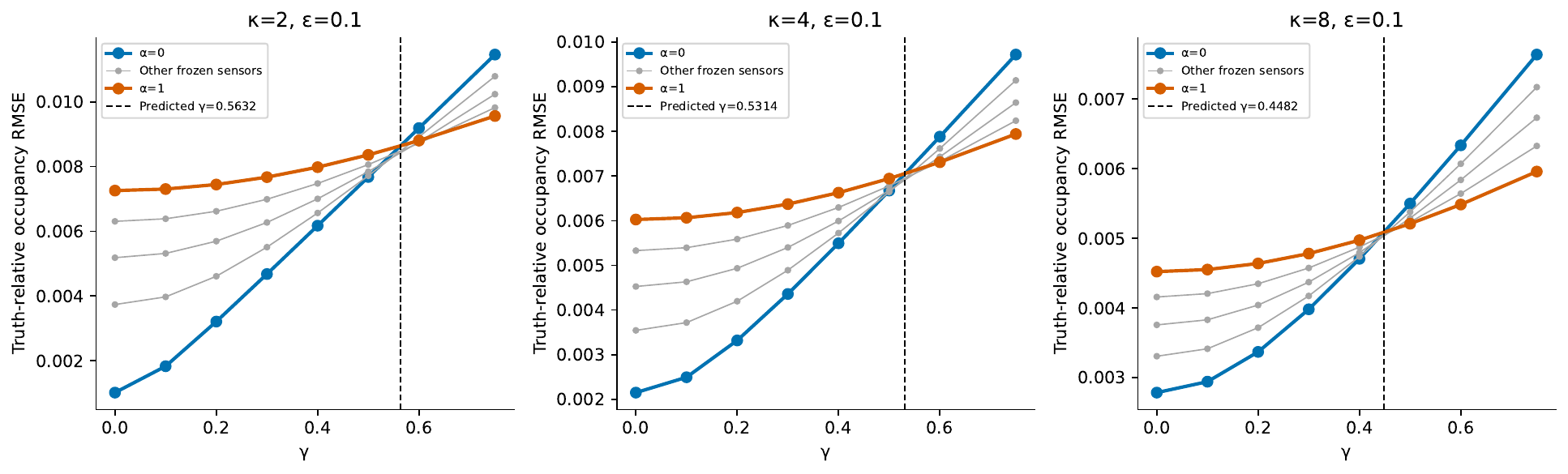}
\caption{\textbf{Candidate-family occupancy error under misspecification.} At $\epsilon=0.10$, truth-relative local-occupancy RMSE is shown across all five frozen geometries as the unprotected departure strength $\gamma$ increases. Blue denotes $\alpha=0$, orange denotes $\alpha=1$, and gray denotes $\alpha\in\{0.25,0.5,0.75\}$. The baseline-preserving $\alpha=0$ geometry is best near $\gamma=0$ but eventually becomes worse than more departure-sensitive geometries. Dashed vertical lines mark the independently predicted small-amplitude $\alpha=0$ versus $\alpha=1$ occupancy crossovers from \eqref{eq:occupancy-misspec-expansion}: $\gamma_\star=0.5632$, $0.5314$, and $0.4482$ for $\kappa=2,4,8$, respectively. No geometry is reselected as $\gamma$ varies; Figure~\ref{fig:fide-comparator-misspecification} gives the selector-level comparison emphasized in the main text.}
\label{fig:baseline-misspecification}
\end{figure}

The independently defined occupancy loss has its own, later, scale-dependent reversal. Its small-amplitude expansion is
\begin{equation}
\operatorname{OccMSE}_{\kappa}(w;\gamma)=\epsilon^2e^{-2\kappa}\left[\frac{A_\kappa(w)}8+\frac{\gamma^2B_\kappa(w)}2\right]+o(\epsilon^2),
\label{eq:occupancy-misspec-expansion}
\end{equation}
with $A_\kappa(w)=\sum_{k\in\{1,3,10\}}w_kI_k(\kappa)^2$ and $B_\kappa(w)=I_1(\kappa)^2(w_1/2-1)^2+\frac{w_1}{4}\{w_3I_3(\kappa)^2+w_{10}I_{10}(\kappa)^2\}$. This gives precomputed $\alpha=0$ versus $\alpha=1$ crossover predictions $0.5632$, $0.5314$, and $0.4482$ for $\kappa=2,4,8$. Using only the prespecified $\gamma$ grid, the corresponding estimates at $\epsilon=0.025$ are $0.56324$, $0.53139$, and $0.44817$. Figure~\ref{fig:baseline-misspecification} shows the complete candidate-family sweep at $\epsilon=0.10$: the baseline-preserving $\alpha=0$ geometry is best near $\gamma=0$, but its advantage erodes as the unprotected departure grows. Other frozen geometries eventually overtake it, and the $\alpha=0$ versus $\alpha=1$ ordering reverses at the separately predicted, scale-dependent thresholds. These occupancy crossovers are substantially different from the $D_Z$ crossover, confirming that the held-out occupancy loss probes a different downstream geometry. The main-text selector-level comparison in Figure~\ref{fig:fide-comparator-misspecification} freezes the actual finite-amplitude selections, namely $\alpha=0$ for Full/static $W_2$ and $\alpha=0.75$ for Tangent/residual $L^2$/forward KL, and compares each with the expected Law tie-breaking policy. At $\epsilon=0.10$ and $\kappa=4$, both differences change sign between the evaluated points $\gamma=0.5$ and $0.6$; linear interpolation of the plotted segments gives $0.5156$ for Full/static $W_2$ and $0.5160$ for Tangent/residual $L^2$/forward KL. These interpolants are descriptive crossing locations rather than separately evaluated roots. 

The same reversal directions hold for all nine combinations of $\epsilon\in\{0.10,0.05,0.025\}$ and $\kappa\in\{2,4,8\}$: Full/static $W_2$ change from lower to higher occupancy error than expected Law, while Tangent/residual $L^2$/forward KL change from higher to lower error. The crossing bracket is $[0.5,0.6]$ for $\kappa\in\{2,4\}$ and $[0.4,0.5]$ for $\kappa=8$ at every tested amplitude.

The main-text selector-level summary in Figure~\ref{fig:fide-comparator-misspecification}, together with the detailed candidate-family views in Figures~\ref{fig:baseline-occupancy}--\ref{fig:baseline-misspecification}, isolates the scope condition behind baseline-preserving residual design. When the task leaves a direction unprotected and the declared baseline is correct there, the geometry selected by Full and static $W_2$ preserves that structure and yields the smallest occupancy error among the selector groups considered here. As a genuine departure in the same unprotected direction grows, that advantage erodes and eventually reverses, while the geometry selected by Tangent, residual $L^2$, and forward KL exhibits the opposite change relative to the Law tie policy. No selector is therefore uniformly favored by this truth-relative downstream loss across baseline regimes. The result supports a conditional role for baseline preservation rather than a robustness guarantee under misspecification, and it does not establish a Full-specific advantage over static $W_2$ in this construction.

No parameter search is used. The three amplitudes, five sensors, three occupancy scales, and misspecification grid are fixed before evaluation. Across all nine amplitude--localization-scale combinations, the correct-baseline selector ordering Full/static $W_2$ $<$ expected Law $<$ Tangent/residual $L^2$/forward KL and the corresponding reversal directions under increasing misspecification are unchanged, although effect magnitudes and crossing locations depend on localization scale. All candidate-family rankings also persist over nine spatial--temporal refinements; the largest relative occupancy-metric change is $3.1\times10^{-7}$, direct quadrature and Fourier/Bessel evaluations agree within $10^{-12}$, and the original baseline-ablation outputs are unchanged.

\subsubsection{Do the held-out designs satisfy their scientific-risk allowances under sampling uncertainty?}
\label{app:risk-feasibility-ablation}
Selection feasibility does not imply that the same relative risk allowance transfers to fresh observations. For studies whose current validation artifacts retain paired per-trial scientific losses, we assess the held-out constraint directly. For paired evaluation losses, define
\begin{equation}
Z_i(p)=\ell_i(\etavec_{\mathrm{Full},p})-\left(1+\frac{p}{100}\right)\ell_i(\etavec_{\mathrm{Law}}),
\label{eq:heldout-risk-feasibility}
\end{equation}
so that $\E[Z_i(p)]<0$ corresponds to satisfaction of the $p\%$ relative allowance on the evaluation distribution. For each benchmark family, we use $100{,}000$ paired bootstrap resamples with common resampled trial indices across allowances. Writing $\bar Z_p$ for the observed paired mean and $\bar Z_p^{\,*}$ for its bootstrap replicate, the pointwise one-sided upper bound is $\bar Z_p+q_{0.95}(\bar Z_p^{\,*}-\bar Z_p)$, while the simultaneous bound replaces the scalar bootstrap deviation by its maximum over all allowances in that family. An operating point is supported below its held-out cap only when this simultaneous upper bound is negative; otherwise we report it as unresolved. Multiplicity control is applied separately within each benchmark family. The current paired-loss analysis covers the Gaussian, prospective double-gyre, and skyrmion validation studies. The current retrospective six-allowance frontier is treated separately because its retained evaluation summaries do not contain the paired per-trial Law and Full scientific-risk losses required for a matching bootstrap analysis.

\begin{figure}[t]
\centering
\includegraphics[width=0.82\linewidth]{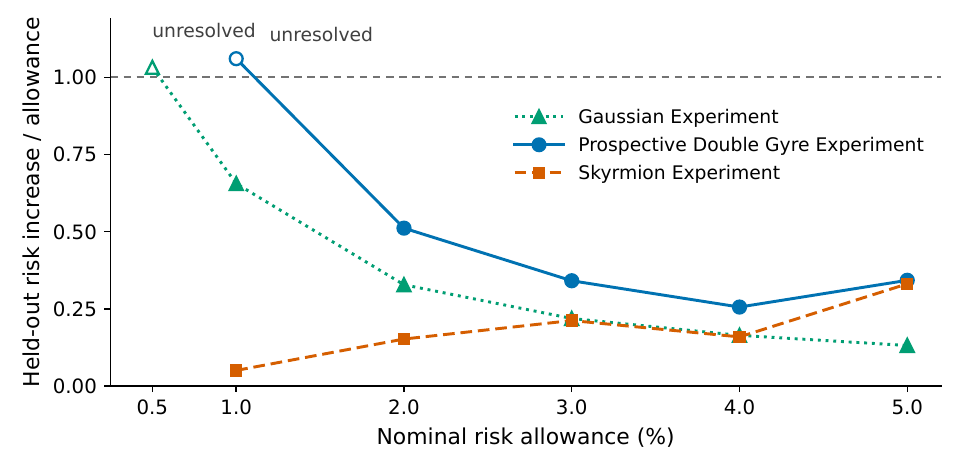}
\caption{\textbf{Do the held-out designs satisfy their scientific-risk allowances under sampling uncertainty?} Descriptive held-out relative risk increase divided by the nominal allowance for the Gaussian-mixture, prospective double-gyre, and skyrmion validation studies. Gaussian and prospective double-gyre use their respective $64$-trial independent holdouts, while skyrmion uses $32$ fresh paired validation trials. The dashed line at one marks equality with the nominal allowance. Filled markers denote operating points whose simultaneous one-sided $95\%$ upper bound for $\E[Z_i(p)]$ is below zero; open markers are unresolved. The plotted ratio is descriptive: inferential status is determined by the paired excess-loss statistic in \eqref{eq:heldout-risk-feasibility}, not by the ratio alone.}
\label{fig:risk-feasibility}
\end{figure}

\begin{table}[t]
\centering
\small
\begin{tabular}{llrrl}
\toprule
Study & Allowance & Risk increase (\%) & Simultaneous upper & Conclusion \\
\midrule
Gaussian & $0.5\%$ & 0.516 & $1.61\times10^{-4}$ & Unresolved \\
Gaussian & $1\%$ & 0.656 & $-7.58\times10^{-5}$ & Below cap \\
Gaussian & $2\%$ & 0.656 & $-7.33\times10^{-4}$ & Below cap \\
Gaussian & $3\%$ & 0.656 & $-1.39\times10^{-3}$ & Below cap \\
Gaussian & $4\%$ & 0.656 & $-2.05\times10^{-3}$ & Below cap \\
Gaussian & $5\%$ & 0.656 & $-2.70\times10^{-3}$ & Below cap \\
\midrule
Prospective & 1\% & 1.060 & $8.40\times10^{-3}$ & Unresolved \\
Prospective & 2\% & 1.022 & $-4.31\times10^{-3}$ & Below cap \\
Prospective & 3\% & 1.022 & $-1.66\times10^{-2}$ & Below cap \\
Prospective & 4\% & 1.022 & $-2.88\times10^{-2}$ & Below cap \\
Prospective & 5\% & 1.711 & $-3.26\times10^{-2}$ & Below cap \\
\midrule
Skyrmion & 1\% & 0.050 & $-2.62\times10^{-3}$ & Below cap \\
Skyrmion & 2\% & 0.304 & $-6.52\times10^{-3}$ & Below cap \\
Skyrmion & 3\% & 0.636 & $-1.00\times10^{-2}$ & Below cap \\
Skyrmion & 4\% & 0.636 & $-1.52\times10^{-2}$ & Below cap \\
Skyrmion & 5\% & 1.653 & $-1.52\times10^{-2}$ & Below cap \\
\bottomrule
\end{tabular}
\caption{\textbf{Held-out risk-feasibility inference.} Relative risk increase is $100(\overline\ell_{\mathrm{Full}}/\overline\ell_{\mathrm{Law}}-1)$. The fourth column gives the simultaneous one-sided $95\%$ upper bound for the paired excess-loss statistic $Z_i(p)$ within the corresponding benchmark family. A point is labeled below cap only when this bound is negative.}
\label{tab:risk-feasibility}
\end{table}

Table~\ref{tab:risk-feasibility} sharpens the held-out scientific-risk assessment for the Gaussian, prospective double-gyre, and skyrmion validation studies. In the Gaussian study, the $0.5\%$ operating point is unresolved under simultaneous one-sided inference, whereas the $1\%$--$5\%$ points are supported below their Law-relative held-out caps. In the prospective double-gyre study, the $1\%$ point likewise remains unresolved, with $\bar Z=7.36\times10^{-4}$ and simultaneous upper bound $8.40\times10^{-3}$, whereas the $2\%$--$5\%$ operating points are supported below their held-out caps. All five fresh skyrmion operating points are supported below their corresponding caps. The current retrospective six-allowance frontier is not included in this bootstrap analysis because the retained evaluation summaries do not contain the paired per-trial scientific-risk losses required to construct matching excess-loss statistics; its verified holdout point estimates are reported in Appendix~\ref{app:exp-vortices}.

The concentration of uncertainty at the tightest allowance is expected from the nested feasibility thresholds: for a fixed held-out loss pair, increasing $p$ shifts $Z_i(p)$ downward by $(p/100)\ell_i(\etavec_{\mathrm{Law}})$ without comparably increasing its sampling variability. This is especially transparent for the prospective $2\%$--$4\%$ points, which use the same frozen Full geometry and have the same $1.022\%$ held-out risk increase; the increasingly negative feasibility margin therefore reflects additional tolerance rather than a change in the validated design.

\subsubsection{Common-pool comparison of Gaussian design criteria}
\label{app:gaussian-common-candidate-comparison}

To complement the criterion-specific optimization results, we compare all six objectives on a shared set of scientifically competitive designs. Let $\mathcal C$ be the union of the geometrically admissible candidates that reached the final evaluation stage under Law, Tangent, Full, residual $L^2$, forward KL, or static $W_2$. Across the six selectors and risk allowances, this gives $17$ unique promoted candidates. The common-pool Full scores use the $64$-trial selection bank, whereas Table~\ref{tab:exp-gaussian-criteria} evaluates the same Law and Full geometries with the same numerical Full evaluator on an independent $64$-trial holdout bank. Lower-fidelity prescreen candidates and raw optimization iterates are excluded so that every design in $\mathcal C$ has undergone the same final evaluation protocol.

\begin{figure}[t]
\centering
\begin{subfigure}[t]{0.49\linewidth}
\centering
\includegraphics[width=\linewidth]{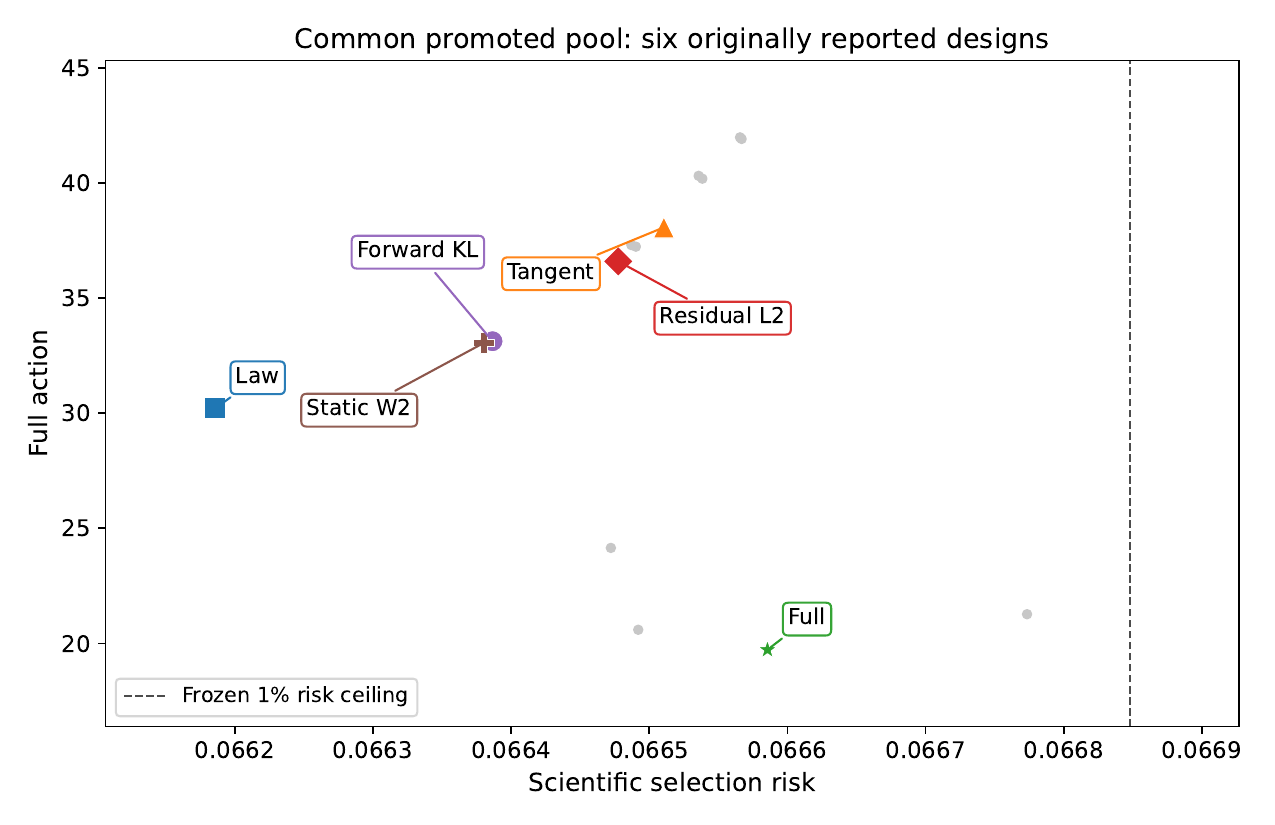}
\caption{Scientific risk versus Full action.}
\label{fig:gaussian-common-pool-risk-action}
\end{subfigure}
\hfill
\begin{subfigure}[t]{0.49\linewidth}
\centering
\includegraphics[width=\linewidth]{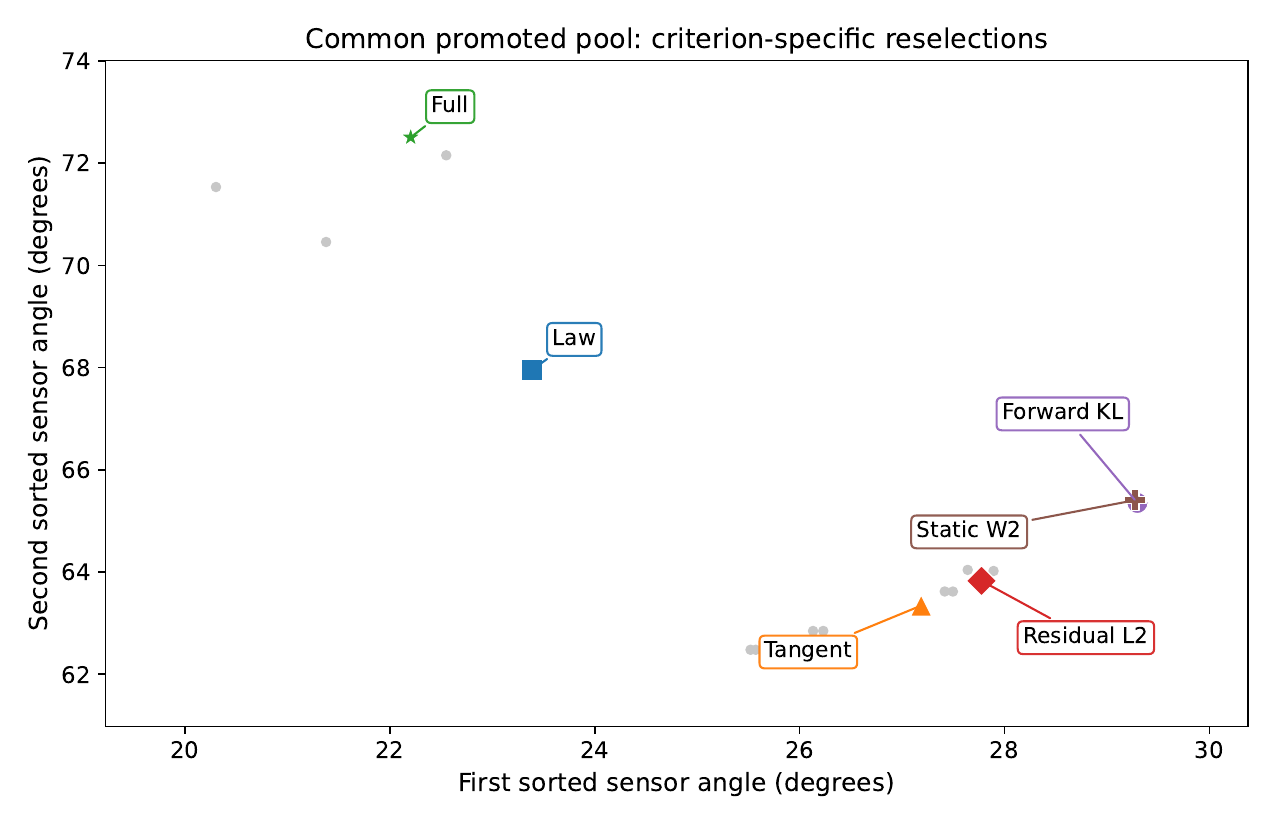}
\caption{Sensor geometries in the two-angle design space.}
\label{fig:gaussian-common-pool-geometry}
\end{subfigure}
\caption{\textbf{Gaussian criterion-specific selections on the common promoted candidate pool.} Gray points denote the remaining promoted candidates and labeled markers denote the six criterion-specific minimizers obtained by reselecting each objective over the shared pool. \textbf{Left:} scientific risks occupy a narrow range while Full action varies substantially across candidate designs; the dashed line marks the frozen $1\%$ operational risk ceiling. Forward KL and static $W_2$ select the same candidate and therefore overlap. \textbf{Right:} the corresponding sensor geometries occupy distinct regions of the design space, again with the forward-KL and static-$W_2$ selections coincident.}
\label{fig:gaussian-common-pool-comparison}
\end{figure}

At the $1\%$ allowance, the operational feasibility threshold is
\begin{equation}
R_{\mathrm{sel}}(\etavec)\le 1.01R_{\mathrm{Law}}=0.06684759,
\end{equation}
together with the population, support, and numerical requirements of the Gaussian experiment. All $17$ candidates in $\mathcal C$ satisfy these conditions. For each criterion $J$, we define its common-pool selection by
\begin{equation}
\etavec_J^{\mathcal C}\in\operatorname*{arg\,min}_{\etavec\in\mathcal C:\,\etavec\ \mathrm{feasible}}J(\etavec).
\label{eq:gaussian-common-pool-selection}
\end{equation}
Table~\ref{tab:gaussian-common-candidate-comparison} reports these selections and their cross-evaluated Full actions. Figure~\ref{fig:gaussian-common-pool-comparison} provides the corresponding geometric view: panel (a) compares scientific risk and Full action across the common pool, while panel (b) locates the same designs in the two-angle sensor space.

\begin{table}[t]
\centering
\small
\begin{tabular}{lrrr}
\toprule
Criterion & Scientific risk & Own objective & Full action \\
\midrule
Law & 0.066186 & 0.066186 & 30.2016 \\
Tangent & 0.066511 & 2.10920 & 38.0653 \\
Full & 0.066586 & 19.71935 & 19.7193 \\
Residual $L^2$ & 0.066478 & 52.50258 & 36.5877 \\
Forward KL & 0.066387 & 0.235235 & 33.1209 \\
Static $W_2$ & 0.066387 & 0.134317 & 33.1209 \\
\bottomrule
\end{tabular}
\caption{\textbf{Gaussian criterion-specific selections on a shared promoted candidate pool.} Each row gives the minimizer of one criterion over the same set of $17$ scientifically feasible designs. The ``own objective'' column is reported on each criterion's native scale and is therefore not comparable across rows; the final column cross-evaluates every selected design using the common Full-action functional. Forward KL and static $W_2$ select the same candidate, whereas Law, Tangent, Full, and residual $L^2$ select distinct candidates.}\label{tab:gaussian-common-candidate-comparison}
\end{table}

Five distinct criterion-specific selections arise from the common pool, with only forward KL and static $W_2$ selecting the same candidate. Law and Full recover the geometries from the original comparison, while static $W_2$ moves to the candidate selected by forward KL. Under common Full cross-evaluation, the selected actions are $19.7193$ for Full, $30.2016$ for Law, $38.0653$ for Tangent, $36.5877$ for residual $L^2$, and $33.1209$ for the shared KL/static-$W_2$ selection. These are selection-bank cross-scores, not values on a common native objective scale, and they should not be identified with the independent held-out scores in Table~\ref{tab:exp-gaussian-criteria}. The scientific risks remain closely matched across the promoted pool, while both the selected geometries and their Full cross-scores differ substantially.

\subsubsection{Controlled reference misspecification in the Gaussian common pool}
\label{app:gaussian-reference-misspecification}

Appendix~\ref{app:baseline-correctness-ablation} isolates the baseline-correctness mechanism in a controlled population setting with known truth and a predeclared failure threshold. Here we ask a complementary benchmark-specific question: does the same reference sensitivity appear when the learned Gaussian reference itself is perturbed while the original scientific task and promoted candidate pool are held fixed? Starting from the frozen reference $(\widetilde Q_t,u_t)$, we construct five endpoint-preserving reference conditions by applying a smooth time-dependent rotation with $\gamma\in\{-\pi/6,-\pi/12,0,\pi/12,\pi/6\}$, corresponding to $-30^\circ,-15^\circ,0,+15^\circ,+30^\circ$. Let $\theta_\gamma(t)=\gamma\sin^2(\pi t)$ and let $R_\theta$ denote the planar rotation matrix. We define
\[
\widetilde Q_t^\gamma=(R_{\theta_\gamma(t)})_\#\widetilde Q_t,\qquad u_t^\gamma(y)=\dot\theta_\gamma(t)Jy+R_{\theta_\gamma(t)}u_t\!\left(R_{-\theta_\gamma(t)}y\right),\qquad J=\begin{pmatrix}0&-1\\1&0\end{pmatrix}.
\]
The transformed law--velocity pair therefore remains dynamically consistent, and because $\theta_\gamma(0)=\theta_\gamma(1)=0$, every perturbation has the same endpoint reference laws as the original condition.

The purpose is to isolate reference dependence of the secondary selection rather than redefine the scientific task. We therefore keep the original $1\%$ scientific-risk threshold, population screen, and $17$ promoted candidates fixed across all five conditions. Every candidate remains admissible and numerically certified. Tangent, residual $L^2$, forward KL, static $W_2$, and Full re-rank this same pool after recalibration against the perturbed reference; Law remains the original action-independent risk minimizer. At $\gamma=0$, all six selectors reproduce their original common-pool selections.

We measure sensitivity to the imposed reference error separately from each selector's native objective. For candidate $\etavec$, define
\[
D_{\mathrm{detect}}(\etavec,\gamma)=\frac{1}{B}\sum_{b=1}^{B}\sum_q\omega_q\,\Delta\mu_{bq}^{\top}V_{bq,\etavec}^{-1}\Delta\mu_{bq},\qquad \Delta\mu_{bq}:=\E_{P_{b,t_q}}[\Phivec_{\etavec}]-\E_{\widetilde Q_{t_q}^{\gamma}}[\Phivec_{\etavec}],
\]
where $V_{bq,\etavec}$ is the existing measurement covariance and $\omega_q$ are the scientific-time quadrature weights. Larger $D_{\mathrm{detect}}$ means that the candidate measurements distinguish the analytic target from the perturbed reference more strongly. This diagnostic is distinct from truth-relative completion error, which we continue to measure using the original hidden-law sliced Wasserstein metric with $128$ fixed directions.

\begin{figure}[t]
\centering
\includegraphics[width=\linewidth]{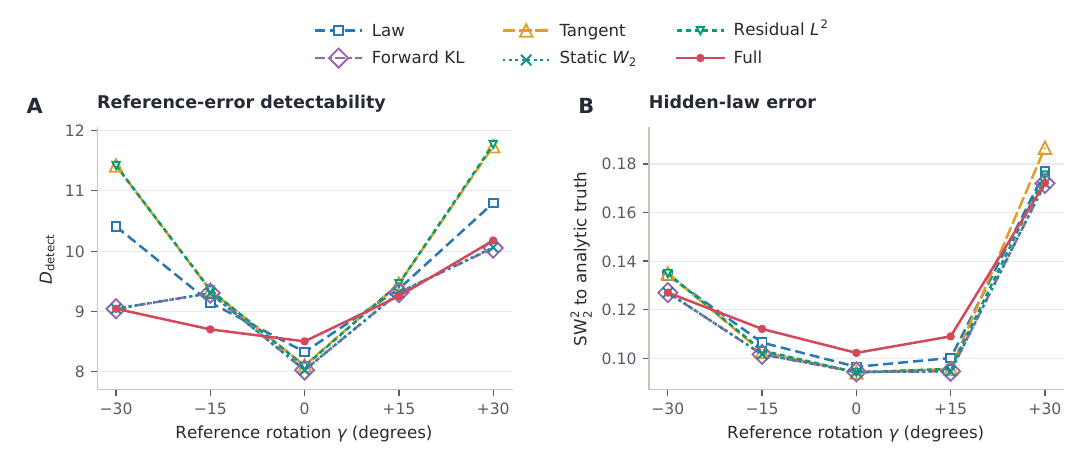}
\caption{\textbf{Controlled Gaussian reference misspecification on the fixed $1\%$ admissible pool.} \textbf{Left:} covariance-normalized detectability of the discrepancy between the analytic target and the perturbed reference; larger values mean that the selected sensors expose the reference error more strongly. \textbf{Right:} truth-relative hidden-law sliced-Wasserstein error of the corresponding information-projected completion; smaller values mean closer agreement with the analytic intermediate law. The two diagnostics need not induce the same ordering.}
\label{fig:gaussian-reference-misspecification}
\end{figure}

The selectors separate clearly under misspecification. At all four nonzero rotations, the Full-selected geometry has lower $D_{\mathrm{detect}}$ than Law and ranks between $14$th and $17$th among the $17$ admissible candidates by detectability. Forward KL and static $W_2$ are also less detectable than Law in three of the four conditions, with nonzero-condition ranks between $11$th and $17$th. Tangent and residual $L^2$, by contrast, are never less detectable than Law in this sweep and occupy ranks $4$--$8$ and $1$--$8$, respectively. Reduced sensitivity to a misspecified reference is therefore neither unique to Full nor a generic consequence of using the reference anywhere in the construction: it appears consistently for Full and often for the static reference-proximity criteria, but not for the two residual-rate criteria in this finite pool.

Detectability and truth fidelity remain separate. Forward KL and static $W_2$ have lower hidden-law $\mathrm{SW}_2^2$ than Law at all four nonzero perturbations despite often selecting less detectable measurements. Full has higher hidden-law error than Law at the two $\pm15^\circ$ conditions and lower error at the two $\pm30^\circ$ conditions. Thus lower sensitivity to reference error does not by itself imply worse reconstruction of the hidden law, nor does higher detectability guarantee better truth-relative completion.

Finally, the rank changes should not be read as evidence that increasing misspecification causes Full to progressively move toward ``blind'' sensor geometries. Full selects the same geometry at $0^\circ$, $\pm15^\circ$, and $-30^\circ$, switching only to a nearby geometry at $+30^\circ$; much of the detectability-rank change occurs because the reference being compared against changes while the geometry does not. The ablation therefore supports a narrower conclusion: under the baseline-preserving preference, Full can favor scientifically admissible measurements that expose less of a misspecified baseline. This behavior is a limitation of the declared reference-preserving scope and should not be interpreted as a truth-recovery guarantee. The analysis is post hoc and conditional on this frozen reference, fixed promoted pool, and perturbation family, so it does not establish a universal robustness threshold or behavior under arbitrary reference misspecification.

\subsubsection{Is the prospective Full-action reduction stable to the numerical evaluator?}
\label{app:prospective-numerical-stability}
The prospective result uses a spatially discretized density--source evaluator, so we test whether its Law--Full contrast depends materially on the spatial raster, smoothing bandwidth, or linear-solver tolerance. The Law and $5\%$ Full geometries, observations, reference dynamics, moment reconstruction, and scientific-risk calculation remain frozen throughout; only the numerical action evaluator is changed.

\begin{table}[t]
\centering
\small
\begin{tabular}{lrrrl}
\toprule
Evaluator & Law action & Full action & Reduction (\%) & Paired Full$-$Law interval \\
\midrule
$128\times64$ & 1.50697 & 1.16113 & 22.95 & $[-0.388,-0.307]$ \\
$192\times96$ & 1.50973 & 1.16290 & 22.97 & $[-0.389,-0.308]$ \\
$256\times128$ & 1.51070 & 1.16352 & 22.98 & $[-0.389,-0.308]$ \\
Bandwidth $0.8\times$ & 1.63096 & 1.24506 & 23.66 & $[-0.430,-0.344]$ \\
Bandwidth $1.2\times$ & 1.38646 & 1.08121 & 22.02 & $[-0.344,-0.269]$ \\
PCG tolerance $10^{-10}$ & 1.50697 & 1.16113 & 22.95 & $[-0.388,-0.307]$ \\
\bottomrule
\end{tabular}
\caption{\textbf{Is the prospective Full-action reduction stable to the numerical evaluator?} Mean numerical Full actions for the frozen Law and $5\%$ Full geometries under spatial refinement, density--source bandwidth perturbation, and solver-tolerance perturbation. The reported paired intervals are for the Full-minus-Law action difference. No setting produces a numerical failure, and the action reduction remains between $22.02\%$ and $23.66\%$.}
\label{tab:prospective-numerical-stability}
\end{table}

As shown in Table~\ref{tab:prospective-numerical-stability}, spatial refinement changes the Law--Full reduction only from $22.95\%$ to $22.97\%$ and $22.98\%$ at $128\times64$, $192\times96$, and $256\times128$, respectively. Across the bandwidth and solver perturbations, the reduction remains between $22.02\%$ and $23.66\%$, and every paired action-difference interval remains strictly below zero. No numerical failures occur, the minimum ESS fraction is $0.168$, and the maximum Poisson residual is $1.02\times10^{-10}$. The reported Law--Full contrast is therefore stable across the tested numerical choices, although these checks do not by themselves establish convergence of the prospective density--source evaluator to the continuum Full functional. The manufactured-solution audit below separately tests continuum convergence for a periodic Full evaluator on an analytically tractable family without density estimation, smoothing, or observation noise.

\subsubsection{Does the periodic Full evaluator converge to analytically known continuum action?}
\label{app:full-manufactured-convergence}

The preceding refinement study tests stability of an empirical Law--Full contrast but does not isolate discretization error from density--source preprocessing. We therefore perform a deterministic manufactured-solution audit on a periodic domain using the exact information-projected Fourier family of Appendix~\ref{app:proof-residual-spectral-comparison}. Let the frozen reference be uniform with $u=0$, and for $k\in\{1,2,4\}$ define
\begin{equation}
q_{\lambda,k}(x)=\frac{\exp(\lambda\cos(kx))}{2\pi I_0(\lambda)},\qquad \lambda(t)=\frac12\sin(2\pi t).
\label{eq:manufactured-convergence-family}
\end{equation}
The corresponding normalized continuity forcing is
\begin{equation}
h_{k,t}(x)=\dot\lambda(t)\left(\cos(kx)-\frac{I_1(\lambda(t))}{I_0(\lambda(t))}\right).
\label{eq:manufactured-convergence-forcing}
\end{equation}
Writing $g(\lambda)$ for the exact transport metric in \eqref{eq:appendix-fourier-transport-metric}--\eqref{eq:appendix-fourier-metric-integral}, the continuum instantaneous action is
\begin{equation}
a_{\mathrm{full}}(t,k)=\frac{\dot\lambda(t)^2}{k^2}g(\lambda(t)).
\label{eq:manufactured-convergence-exact-action}
\end{equation}
The continuum benchmark is evaluated independently by adaptive one-dimensional quadrature and therefore does not use the discrete Poisson solver under test. Equation~\eqref{eq:manufactured-convergence-exact-action} also gives the exact integrated-action ratios $\Act(1)/\Act(2)=4$ and $\Act(1)/\Act(4)=16$.

The available periodic weighted-Poisson implementation is tested by extruding the one-dimensional family to the square torus as $q(x,y)=q(x)/(2\pi)$ and $s(x,y)=q(x)h(x)/(2\pi)$. Because the solution is independent of $y$, this extrusion leaves the continuum action unchanged. Exact cell-centered densities and sources are supplied directly, with no particle sampling, smoothing, density floor, source recentering, or observation noise. Spatial convergence is evaluated at $N\in\{32,64,128,256,512\}$ using $257$ time nodes, while temporal convergence is evaluated separately on the finest spatial grid.

\begin{table}[t]
\centering
\small
\begin{tabular}{rrr}
\toprule
Frequency $k$ & Relative error at $N=512$ & Empirical spatial order \\
\midrule
$1$ & $1.156\times10^{-5}$ & $2.00004$ \\
$2$ & $4.623\times10^{-5}$ & $2.00015$ \\
$4$ & $1.850\times10^{-4}$ & $2.00061$ \\
\bottomrule
\end{tabular}
\caption{\textbf{Manufactured-solution convergence of the periodic Full evaluator.} Relative errors compare the discrete time-integrated action with an independently quadrature-evaluated continuum benchmark. The empirical order is computed from the final two spatial refinements. At fixed $N$, the larger relative error at higher frequency is consistent with the expected dependence of a second-order spatial discretization on the dimensionless resolution $kh$.}
\label{tab:full-manufactured-convergence}
\end{table}

The evaluator exhibits approximately second-order spatial convergence for all three frequencies. At $N=512$, the numerical action ratios are $3.999861$ for $k=1$ versus $k=2$ and $15.997226$ for $k=1$ versus $k=4$, compared with the exact values $4$ and $16$. The isolated temporal-quadrature error is approximately $6.7\times10^{-13}$ already at $17$ nodes and reaches floating-point roundoff under further refinement, so the finest-grid continuum errors are dominated by spatial discretization. The largest physical Poisson residual, $1.79\times10^{-8}$, occurs for the coarsest $N=32$, $k=4$ case, where only eight cells resolve each wavelength; for $N\ge64$, the residual remains below $2.3\times10^{-11}$ and no solver failures occur.

This manufactured-solution audit shows that the tested periodic Full evaluator converges to analytically known continuum action and reproduces the inverse-frequency transport scaling predicted by the theory. 

\subsubsection{Ablation: Which parts of the prospective predictive interface matter?}
\label{app:prospective-predictor-sensitivity}

Prospective selection is conditional on three classes of predicted inputs: sensor-response means, single-particle response covariances, and scientific quantities used by the primary risk. This ablation isolates their roles using the $541$ labeled sensor geometries discovered by the original prospective search, of which $406$ satisfy the original geometric constraints, while holding the candidate pool fixed across perturbations. Fixing the pool removes variation from continuous design optimization and isolates how errors in the predictive interface change the ranking and admissibility of already-discovered measurement systems. The unperturbed condition is therefore a new selection restricted to this fixed pool rather than a reevaluation of the originally reported Law--Full pair, and its held-out action reduction need not equal that of the original prospective frontier. Under each perturbed interface, Law minimizes the corresponding scientific risk over the fixed pool and Full minimizes action within the condition-specific $5\%$ Law-relative risk band and numerical-support constraints; all selected geometries are then frozen before evaluation with the common unperturbed post-freeze validation reference.

\begin{figure}[t]
\centering
\includegraphics[width=\linewidth]{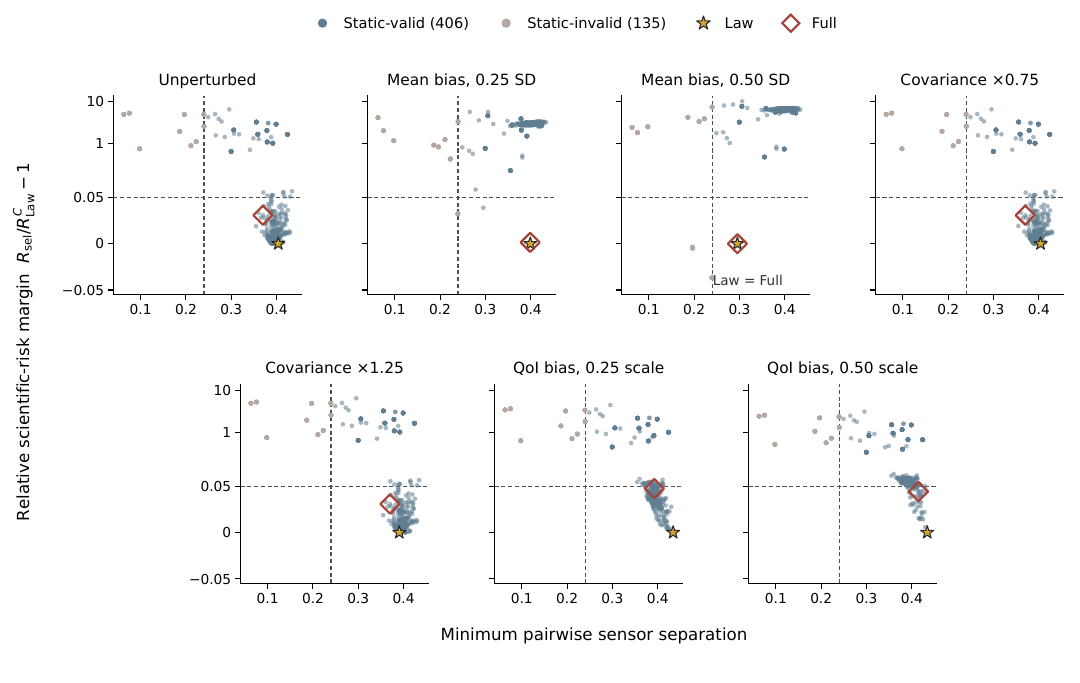}
\caption{\textbf{Predictor perturbations reshape the prospective design-time risk landscape in different ways.} Each panel shows the same fixed pool of $541$ candidate sensor geometries, with minimum pairwise sensor separation on the horizontal axis and scientific-risk margin $R_{\mathrm{sel}}/R_{\mathrm{Law}}^{\mathcal C}-1$ relative to the condition-specific candidate-pool Law anchor on the vertical axis. The vertical dashed line marks the minimum allowed separation $0.24$, and the horizontal dashed line marks the $5\%$ Law-relative risk allowance. Highlighted markers denote the Law and Full designs selected from the fixed pool. The tested covariance rescalings leave the landscape nearly unchanged, whereas sensor-mean perturbations substantially reorganize the relative risk ordering and move an increasing fraction of the below-cap candidates into the separation-invalid region. QoI-target perturbations alter the scientific-risk landscape without changing the underlying reconstructed trajectories. Because each panel is normalized by its own Law anchor, vertical positions describe within-condition relative risk rather than absolute risk across perturbations. The two dashed thresholds alone do not define final feasibility, which also requires the condition-dependent numerical and support criteria.}
\label{fig:prospective-predictor-landscape}
\end{figure}

We consider three perturbation families. First, we add a fixed bias direction to the predicted sensor-response means at amplitudes $0.25$ and $0.50$ times the local response standard deviation. Second, we rescale the predicted single-particle sampling covariance by factors $0.75$ and $1.25$, leaving detector noise unchanged. Third, we perturb the predicted scientific-QoI targets by $0.25$ and $0.50$ frozen QoI scales along a fixed direction. Table~\ref{tab:prospective-predictor-sensitivity} summarizes the resulting finite-pool selections and their held-out validation performance; the reported rank correlation is computed over the same $406$ statically admissible geometries in every condition. Figure~\ref{fig:prospective-predictor-landscape} complements this summary by showing how each perturbation reorganizes the design-time scientific-risk landscape over the common candidate pool as a function of minimum sensor separation.

\begin{table}[t]
\centering
\small
\setlength{\tabcolsep}{4pt}
\resizebox{\linewidth}{!}{%
\begin{tabular}{lrrrr}
\toprule
Predictor condition & Feasible & Risk-rank $\rho$ &
\shortstack{Held-out risk\\change (\%)} &
\shortstack{Held-out action\\reduction (\%)} \\
\midrule
Unperturbed & 246 & 1.000 & 2.030 & 24.324 \\
Mean bias, $0.25$ local SD & 16 & $-0.294$ & $-0.141$ & 0.327 \\
Mean bias, $0.50$ local SD & 1 & $-0.355$ & 0.000 & 0.000 \\
Sampling covariance $\times0.75$ & 246 & 1.000 & 2.030 & 24.324 \\
Sampling covariance $\times1.25$ & 246 & 1.000 & 2.279 & 23.133 \\
QoI-target bias, $0.25$ scale & 224 & 0.800 & $-1.749$ & 26.747 \\
QoI-target bias, $0.50$ scale & 46 & 0.746 & 0.641 & 25.642 \\
\bottomrule
\end{tabular}%
}
\caption{\textbf{Sensitivity of prospective selection to the aggregate predictive interface.} All conditions rank the same fixed pool of $541$ candidate geometries. ``Feasible'' is the number of candidates satisfying the original numerical/support requirements and the condition-specific $5\%$ Law-relative scientific-risk band. Risk-rank $\rho$ is the Spearman correlation with the unperturbed condition over the $406$ statically admissible geometries. Held-out quantities compare the Full and Law designs selected under the same predictor condition on the common post-freeze validation experiment.}
\label{tab:prospective-predictor-sensitivity}
\end{table}

As summarized in Table~\ref{tab:prospective-predictor-sensitivity} and visualized in Figure~\ref{fig:prospective-predictor-landscape}, the covariance perturbations leave the design landscape essentially unchanged. At both covariance scales, the feasible set remains at $246$ candidates, the scientific-risk rankings have correlation above $0.9999$ with the unperturbed predictor, and the Full-selected geometry is unchanged. The corresponding held-out action reductions are $24.32\%$ and $23.13\%$. Thus, within the tested range, moderate covariance rescaling has little effect on either admissibility or the secondary action ranking.

Systematic errors in the predicted sensor means have a markedly different effect. At the $0.25$-SD perturbation, all $406$ statically admissible candidates continue to pass the non-risk numerical and support requirements, but only $16$ remain within the $5\%$ scientific-risk band. The scientific-risk ranking over these valid geometries is negatively correlated with the unperturbed ranking, $\rho=-0.294$, and the selected Law and Full geometries become nearly identical, leaving only a $0.33\%$ held-out action reduction. At the $0.50$-SD perturbation, only one candidate remains feasible, so Law and Full necessarily coincide. This stronger perturbation also affects support: only $67$ statically valid candidates satisfy the non-risk gates in both the perturbed and unperturbed conditions, yet their scientific-risk rankings are still strongly reversed, with $\rho=-0.761$. The sensitivity therefore persists after excluding both geometrically invalid designs and condition-dependent support failures.

The same mean perturbations also shift the apparent low-risk region of the fixed pool toward closely spaced sensors. Before applying the geometric constraint, none of the $246$ unperturbed below-cap candidates violates the minimum-separation rule, compared with $15$ of $31$ candidates under the $0.25$-SD perturbation and $30$ of $31$ under the $0.50$-SD perturbation. Candidate-level decompositions show that the effect is not a uniform offset of the risk values: the perturbed sensor means change reconstructed moment trajectories and projected scientific quantities in a geometry-dependent way. For example, the original Law geometry changes from design-time risk $1.13$ to $6.63$ and $19.35$ under the two mean perturbations, whereas the geometry selected under the larger perturbation changes from risk $8.52$ in the unperturbed condition to $2.53$. The dominant changes in the inspected candidates arise from squared mean QoI errors rather than trial-to-trial variance.

Perturbing the scientific-QoI targets produces a third, more localized mechanism. For a fixed geometry, these perturbations leave predicted sensor responses, reconstructed moments, projected law trajectories, and Full actions unchanged; only the scientific admissibility criterion changes. Consequently, the Full-action ranking is identical to the unperturbed ranking on the common pre-risk-valid pool, while the scientific-risk rank correlation decreases to $0.800$ and $0.746$ and the feasible set contracts to $224$ and $46$ candidates. The selected Full designs nevertheless retain held-out action reductions of $26.75\%$ and $25.64\%$. This separates sensitivity of the scientific-risk constraint from sensitivity of the law-level action itself.

The held-out results also illustrate the scope of a relative risk constraint. Every Full design remains within $5\%$ of the Law design selected under the same perturbed predictor, but this within-condition comparison does not constrain absolute performance relative to the unperturbed design problem. In particular, the held-out Full risks under the two sensor-mean perturbations are $4.70$ and $8.94$, compared with $1.22$ for the unperturbed Law design. Thus, the relative constraint continues to control the Full--Law tradeoff under a given predictive model, while systematic predictor error can shift the scientific-risk landscape shared by both methods.

The fixed-pool study identifies three forms of predictor sensitivity. The tested covariance rescalings leave both scientific and action rankings nearly unchanged; QoI-target shifts alter scientific admissibility while preserving the law-level action ranking; and sensor-mean bias changes the reconstructed moments themselves, substantially reorganizing both the scientific-risk landscape and the available Law--Full tradeoff. These results are conditional on the fixed discovered candidate pool, one selection reference and one separately trained post-freeze validation reference, and the specified perturbation directions.

\subsubsection{Are the skyrmion action reductions robust to numerical approximation?}
\label{app:skyrmion_ablations}

We test two possible explanations for the skyrmion action reductions: sensitivity to the potential dictionary used during design optimization, and disproportionate influence of a trial with concentrated importance weights. Sensor geometries remain frozen throughout, and all 32 paired validation trials remain in the primary analysis.

\paragraph{Does the action reduction persist under a richer potential space?}
A design optimized against a restricted Galerkin evaluator could appear favorable because the evaluator misses correction directions that matter more for that design. Proposition~\ref{prop:galerkin-monotonicity} makes this issue explicit: for a fixed projected law and continuity forcing, the exact restricted variational action is a lower bound on Full, and nested potential-space enrichments can only tighten that bound. This property alone does not show that the Base dictionary captures enough of the Full correction for reliable design comparison, so we re-evaluate the frozen frontier using four potential spaces: Base is the original dictionary of 248 one-particle Fourier functions and 32 radial pair functions, Fourier-400 and Fourier-560 retain these coordinates and append 120 and 280 Fourier functions respectively, while Interaction-416 instead appends 136 pre-specified distinct-particle products of low-frequency Fourier channels, testing pair-dependent potential directions rather than additional one-particle resolution. Within each trial and geometry, reconstructed moments, projected weights and continuity forcing are held fixed across evaluators, with fitting and audit banks kept separate. Original coordinates, normalization, solver settings and certification thresholds are unchanged; only the potential space is enriched.

For evaluator $E$, let $\overline A_E(\etavec)$ denote mean fitted action over the $32$ trials. We compare
\begin{equation}
\widehat D_{p,E}=1-\frac{\overline A_E(\etavec_{\mathrm{Full},p})}{\overline A_E(\etavec_{\mathrm{Law}})},
\qquad
\Delta_{p,E}=100\bigl(\widehat{D}_{p,E}-\widehat{D}_{p,\mathrm{Base}}\bigr),
\label{eq:ablation_contrast}
\end{equation}
where $100\widehat{D}_{p,E}$ is the percentage action reduction and $\Delta_{p,E}$ is its change in percentage points relative to Base. Reductions are ratios of mean actions, not means of trial-wise ratios. The 3\% and 4\% allowances select identical coordinates and are evaluated once.

\begin{table}[t]
\centering
\small
\begin{tabular}{lrrrr}
\toprule
Risk allowance & Base & Fourier-400 & Fourier-560 & Interaction-416 \\
\midrule
1\%       & 3.51 & 3.51 & 3.52 & 3.53 \\
2\%       & 6.91 & 6.92 & 6.93 & 6.95 \\
3\%       & 9.38 & 9.40 & 9.41 & 9.43 \\
4\%       & 9.38 & 9.40 & 9.41 & 9.43 \\
5\%       & 12.22 & 12.25 & 12.28 & 12.31 \\
\bottomrule
\end{tabular}
\caption{\textbf{Does the action reduction persist under a richer potential space?}
Percentage reductions in mean fitted action relative to Law under each evaluator. All 32 trials are retained, including failed-certification cases; the 3\%/4\% row represents one shared geometry. Fourier-400 and Fourier-560 refine the Fourier dictionary, while Interaction-416 adds interaction features. Fourier-560 has six held-out energy-check failures, and ESS support failures remain under every evaluator.}
\label{tab:skyrmion_basis}
\end{table}

The action reduction persists at every operating point under all four evaluators (Table~\ref{tab:skyrmion_basis}), with a largest change of only $0.096$ percentage points relative to Base. Full retains lower fitted action than Law in every individual trial under every evaluator, and the ordering across allowances is unchanged. Table~\ref{tab:skyrmion_absolute_enrichment} additionally shows the absolute-action change under the nested Base-to-Fourier-560 enrichment: both frozen designs acquire a small amount of additional fitted action while their relative contrast remains essentially unchanged.

\begin{table}[t]
\centering
\small
\begin{tabular}{lrrr}
\toprule
Design & Base ($K=280$) & Fourier-560 ($K=560$) & Increase (\%) \\
\midrule
Law & $0.024433$ & $0.024581$ & $0.61$ \\
Full ($5\%$ allowance) & $0.021449$ & $0.021564$ & $0.54$ \\
\bottomrule
\end{tabular}
\caption{\textbf{Absolute fitted action under nested Galerkin enrichment.} Mean fitted actions over the same $32$ validation trials. In agreement with the monotonicity of the underlying restricted variational problem in Proposition~\ref{prop:galerkin-monotonicity}, enlarging the Fourier potential space raises the fitted action for both frozen designs. The corresponding Full-versus-Law reduction changes only from $12.22\%$ to $12.28\%$.}
\label{tab:skyrmion_absolute_enrichment}
\end{table}

Thus, the observed contrasts are stable under the tested Fourier and interaction enrichments, although these checks do not bound correction energy in untested potential directions.

Contrast stability is distinct from numerical certification. The positive, ordered frontier is reproduced by Fourier-400 and Interaction-416, both of which pass the reported held-out residual checks, so the enrichment result does not rely on the largest Fourier-560 system. Fourier-560 exceeds the held-out energy threshold in six of $160$ unique trial/design evaluations, identifying a numerical-certification limit of that larger basis rather than a reversal of the design ordering. Separately, the low-support validation pair fails one ESS support check for Law and all four distinct Full geometries under every evaluator. We therefore do not claim full numerical certification or continuum convergence; Proposition~\ref{prop:galerkin-monotonicity} characterizes what convergence of the underlying restricted variational problem would require, while the present experiment supports only that the observed Full-versus-Law contrast is stable under the tested one-particle and interaction-space enrichments.

\paragraph{Does the low-support trial drive the aggregate action reduction?}
We examine the low-support validation pair's influence using Base at the $5\%$ allowance. Retaining all $32$ pairs gives a $12.215\%$ action reduction; excluding only this prespecified pair gives $12.380\%$, a change of $+0.165$ percentage points. Across all leave-one-pair-out calculations, the reduction ranges from $12.083\%$ to $12.442\%$. The low-support validation pair ranks third in absolute deletion influence, but removing any single pair preserves the aggregate action reduction. Thus the low-support trial does not drive the reported contrast: every leave-one-pair-out analysis preserves an action reduction above $12\%$. These deletions are sensitivity checks rather than replacements for the primary analysis, and they do not resolve the trial's separate support-certification issue.

We additionally evaluate both frozen geometries on independent reference banks, preserving their saved moment trajectories and Base dictionary while recalibrating projected weights and forcing on each bank. Increasing the fitting bank from 65,536 to 131,072 and 262,144 configurations changes the low-support validation pair's fitted-action reduction from $9.685\%$ to $8.421\%$ and $8.877\%$, respectively. The trial-specific estimate varies with reference-bank size, but its influence on the aggregate contrast is small under this diagnostic: replacing only this pair's actions by the largest-bank estimates changes the overall reduction by $-0.048$ percentage points.

\subsubsection{Sensitivity to the transport--reaction metric}
\label{app:unbalanced-kappa-ablation}

\begin{figure}[t]
\centering
\includegraphics[width=0.72\linewidth]{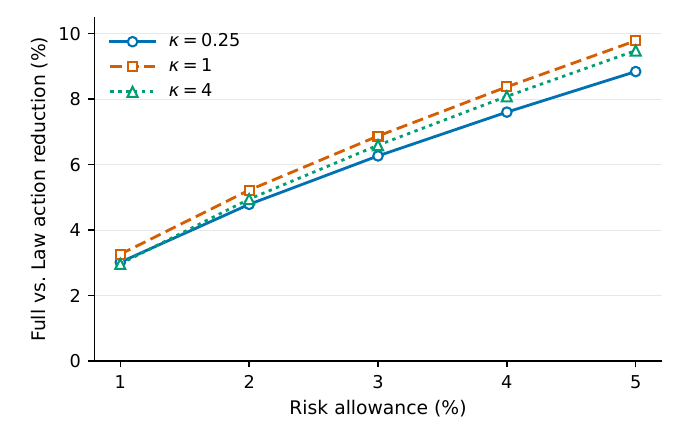}
\caption{\textbf{The within-metric Full-versus-Law action reduction persists across reaction penalties.} Each curve reports the percentage reduction in mean Full action relative to Law at a fixed $\kappa$. Scientific observations are shared across the sweep, while $\kappa$ changes only the transport--reaction correction metric; vertical comparisons of absolute action across $\kappa$ are therefore not meaningful.}
\label{fig:unbalanced-kappa-sensitivity}
\end{figure}

In the unbalanced setup, the reaction penalty $\kappa$ determines the relative geometry assigned to movement and reaction and is therefore a modeling choice rather than a parameter whose absolute action values can be compared across settings. We test whether the within-metric Full-versus-Law contrast depends strongly on this choice by evaluating the frozen construction at $\kappa\in\{0.25,1,4\}$ on the same shared physical validation bank. As shown in Table~\ref{tab:unbalanced-kappa-sensitivity} and Figure~\ref{fig:unbalanced-kappa-sensitivity}, Full retains lower mean action than Law at every allowance under all three metrics. The reduction ranges from $2.97\%$ to $3.26\%$ at the $1\%$ allowance and from $8.85\%$ to $9.79\%$ at the $5\%$ allowance, with the intermediate allowances showing a similar pattern, indicating that the within-metric Full-versus-Law contrast is not tied to the default reaction penalty.

\begin{table}[t]
\centering
\small
\begin{tabular}{rrrr}
\toprule
Allowance & $\kappa=0.25$ & $\kappa=1$ & $\kappa=4$ \\
\midrule
$1\%$ & 3.02 & 3.26 & 2.97 \\
$2\%$ & 4.79 & 5.22 & 4.95 \\
$3\%$ & 6.27 & 6.88 & 6.59 \\
$4\%$ & 7.61 & 8.38 & 8.09 \\
$5\%$ & 8.85 & 9.79 & 9.49 \\
\bottomrule
\end{tabular}
\caption{\textbf{Sensitivity of the Full-versus-Law contrast to the transport--reaction metric.} Entries are percentage reductions in mean Full action relative to Law, computed separately within each value of $\kappa$. Because $\kappa$ changes the correction metric, absolute action values are not compared across columns.}
\label{tab:unbalanced-kappa-sensitivity}
\end{table}

\subsubsection{Unbalanced mechanism decompositions}
\label{app:unbalanced-mechanism-ablation}

\begin{table}[t]
\centering
\small
\setlength{\tabcolsep}{4pt}
\resizebox{\linewidth}{!}{%
\begin{tabular}{rrrrrr}
\toprule
$\kappa$ &
\shortstack{Move / Full\\(\%)} &
\shortstack{Reaction / Full\\(\%)} &
\shortstack{Hidden / Full\\(\%)} &
\shortstack{Full vs. Tangent\\(\%)} &
\shortstack{Full vs. GlobalReaction\\(\%)} \\
\midrule
0.25 & 21.99--22.65 & 77.35--78.01 & 43.15--43.80 & 1.95--6.12 & 2.23--6.28 \\
1    & 23.97--24.21 & 75.79--76.03 & 64.58--65.51 & 2.57--6.30 & 2.16--6.00 \\
4    & 38.79--39.33 & 60.67--61.21 & 73.51--74.62 & 2.27--5.62 & 1.76--5.20 \\
\bottomrule
\end{tabular}%
}
\caption{\textbf{Unbalanced mechanism summary across the $1\%$--$5\%$ allowances.} The first three quantities describe the Full-selected geometry and are reported as ranges over the five allowances. Move and Reaction form one decomposition of Full, while Tangent and Hidden form a separate orthogonal decomposition. The final two columns report the range of reductions in the common Full cross-score obtained by Full relative to the Tangent- and GlobalReaction-selected geometries.}
\label{tab:unbalanced-mechanism-summary}
\end{table}

\begin{figure}[t]
\centering
\includegraphics[width=\linewidth]{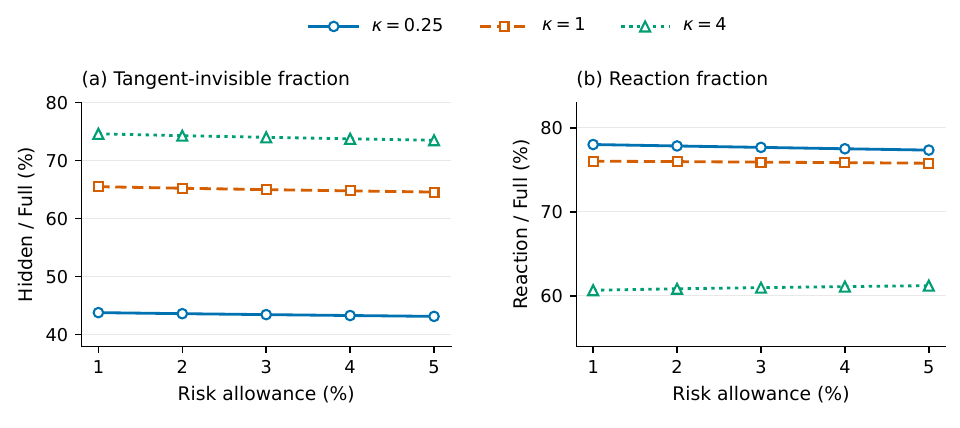}
\caption{\textbf{Mechanism decomposition of the unbalanced Full correction.} \textbf{Left:} fraction of Full action lying in the Tangent-invisible Hidden component. \textbf{Right:} fraction of Full action allocated to reaction rather than spatial movement. Fractions are computed within each fixed transport--reaction metric; the figure does not compare absolute action values across $\kappa$.}
\label{fig:unbalanced-mechanism-decomposition}
\end{figure}

The screened Full correction admits two complementary decompositions established in the finite-measure theory. First, the Full energy in \eqref{eq:ub-full-action-energy} separates into Move and Reaction contributions, corresponding respectively to the transport and reaction terms of the optimal correction. Second, the orthogonal decomposition in \eqref{eq:ub-action-decomposition} writes Full as Tangent plus Hidden, where Tangent is the minimum correction visible through the measured raw moment rates and Hidden is the remaining law-level component. GlobalReaction provides a separate restriction diagnostic: unlike Tangent, which restricts the correction through the measured moment-rate equations, GlobalReaction restricts the admissible reaction field while retaining transport. Table~\ref{tab:unbalanced-mechanism-summary} and Figure~\ref{fig:unbalanced-mechanism-decomposition} show that the Tangent-invisible component remains substantial throughout the tested regime, while the common-Full cross-score of the Full-selected geometry is lower than those of both restricted comparators in every tested $\kappa$--allowance combination.

\section{Experimental details and reproducibility}
\label{app:experiments}

\paragraph{Shared experimental conventions.} For particle-based studies, a trial observes $\bm y_k=n^{-1}\sum_{i=1}^{n}\Phivec_{\etavec}(X_{t_k}^{(i)})+\sigma_{\mathrm{obs}}\bm\varepsilon_k$ with $\bm\varepsilon_k\sim\mathcal N(0,I_R)$; one microscopic state is a tracer position in the transport studies and a complete interacting configuration in the skyrmion study. Reconstruction produces a differentiable moment trajectory that is information-projected relative to the frozen reference. Except for the stricter skyrmion development constraint stated below, allowance $p\%$ means $R_{\mathrm{sel}}(\etavec)\le(1+p/100)R_{\mathrm{Law}}$, where $R_{\mathrm{Law}}$ is the selection risk of the numerically optimized Law design. For a fixed reference, $\widehat D_p=1-\overline A(\etavec_{\mathrm{Full},p})/\overline A(\etavec_{\mathrm{Law}})$, with $\overline A$ the trial-mean numerical Full action; when several frozen references are used, reference-wise effects are computed before aggregation as specified by the study. Selection feasibility, held-out scientific risk, numerical certification, and action improvement are reported separately. Numerical Full denotes the declared raster or Galerkin approximation, and absolute action values are not compared across systems or reference regimes. For normalized empirical projection weights $p_1,\ldots,p_N$, $\operatorname{rESS}=(N\sum_i p_i^2)^{-1}$; the Gaussian and double-gyre studies require $\operatorname{rESS}\ge0.03$ for their declared selection-time support checks and the skyrmion study requires $0.05$. These thresholds diagnose support concentration rather than empirical moment feasibility, which is characterized separately by Proposition~\ref{prop:empirical-iprojection-feasibility}.

\paragraph{Reference flow matching.}
The Gaussian, retrospective and prospective double-gyre, and skyrmion studies use reference velocities trained by flow matching on endpoint pairs. For an experiment-specific endpoint coupling $(A_0,A_1)\sim\Pi$, bridge $X_t^{\mathrm{br}}$, and $T\sim\mathrm{Unif}[0,1)$, we minimize
\begin{equation}
\mathcal L_{\mathrm{FM}}(\theta)=\E\!\left[\left\|f_\theta(X_T^{\mathrm{br}},T)-\left.\partial_tX_t^{\mathrm{br}}\right|_{t=T}\right\|_2^2\right].
\label{eq:reference-flow-matching}
\end{equation}
For the Gaussian benchmark, $X_t^{\mathrm{br}}=(1-t)A_0+tA_1+0.15\sin(\pi t)Z$ with independently sampled endpoint states and $Z\sim\mathcal N(0,I)$. For both double-gyre studies, independently sampled empirical endpoints are first mapped to box-logit coordinates and connected by the noiseless linear bridge $X_t^{\mathrm{br}}=(1-t)A_0+tA_1$; the learned latent velocity is mapped back to physical coordinates by the logistic Jacobian. For the skyrmion reference, endpoint configurations use the frozen particle-matched coupling and the periodic bridge $X_t^{\mathrm{br}}=[A_0+t\Delta+0.01\sin(\pi t)Z]\bmod L$, where $\Delta$ is the minimum-image displacement. After training, the reference is frozen and $\widetilde Q_t$ is represented by deterministic forward rollout of the learned velocity from the initial law; the training bridge itself is not used as the reference marginal path. The unbalanced experiment instead uses the fixed analytical reference described in Appendix~\ref{app:exp-unbalanced}.

\subsection{Information boundaries}
\label{app:information-boundary}

The studies differ in the target information available during measurement design. Table~\ref{tab:information-boundary} summarizes these regimes. In particular, the prospective double-gyre optimizer accesses intermediate target information only through a frozen aggregate interface: the interface is constructed offline from a separate microscopic simulation bank, but the underlying intermediate particles and target simulator are not available during sensor optimization.

\begin{table}[t]
\centering
\footnotesize
\setlength{\tabcolsep}{3pt}
\renewcommand{\arraystretch}{1.12}
\begin{tabularx}{\linewidth}{
>{\raggedright\arraybackslash}p{2.25cm}
>{\raggedright\arraybackslash}X
>{\raggedright\arraybackslash}X}
\toprule
\textbf{Study} & \textbf{Information available during selection} & \textbf{Information introduced after design freeze} \\
\midrule
Gaussian mixture & Analytic target law, target-derived moments and covariances, noisy finite-sample observations, population-risk screen, and one frozen endpoint-trained reference. & Fresh target draws, particle sampling, and detector noise from the same analytic family. The reference is unchanged. \\
\addlinespace
Retrospective double-gyre & Frozen microscopic target simulation bank and three endpoint-trained references; candidate observations and scientific-risk quantities are evaluated directly from the target bank. & Fresh particle resampling and detector noise, conditional on the same target bank and reference ensemble. \\
\addlinespace
Prospective double-gyre & Endpoint ensembles, one frozen endpoint-trained reference, and fixed aggregate sensor-response means, cross-sensor covariances, and scientific-QoI predictions. These aggregate fields are precomputed from a separate microscopic simulation bank and its intermediate particles are not exposed to the optimizer. & A separately seeded evaluation reference trained from the same endpoint dataset, together with new microscopic validation trajectories, sampling, and detector noise for action and scientific-risk validation. \\
Skyrmions & A fixed development set of interacting configurations, one frozen endpoint-trained permutation-equivariant reference, aggregate sensor observations, and the nine permutation-invariant scientific features used to evaluate development risk. & A separate set of $32$ fresh paired measurement trials for scientific-risk and action validation. The reference model and selected sensor geometries remain frozen. Richer potential dictionaries, leave-one-pair-out analyses, and larger reference banks are introduced only as post-selection numerical sensitivity checks. \\
\addlinespace
Unbalanced Gaussian & A fixed analytical reference and finite-measure simulator, together with development observations and simulator truth used for scientific-risk evaluation. & A fresh bank of $128$ paired trials with new physical nuisances, particle observations, detector noise, and log-mass noise. The same frozen bank is reused across the $\kappa$ sensitivity analysis. \\
\bottomrule
\end{tabularx}
\caption{\textbf{Information boundaries of the reported studies.} ``After design freeze'' denotes information unavailable when optimizing the reported sensor geometries. Fresh evaluation is conditional on the quantities explicitly retained in the corresponding row.}
\label{tab:information-boundary}
\end{table}

\subsection{Gaussian-mixture criterion comparison}
\label{app:exp-analytic}

\paragraph{Gaussian system and sensor family.} For $d(\beta)=(\cos\beta,\sin\beta)^\top$, let $$g_\beta(x)=\tfrac12\mathcal N(x;1.5d(\beta),0.3^2I_2)+\tfrac12\mathcal N(x;-1.5d(\beta),0.3^2I_2)$$ and $$\bar p_t^\alpha=(1-t)^2g_0+2t(1-t)g_\alpha+t^2g_{\pi/2}$$ for $\alpha\in[\pi/6,\pi/3]$. Benchmark truth is the normalized mass of $\bar p_t^\alpha$ on a $51\times51$ grid in $[-3.2,3.2]^2$. Two sensors have centers $z_j=1.5d(\theta_j)$ on the radius-$1.5$ ring, angular separation at least $20^\circ$, and responses $\Phi_{\etavec,j}(x)=\exp(-\|x-z_j\|^2/(2\ell^2))$ with $\ell=0.45$. Each trial observes $100$ particles at eleven acquisition times with $\sigma_{\mathrm{obs}}=0.01$, and endpoint-anchored quadratic generalized least squares reconstructs moments on $21$ scientific times.

\paragraph{Reference, selection, and evaluation protocol.} The reference is trained using \eqref{eq:reference-flow-matching} with four width-$128$ SiLU hidden layers for $12{,}000$ Adam steps, batch size $2{,}048$, initial learning rate $10^{-3}$ with cosine decay to $5\times10^{-5}$, and gradient clipping at $10$. Scientific risk is Gaussian-kernel MMD$^2$ with bandwidth $0.55$ on the unobserved time nodes. The frozen Law anchor is $R_{\mathrm{Law}}=0.0662$ with population-risk ceiling $0.0692$. Law, Tangent, Full, residual $L^2$, forward KL, and static $W_2$ are optimized at allowances $0.5,1,2,3,4,5\%$ using the same information-projected construction; each receives seven common starts and $40$ Adam steps per allowance with learning rate $0.012$ and constraint penalty $10^4$, together with the same number of final candidate evaluations. Selection uses $4$ trials for gradient estimation, $32$ for scientific-risk evaluation, $16$ for preliminary action evaluation, and $64$ for final objective evaluation. Full uses a $41^2$ optimization raster and a $101^2$ final evaluation raster with no added operator-density floor. Frozen geometries are evaluated on $64$ paired holdout trials. Static $W_2$ was added after the original five-selector holdout had been inspected. It received the same search opportunity but reused that holdout and is therefore post hoc.

\paragraph{Allowance sweep and criterion-specific selections.} Across the prescribed risk allowances, the secondary targets select different sensor geometries despite operating under the same scientific-risk restriction. At the $0.5\%$ allowance, the Full-selected design reduces held-out Full action by $30.67\%$ relative to Law; from $1\%$ through $5\%$, Full selects the same geometry and gives a $33.59\%$ reduction. The $1\%$ cross-objective results are reported in Table~\ref{tab:exp-gaussian-criteria}. Appendix~\ref{app:gaussian-common-candidate-comparison} repeats the selection over a common promoted candidate pool and again produces criterion-specific geometries, showing that the separation is not solely a consequence of criterion-specific optimization trajectories. All reported designs satisfy their selection-time scientific and numerical requirements; held-out scientific-risk transfer is assessed separately in Appendix~\ref{app:risk-feasibility-ablation}, where the $0.5\%$ point remains unresolved under simultaneous one-sided inference and the $1\%$--$5\%$ points are supported below their corresponding held-out caps.

\subsection{Prospective double-gyre study}
\label{app:exp-vortices-prospective}

\paragraph{Double-gyre system.} Passive tracers evolve on $\Omega=[0,2]\times[0,1]$ under the time-dependent double-gyre with $a(\tau)=0.25\sin(2\pi\tau/10)$, $f(x_1,\tau)=a(\tau)x_1^2+(1-2a(\tau))x_1$, $v(x,\tau)=0.1\pi(-\sin(\pi f)\cos(\pi x_2),\,\cos(\pi f)\sin(\pi x_2)\partial_{x_1}f)^\top$, and $dX_t/dt=10v(X_t,10t)$. The initial population is a mixture of four truncated Gaussian components with coordinate standard deviation $0.07$ and a $10\%$ uniform background. The retrospective and prospective studies use this common transport system; their information boundaries and selection procedures differ as described below and in Appendix~\ref{app:exp-vortices}.

\paragraph{Information boundary and scientific objective.}
The optimizer receives endpoint ensembles, one frozen endpoint-trained reference, and a frozen aggregate predictive interface constructed offline from a separate microscopic simulation bank; the underlying intermediate particles, hidden trajectories, and target simulator are not exposed during sensor optimization. On a $129\times65$ aggregate grid, the interface supplies the sensor-response mean $\widehat{\bm\mu}_{\etavec}(t)$, which provides the design-stage moment trajectory $\cvec_{\etavec}(t)$ used by FIDE, together with the full single-particle response covariance $\widehat\Sigma_{\etavec}(t)$ and predicted trajectories of the scientific quantities used by the design-stage risk. For $2{,}000$ sampled particles and detector-noise standard deviation $0.005$, selection uses $\E[\bm y_t\mid\etavec]=\widehat{\bm\mu}_{\etavec}(t)$ and $\Cov(\bm y_t\mid\etavec)=\widehat\Sigma_{\etavec}(t)/2{,}000+0.005^2I_4$, retaining cross-sensor sampling dependence. The scientific features are $\bm\xi(x_1,x_2)=(x_1/2,x_2,(x_1/2)^2,x_2^2,x_1x_2/2)^\top$. Writing $m_j(t)$ for their frozen aggregate predictions and $s_j=\max\{\operatorname{std}_t m_j(t),0.02\}$, selection risk is $R_{\mathrm{sel}}(\etavec)=B^{-1}\sum_{b=1}^{B}\sum_{q=0}^{20}\omega_q\sum_{j=1}^{5}[(\E_{Q_{b,t_q}^{\etavec}}[\xi_j]-m_j(t_q))/s_j]^2$, where $\omega_q$ are normalized trapezoidal weights on the $21$ scientific times. The final Law anchor is $R_{\mathrm{Law}}=1.1344$ and is obtained entirely from selection-stage information.

\paragraph{Optimization, freeze, and validation.} The selection reference uses $32{,}768$ rollout particles; Law and Tangent use $16$ starts and $60$ Adam steps, while Full uses $16$ starts and $50$ steps followed by higher-resolution evaluation of eight candidates and local refinement of three. Final selection uses $32$ common trials and a $128\times64$ Full raster at all $21$ scientific times. After the frontier for $p\in\{1,2,3,4,5\}$ is frozen, a separately seeded validation reference trained from the same endpoint dataset and $64$ paired trials with fresh microscopic observations and hidden scientific targets are introduced, with no validation result used to alter the geometries or Law anchor. The Full-versus-Law action reductions are $13.71\%$, $19.05\%$, $19.05\%$, $19.05\%$, and $22.95\%$; the $2\%$--$4\%$ allowances select the same Full geometry and have simultaneous $95\%$ interval $[16.72\%,21.37\%]$, while the $5\%$ interval is $[20.63\%,25.27\%]$. At $5\%$, Full uses $2.834\%$ additional selection risk and has mean Full action $1.16113$, compared with $1.50697$ for Law and descriptive Tangent cross-score $1.32591$. Held-out scientific-risk increases are $1.060\%$, $1.022\%$, $1.022\%$, $1.022\%$, and $1.711\%$; simultaneous one-sided inference leaves the $1\%$ point unresolved and supports the $2\%$--$5\%$ points below their corresponding caps.

\subsection{Interacting-skyrmion study}
\label{app:exp-skyrmions}

\paragraph{System, observations, and scientific objective.} One microscopic state is a configuration $X_t=(r_{1,t},\ldots,r_{16,t})\in\mathcal X_{\mathrm{sky}}:=\Omega_{\mathrm{per}}^{16}$, where $\Omega_{\mathrm{per}}=(\R/2\mathbb Z)\times(\R/\mathbb Z)$, so $\dim\mathcal X_{\mathrm{sky}}=32$ and $\dot r_{i,t}=V_i(X_t,t)$ for $i=1,\ldots,16$. The implemented simulator uses $24$ Euler--Maruyama substeps per scientific interval, interaction strength $0.035$ and range $0.16$, pinning strength $0.055$ and width $0.10$, Magnus coefficient $0.32$, and diffusion amplitude $0.006$. Four width-$0.12$ aggregate sensors have periodic pairwise separation at least $0.20$; each measurement averages $512$ configurations and adds detector noise with standard deviation $0.002$ at seven acquisition times, while reconstruction is evaluated at $13$ scientific nodes. Each fresh validation truth contains $5{,}000$ configurations. Scientific risk compares nine permutation-invariant structural features—four radial-pair summaries, four structure factors, and one hexatic-order summary—using fixed covariance whitening and normalized trapezoidal time weights.

\paragraph{Reference, configuration-space approximation, and development protocol.} The frozen reference is an endpoint-only permutation-equivariant flow trained for $6{,}000$ Adam steps with batch size $512$; validation does not retrain or replace this reference. Full is approximated in configuration space by $\psi_{t,K}(X)=\sum_{a=1}^{K}\beta_a(t)\chi_a(X)$ and $\delta_{t,K}(X)=-\nabla_X\psi_{t,K}(X)$ with an immutable $K=280$ dictionary containing $248$ one-particle Fourier functions and $32$ radial-pair functions. Risk, Galerkin fitting, and independent audit banks contain $32{,}768$, $65{,}536$, and $262{,}144$ complete configurations, respectively. The solver uses float64 arithmetic with Jacobi equilibration and relative eigenvalue cutoff $10^{-12}$, and the prespecified support requirement is $\operatorname{rESS}\ge0.05$. The reported frontier is optimized on $17$ previously inspected development draws using the stricter paired development-risk ceiling $R_{\mathrm{dev}}(\etavec)\le(1+0.8p/100)R_{\mathrm{Law}}$ for nominal allowances $p\in\{1,2,3,4,5\}$. Development action reductions are $4.04\%$, $7.94\%$, $10.70\%$, $10.70\%$, and $12.87\%$; the $3\%$ and $4\%$ operating points use the same selected geometry. These values describe the development frontier only and are separated from the fresh validation analysis.

\paragraph{Fresh validation and numerical sensitivity.} After selection, the frozen frontier is evaluated on $32$ new paired trials, with the reference model and sensor geometries unchanged. Full reduces mean fitted Galerkin action relative to Law by $3.51\%$, $6.91\%$, $9.38\%$, $9.38\%$, and $12.22\%$ for nominal allowances $1\%$ through $5\%$, while the corresponding mean scientific-risk increases are $0.050\%$, $0.304\%$, $0.636\%$, $0.636\%$, and $1.653\%$. Simultaneous one-sided held-out risk inference supports all five operating points below their nominal caps. Action uncertainty uses $100{,}000$ paired bootstrap resamples conditional on the fixed reference and shared quadrature samples; exploratory simultaneous $95\%$ intervals have positive lower bounds throughout, including $[2.40\%,4.61\%]$ at $1\%$ and $[11.11\%,13.32\%]$ at $5\%$. One validation pair falls below the prespecified $\operatorname{rESS}=0.05$ support threshold, so complete numerical support certification is not claimed; all $32$ pairs are retained, and every leave-one-pair-out analysis at $5\%$ preserves an action reduction above $12\%$. Post-selection re-evaluation with richer Fourier and interaction potential spaces changes the reported reduction by at most $0.096$ percentage points relative to the base evaluator; these checks assess stability under the specified numerical perturbations and do not establish Galerkin convergence.

\subsection{Unbalanced Gaussian study}
\label{app:exp-unbalanced}

The finite-measure study uses a two-dimensional three-component Gaussian mixture on $\mathbb R^2$ over $t\in[0,1]$. Writing $s(t)=\sin^2(\pi t)$ and $\theta_j(t)=2\pi j/3+\pi t/2$, the component centers are $u_j(t)=1.35(\cos\theta_j,\sin\theta_j)+a\,s(t)(-\sin\theta_j,\cos\theta_j)$ and the common standard deviation is $\sigma(t)=0.38+0.06s(t)$. Trial-level nuisances $a\sim U[-0.30,0.30]$, $b_1,b_2\sim U[-0.65,0.65]$, and $c\sim U[-0.20,0.20]$ perturb the intermediate shape and mass while vanishing at the endpoints. With $\beta=b_1(1,-1,0)+b_2(1,1,-2)/\sqrt3$, the component masses are $m_j(t)=e^{cs(t)}\operatorname{softmax}(s(t)\beta)_j$, so the total mass is $M_t=e^{c\sin^2(\pi t)}$. The frozen analytical reference sets all four nuisances to zero, giving equal component weights, unit mass, zero reference source, and the same rotating centers and time-dependent variance. No reference model is trained.

Two labeled Gaussian sensors of width $0.45$ have centers in $[-2.2,2.2]^2$ with minimum pairwise separation $0.45$. Each trial observes $100$ particles from the normalized shape at eleven acquisition times $j/10$ and adds independent detector noise with standard deviation $0.01$; log mass is observed separately with noise standard deviation $0.02$. Shape moments and log mass are reconstructed independently by endpoint-anchored generalized least squares using three $\sin^2(\pi t)$-weighted polynomial basis functions, after which the normalized shape is information-projected relative to the analytical reference and recombined with the reconstructed mass. Scientific risk is finite-measure Gaussian-kernel $\mathrm{MMD}^2$ with bandwidth $0.55$, averaged over the ten non-acquisition interior times, so both shape and total-mass errors contribute to the admissibility criterion.

Law, Tangent, GlobalReaction, and Full are selected on $32$ common development trials using constrained SLSQP with differentiable objectives and the full $1\%$--$5\%$ Law-relative scientific-risk allowances. Tangent uses the transport--reaction Galerkin correction generated by the two measured observables together with the constant mass coordinate. GlobalReaction, defined in \eqref{eq:ub-global-reaction-action}, retains unrestricted transport but restricts reaction to the spatially uniform mass-rate correction, whereas Full solves the unrestricted screened transport--reaction problem. The authoritative action evaluator uses a cell-centered finite-volume discretization with no density smoothing or floor, float64 arithmetic, diagonally preconditioned conjugate gradients at relative tolerance $10^{-10}$, and $193$ trapezoidal action nodes. The primary $\kappa=1$ evaluation uses a $97^2$ spatial grid; the $\kappa=0.25$ sensitivity evaluation uses $129^2$, while $\kappa=4$ again uses $97^2$. Numerical certification checks moment calibration, projected effective sample size, captured Gaussian mass, projected covariance conditioning, screened and balanced PDE residuals, forcing centering, and the Move--Reaction and Tangent--Hidden energy identities.

After the five-point frontiers are frozen, validation uses one new bank of $128$ paired trials shared across all methods, allowances, and $\kappa\in\{0.25,1,4\}$, with no pooling of development observations. The primary confirmatory analysis is the $\kappa=1$ Full-versus-Law comparison. Action uncertainty uses $100{,}000$ paired bootstrap resamples with simultaneous intervals over the five allowances; held-out scientific-risk feasibility is assessed separately using simultaneous one-sided bounds for the paired excess-loss statistic. Tangent and GlobalReaction are retained as separate descriptive five-point comparator families. The $\kappa$ sweep reuses the same physical validation bank so that changes across $\kappa$ reflect the transport--reaction metric rather than different scientific observations; absolute action values are therefore compared only within a fixed $\kappa$.

\subsection{Computing environment}

Experiments were run locally on a system with an Intel Core Ultra 9 275HX processor, comprising $24$ physical cores with maximum turbo frequency $5.4$~GHz, and an NVIDIA GeForce RTX 5090 Laptop GPU with $24$~GB of dedicated GDDR7 memory.

Execution routes differed across studies. JAX-based simulation, reference training, and selected optimization or evaluation stages used GPU acceleration where applicable; empirical information projection and several differentiable proxy calculations used native C++/OpenMP implementations.

The computational environment was hosted in Windows Subsystem for Linux~2 (WSL2), running Ubuntu 24.04.3 LTS with Linux kernel 6.6.87.2-microsoft-standard-WSL2. The principal implementation used Python 3.12.3 and JAX 0.8.3 with 64-bit numerical precision enabled. GPU acceleration was provided through the JAX CUDA~12 backend.

Empirical information projection used a C++ backend where specified by the study protocol. At a given trajectory node, for empirical reference particles $x_i$ with base weights $\bar w_i$ and reconstructed target moment $\widehat c$, the double-precision Newton solver finds a finite multiplier $\lambda$ satisfying $F(\lambda)=0$, where $p_i(\lambda)$ is given by \eqref{eq:empirical-iprojection-weights} and
\begin{equation} F(\lambda)=\sum_{i=1}^{N}p_i(\lambda)\Phivec_{\etavec}(x_i)-\widehat c,\qquad \nabla_\lambda F(\lambda)=\Cov_{p(\lambda)}(\Phivec_{\etavec},\Phivec_{\etavec}). \label{eq:empirical-newton-system} \end{equation}
Thus failure to attain the requested moment and ill-conditioning of an otherwise calibrated interior solution are distinct numerical events: the former concerns empirical feasibility or solver convergence, whereas the latter concerns local stability of the multiplier coordinates and implicit derivatives. Multiplier estimates are warm-started between consecutive time nodes, while independent trajectories and linear-algebra operations are parallelized using OpenMP. The native solver is exposed to Python through pybind11 and integrated with JAX through Tesseract~\citep{hafner2025tesseract}; Jacobian--vector and vector--Jacobian products use implicit solves with the moment covariance in \eqref{eq:empirical-newton-system} rather than differentiation through the individual Newton iterations.

The C++ extension was compiled in release mode using the GNU C++ 13.3 compiler, the C++17 language standard, and the optimization flags \texttt{-O3} and \texttt{-march=native}. Fast-math optimizations were not enabled in order to preserve the numerical behavior of the double-precision projection solver. The complete software environment is summarized in Table~\ref{tab:software_specs}.

\begin{table}[!htb]
\centering
\caption{Hardware and software environment used for the experiments.}
\label{tab:software_specs}
\small
\begin{tabular}{lp{8.2cm}}
    \toprule
    \textbf{Component} & \textbf{Configuration} \\
    \midrule
    Operating system
        & Ubuntu 24.04.3 LTS under WSL2 \\
    Linux kernel
        & 6.6.87.2-microsoft-standard-WSL2 \\
    Processor
        & Intel Core Ultra 9 275HX (24 cores, up to 5.4~GHz) \\
    GPU
        & NVIDIA GeForce RTX 5090 Laptop GPU \\
    GPU memory
        & 24~GB GDDR7 \\
    NVIDIA driver
        & 581.57 \\
    Python
        & 3.12.3 \\
    JAX
        & 0.8.3 \\
    jaxlib
        & 0.8.3 \\
    Accelerator backend
        & CUDA~12 via the JAX CUDA plugin 0.8.3 \\
    CUDA runtime
        & 12.9 (JAX-packaged runtime) \\
    NumPy
        & 2.5.1 \\
    SciPy
        & 1.18.0 \\
    Tesseract Core
        & \texttt{tesseract-core} 1.11.0 \\
    Tesseract JAX interface
        & \texttt{tesseract-jax} 0.2.3 \\
    Projection backend
        & Native Tesseract C++/OpenMP batched trajectory solver \\
    C++ language standard
        & C++17 \\
    C/C++ compiler
        & GCC/G++ 13.3.0 \\
    Build system
        & CMake 4.4.2, release configuration \\
    Python/C++ bindings
        & pybind11 3.1.0 \\
    CPU parallelization
        & OpenMP 4.5 \\
    Native compilation
        & \texttt{-O3}, \texttt{-march=native}, without
          \texttt{-ffast-math} \\
    Native numerical precision
        & IEEE 754 double precision (\texttt{float64}) \\
    Neural optimization
        & Custom JAX implementation of Adam \\
    JAX numerical precision
        & 64-bit floating point enabled \\
    Job scheduling
        & Local execution; no external scheduler \\
    \bottomrule
\end{tabular}
\end{table}

\end{document}